# REAL-TIME ROAD TRAFFIC INFORMATION DETECTION THROUGH SOCIAL MEDIA

A Thesis presented to
The Academic Faculty

by
Chandra P. Khatri

In Partial Fulfillment of the Requirements for
Masters in
**Computational Science & Engineering, Civil Engineering**

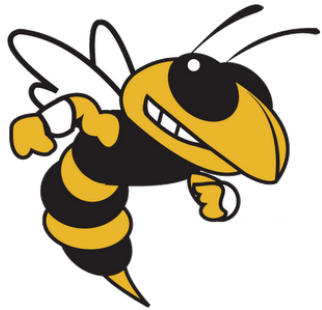

**Georgia Institute of Technology**
August, 2015

# REAL-TIME ROAD TRAFFIC INFORMATION DETECTION THROUGH SOCIAL MEDIA

Approved by:

Dr. Michael P. Hunter, Advisor
School of Civil and Environmental Engineering
*Georgia Institute of Technology*

Dr. Richard Fujimoto
School of Computational Science & Engineering
*Georgia Institute of Technology*

Dr. Bistra Dilkina
School of Computational Science & Engineering
*Georgia Institute of Technology*

Dr. Kari E. Watkins
School of Civil and Environmental Engineering
*Georgia Institute of Technology*

Date Approved: May 14, 2015

# ACKNOWLEDGEMENTS

I wish to thank Dr. Michael Hunter for helping me in changing my home unit to Transportation Systems Engineering and also for providing an opportunity to pursue research in the domain which I am passionate about. Furthermore, I would like to thank my parents, without whose support I would not be here.

# TABLE OF CONTENTS

# LIST OF TABLES

# LIST OF FIGURES

# LIST OF SYMBOLS AND ABBREVIATIONS

| | |
|---|---|
| TTI | Texas A&M Transportation Institute |
| Tf-Idf | Term frequency - Inverse document frequency |
| NLP | Natural Language Processing |
| AI | Artificial Intelligence |
| API | Application Program Interface |
| NN | Neural Networks |
| SGD | Stochastic Gradient Descent |
| NB | Naïve Bayes |
| POS | Parts of Speech |
| NER | Named Entity Recognition |
| LDA | Latent Dirichlet Allocation |
| NN | Neural Networks |

# SUMMARY

In current study, a mechanism to extract traffic related information such as congestion and incidents from textual data from the internet is proposed. The current source of data is Twitter, however, the same mechanism can be extended to any kind of text available on the internet. As the data being considered is extremely large in size automated models are developed to stream, download, and mine the data in real-time. Furthermore, if any tweet has traffic related information then the models should be able to infer and extract this data. To pursue this task, Artificial Intelligence, Machine Learning, and Natural Language Processing techniques are used. These models are designed in such a way that they are able to detect the traffic congestion and traffic incidents from the Twitter stream at any location.

Currently, the data is collected only for United States. The data is collected for 85 days (50 complete and 35 partial) randomly sampled over the span of five months (September, 2014 to February, 2015) and a total of 120,000 geo-tagged traffic related tweets are extracted, while six million geo-tagged non-traffic related tweets are retrieved. The classification models for detection of traffic congestion and incidents are trained on this dataset. Furthermore, this data is also used for various kinds of spatial and temporal analysis. A mechanism to calculate level of traffic congestion, safety, and traffic perception for cities in U.S. is proposed. Traffic congestion and safety rankings for the various urban areas are obtained and then they are statistically validated with existing widely adopted rankings. Traffic perception depicts the attitude and perception of people towards the traffic.

It is also seen that traffic related data when visualized spatially and temporally provides the same pattern as the actual traffic flows for various urban areas. When visualized at the city level, it is clearly visible that the flow of tweets is similar to flow of vehicles and that the traffic related tweets are representative of traffic within the cities. With all the findings in current study, it is shown that significant amount of traffic related information can be extracted from Twitter and other sources on internet. Furthermore, Twitter and these data sources are freely available and are not bound by spatial and temporal limitations. That is, wherever there is a user there is a potential for data.

# CHAPTER 1

# INTRODUCTION

Increasing traffic-congestion in urban areas is costing the U.S. economy billions of dollars (TTI Urban Mobility Report) [95]. Apart from economic impacts, traffic delays not only cause frustration to the drivers but also reduce the efficiency of people. As transportation is the backbone of any economy, when it is inefficient it impacts each and every person in the ecosystem. In 2011, each commuter in the U.S. had to face on average 38 hours of traffic delays, a 137% increase from 1982 (TTI Urban Mobility Report) [95]. Furthermore, traffic delays and congestion cost approximately 120 billion dollars in the U.S. in 2011 (TTI Urban Mobility Report) [95]. Surprisingly, congestion is now not restricted to "rush hours" alone. Approximately 40% of delay occurs during mid-day and overnight hours (TTI Urban Mobility Report) [95]. Therefore, those businesses which depend on deliveries and constant production are consistently affected.

With increasing congestion and delays, it is important to derive new and innovative solutions which are capable of addressing these challenges. Creating new infrastructure is costly. A better approach is to first optimize the transportation system where possible, improving the efficiency of existing infrastructure. However, more efficient system management requires improved real-time operational knowledge. For example, utilizing real-time monitoring to detect incidents or congestion may enable significant reductions in road traffic congestion through real-time optimization and system management. Effective management and near to real-time monitoring of traffic is important for smoother traffic flow. When commuters know that there is a blockage, a jam, or an incident ahead they can reroute their trip. DOTs and other local agencies may be able to control the situations faster when they are aware of these incidents or blockages.

In the last decade, several predictive technologies have been developed which when incorporated in transportation systems can be used to reduce the costs and increase the efficiency of the economy. Real-time frameworks not only help the users to pre-plan their trip but also improve mobility and safety within the city. Apart from real-time, when information is aggregated over longer time periods important insights may also be extracted, such as identifying locations which are most vulnerable to incidents, congestion, or gridlock. After identifying key locations, site specific treatments may be undertaken to improve the overall system performance.

Clearly, awareness of operations in the system is critical and this requires operations monitoring and data. Currently, most monitoring systems use sensors and other infrastructure that is expensive, time consuming, high maintenance, and limited to few spatiotemporal values. Thus it is not feasible to deploy monitoring systems that completely cover in time and/or space a transportation system. Therefore, many incidents and congestion occurrences go unreported in real-time, especially on the road where no cameras or other sensors are available. Moreover, conventional means of data collection often provide generic predictions that are ineffective during non-recurrent circumstances, such as incidents and emergencies. In additional more advance systems are still far off. For example, Dedicated Short Range Communication (DSRC) onboard units are expensive, not common, and not yet standardized. Advancing technologies (smart phones and IT applications) and rise of Internet of Things are providing alternative, and potentially cheaper, efficient information sources.

### 1.1 Goals

In the current study, alternative potential data sources for traffic information are identified and are used for various kinds of analysis. The goal of this study is to develop a novel framework and traffic monitoring system using these non-traditional data sources. It is hoped that this system will help enable future efforts to minimize the congestion and to detect incidents in real-time using these alternative data sources. Real-time information at a large scale is commonly obtained either through sensors embedded in the infrastructure or through probe vehicle data. Recently, with the emergence of social media, significant amounts of information are being shared in real-time among various users and communities. Social media has emerged as the mainstream medium to propagate news and information. Hence, Twitter, Facebook, Instagram, Snapchat, and other social media websites are acting as new platforms for data, where people themselves act as a sensors and share the information which they possess.

In this effort information shared by people on web is used as a source for extracting traffic related information. For this research, Twitter is used as the primary source of data. To date, limited research has been completed obtaining traffic information or incident identification using Twitter or other social media data. In first phase of this research the data is obtained in real-time and then stored in a database for offline analysis. In the second phase some efforts are made to demonstrate potential use of the data in real-time.

To accomplish this goal the analysis will initially leverage tweets that include geolocation information. For this effort the collected tweets are focused on urban areas to increase the likelihood of meaningful sample sizes. The included urban areas are those found in the TTI congestion index rankings (TTI Urban Mobility Report) [95], allowing for comparison of the develop congestion rankings to other known rankings. After obtaining the data, it is mined, i.e. cleaning and extraction, for traffic related data. Machine Learning, Artificial Intelligence, and Natural Language Processing techniques are utilized on the captured textual data so that traffic related relevant information can be extracted. Textual data contains contextual information and automated mathematical models do not readily understand this kind of contextual information. Automated mathematical models have difficulty making sense out of what is being reflected through the tweets. On the other hand humans are good at understanding the contextual information from the text and language related tasks, however, they are not good at scalability. That is, when millions of samples are presented to humans it is hard and very time consuming to make sense out of each and every single tweet. Therefore, Machine Learning and Artificial Intelligence techniques are used to achieve this goal.

To convert the data into a usable form, processing is required. Data Mining techniques are used to perform this task. Automated mathematical models are then prepared and used to extract the information and handle Big Data. Information extraction out of the textual data is aimed to retrieve as much traffic related information as possible. In the current study, the goal is to achieve the following traffic related information: traffic incidents, traffic congestion, gridlock, congestion, spatial and temporal variation of traffic, etc.

**1.2 Objectives**

Five hundred million tweets are shared each day (Twitter Statistics) [105]. These tweets contain significant amounts spatial-temporal information. Because users utilize twitter for personalized news, information, or thoughts/reflections, people publically share more information on Twitter than many other social networking websites. For instance, many times, people share their frustration through tweets, in real-time. In the transportation domain, this is an example of the type of information experts seek. It is also observed that many people share traffic incidents, which they find on the go. From a safety point of view this kind if information is very important and can be utilized in machine learning models to identify congestion and incidents.

In current study, an extensive analysis on tweets is performed to extract real-time traffic information such as congestion, incidents, traffic perception, etc. A platform for only traffic related data mining is developed. A method for collecting the traffic and incident data is proposed. Variation of vehicular traffic activity obtained from twitter data across different times and days of the week is observed. Two different types of analyses are performed: (i) offline analysis and (ii) online analysis. In the offline analysis, past data is aggregated over a five month period and analyzed to obtain the following information at a cities level:

i) congestion rankings

ii) safety rankings

iii) traffic perception

iv) aggregated vehicular traffic activity mapping

v) traffic flow

vi) topic modeling (obtaining topics such as accidents, grid-lock, congestion etc. in textual form from group of tweets aggregated at cities level)

While, online analysis is another term for real-time analysis and in current study following kinds of online analysis are performed:

i) vehicular traffic activity mapping and visualization in real-time

ii) classification whether a given tweet is traffic related

iii) incident classification.

Apart from various kinds of analysis, one primary objective of current study is to verify and validate the developed models. Traffic activities (incidents and congestion) for various cities and their validation with most widely adopted rankings such as TTI Congestion Rankings, Allstate Safety Rankings [95, 4], etc. is performed. Verification of the models is performed by visual inspection (mapping of the data) and checking whether the outputs of the models conform to the observed and expected results.

### 1.3 Implementation

In the first step tweets are streamed from Twitter for a set of urban areas. These tweets are analyzed to determine whether they contain any traffic related information. This task is achieved by using machine learning classification models. Once traffic related tweets are identified, they are then stored in the database for offline analysis. For this effort data is collected intermittently over the five months period from September, 2014 to February, 2015.

Traffic related tweets are then segmented in two groups: congestion related tweets and incident related tweets. Another machine learning algorithm is trained to perform this task. After aggregating traffic congestion and traffic incident tweets into separate databases they are used to perform spatiotemporal analysis. An algorithm for calculating a congestion index for the urban areas based on the tweets is included in the study. The developed ranking is then compared to the TTI ranking. In similar way, an incident index is developed for same cities. Incident index is also compared with existing rankings.

Also, a novel technique for obtaining a perception index for various cities is developed. The perception index attempts to reflect how people perceive traffic incidents, congestion, and the driving or commuting experience overall. Currently, there is no quantitative method to obtain traffic perception. Most of the existing perception studies are based on surveys. In the current study, a quantitative approach based on Natural Language Processing and AI is proposed. Furthermore, topmost topics for each city are derived. Topmost topics are those sets or combination of words which best represent the data (Topic Modeling, Wikipedia) [100]. For instance, suppose one city has majority of the tweets related to an incident while a second city has majority of tweets related to congestion. Then the first city will have topmost topics related to incidents while the second city will have topmost topics related to congestion. A topic model is a type of statistical model for discovering the underlying abstract topics that occur in a collection of documents. Documents in current study are tweets. The data which is analyzed in current study is text extracted topics for each city as a combination of words. Topics represent those combination of words which best describe the data.

Apart from these analyses, Tweet activity is mapped in real-time and offline on top of the U.S. maps for better insights. Tweet activity studies (i.e. the number of congestion or incident tweets being sent from an area) are also performed through visualizing the number of tweets sent. Tweets are sent by users acting as sensors. Congestion and incident related tweets are assumed to be sent by drivers, passengers, pedestrians, and other commuters. In the current study it is assumed that the tweet activity obtained from tweets is representative of the level of vehicle traffic (i.e., congestion and incidents) being observed by the vehicle traffic sensors (here users). These sensors might imitate the real vehicle traffic activity on various streets. Therefore, the tweet activity is considered as representative of vehicle traffic flow. Furthermore, the results obtained by this study are also verified and validated to depict that twitter and social media may provide an alternative source for extracting vehicle traffic information. Because information regarding vehicle traffic can be extracted from tweets, a concept of "Digital Traffic Signature" is introduced. Digital traffic signature (i.e. traffic tweets) for a city represents its "fingerprint" and vehicle traffic characteristics for that city. Every city has a unique representation or pattern of vehicle and tweet traffic flow. Therefore this concept is termed as a digital signature as it is unique for each city. Finally, the data definition, methodology, and observation sections provides descriptive information regarding the studies performed.

# CHAPTER 2
# BACKGROUND

Significant vehicular traffic related information likely exists on the internet which is being unused. In 2013, there was a study which revealed that 90% of the data on web has been generated in the past two years (IBM, SINTEF ICT, 2013) [35]. The reason behind this sudden growth is the emergence of smartphones. Significant quantities of data are being generated because smartphones are being used as sensors. Many researchers are using this information for various kinds of analysis and in various domains, such as medicine, stocks, emergency management, etc. Some researchers are also using this information in real-time to take immediate actions during mass emergencies such as floods, earthquakes, medical crisis, etc. The concept is to use people as sensors (who actually are the source of all the events) and obtain as much traffic related information as possible in real-time or offline.

## 2.1 Real-time Traffic Information

It is helpful to know traffic incidents and traffic blockages in real-time as actions can be then taken to resolve them. Furthermore some incidents, which are not on the main streets, do not get resolved quickly because they are not covered by traffic cameras, etc. However, such incidents can be reported by commuters through tweets and other social media. Real-time traffic information is so important that there are some smartphone navigation applications such as Google Waze [113] which provides real-time traffic updates. The kind of information these real-time applications provide are: construction related details on various streets, traffic congestion due to events and games, safety and visibility, accidents, objects in the street, congestion/jams on the streets, etc. These kind of information are very helpful for the commuters. For example, if a commuter knows that there is construction on a street and that the traffic is blocked he/she can re-route their trip. The real-time reporting on Waze happens through the Waze application. The application provides the interface for people to provide reports and update the real-time maps. In fact, people get rewards in the application for reporting. This kind of schemes attracts people to share the information. There are other applications as well which provide similar information such as Here, TomTom, Beat The Traffic and Mapquest [33, 101, 6, 50]. Mapquest provides the real-time traffic on a street from cameras deployed on the streets. All these applications help in reducing the congestion and incident hence improve the productivity.

## 2.2 Congestion Index

Controlling the traffic congestion is an important aspect to facilitate a more competitive metropolitan economy. Traffic congestion leads to increased travel times hence causes lesser time spent on jobs or lesser efficiency on various economic activities. An urban area where employees have lower travel times is likely to be more productive than the one where travel times are longer. Recently, Remy Prud'homme and Chang-Woon Lee at the University of Paris [72] depicted that productivity improves as number of jobs that can be reached by employees in a particular period of time increases. This study is performed after reviewing Korean and French urban areas. More time spent costs money to employers, employees, and shippers.

Because of the above mentioned issues with congestion, some studies even estimate the costs of traffic congestion (value of lost time and excess fuel costs). There exist various rankings for measuring congestion across the United States. Some of the most widely accepted congestion reports and rankings are from TomTom, INRIX and Texas A&M Transportation Institute (TTI) [101, 36, 95]. The methodologies adopted by these organizations is different from each other. One of the fundamental metrics used for providing the ranks in all these reports is excess time lost in traffic congestion during peak hours. There exists strong correlation between these three rankings, however there are significant differences as well. TTI adopts the data from INRIX and still presents different rankings than that of INRIX. TomTom index is simplest of all the reports and considers only travel time index to rank the cities according to congestion.

INRIX uses several parameters for calculating congestion level for various metropolitan areas (INRIX Ranking Methodology) [37]. Firstly, it calculated how much extra travel time is required when compared to free flow conditions during the peak hours. A ratio between the travel time during peak hours and travel time during free flow is considered as one input attribute. The next parameter used for ranking is "Wasted Hours" in congestion. The number of annual commute trips is assumed at 440 – equivalent to traveling to and from work 5 days a week for 44 weeks. "Wasted Hour" estimates are annualized and to create a monthly estimate of wasted hours, the annual result is divided by 12. Third and last parameter used by INRIX is travel times on congested corridors across most congested corridors and road segments. These travel times are then considered for obtaining the congestion level for each city.

TTI's study is the most exhaustive of all (TTI Annual Mobility Report) [95]. Although, it inherits the data from INRIX but they use many different kinds of parameters and provide different kinds of rankings based on these parameters. The variable and performance measures considered in this study are: travel speed, travel delay, annual person delay, annual delay per auto commuter, total peak hour travel time, travel time index, commuter stress index (travel time index in peak hours), planning time index, $CO_2$ production and wasted fuel, total congestion cost and truck congestion cost, truck commodity value, number of rush hours, percent of daily and peak travel in congested conditions and percent of congested travel. Because different people perceive congestion differently therefore ranks for each metric or performance measure is provided in this study.

It is speculated that some of the parameters which are used for calculating congestion create a bias towards large cities. Some of these biased parameters are: "Wasted Hours" in INRIX's report and most of the parameters in TTI. Reason for bias in these indexes is due to the fact that delays and travel times for commuters in big cities such as Washington, San Francisco, New York, etc. are much higher than those of the mid-sized cities. Comparing cities based on travel times and average delays may not capture commuter's perceptions. For people in San Jose 30 minutes of travel time and 15 minutes of delay is too much while for a commuter in New York that might be acceptable daily occurrence. Commuters get used to these delays and travel times. Therefore, it is important to also incorporate how people in different cities perceive traffic.

## 2.3 Existing Data Sources and Infrastructure

There exist different kinds of databases which are being monitored by various organizations. Some of the biggest databases are provided by U.S. Department of Transportation. There exist Intermodal Transportation Database [38] which includes information on the volumes of freight and passenger movement by mode, origin, and destinations; location and connectivity of transportation facilities; and a national accounting of expenditures and capital stocks on each mode of transportation and intermodal combination. Texas A&M Transportation Guides [94] also maintains the database or the links to various database which are related to transportation. Texas A&M guides contains and provides links to a wide variety of databases such as: air pollution trends, transportation planning records, congestion data, freight transportation, fatality analysis reports, data analysis on crash and fatalities, highway safety information, highway statistics, highway safety design models, lifecycle analysis models, bridge inventory and management, national household survey, transit database, automotive sampling system, state traffic safety information and data, energy data book, and many more. United States Census Bureau [110] also maintains several different databases, most of which are similar to those mentioned above. Many state DOTs also maintain their local databases. GDOT for example supports Georgia 511 [28] which provides crash, incident, construction, and other related information in real-time.

Apart from these government and publicly available data, there are some private players as well who collect traffic data. Transportation Data Sources [98] is a private entity which provides the data in the form of products: real-time and offline fleet information. INRIX [36] maintains one of the largest transportation database and provide service to various organizations including DOTs and automobile manufacturing companies for efficient commuting. TomTom [101] is another product which provides and maintains a similar database as that of INRIX. With emergence of smartphones Google, Apple, and Microsoft also collect data through GPS on the smartphones (CNET, 2014) [18]. This data is very crucial in estimating trips, speed, and volume data and also incidents and congestion. With almost everyone carrying smartphones, these smartphone databases are becoming the richest source of data. However, this probe data has lot of noise especially at arterials.

## 2.4 Crowdsourcing and Microblogging

Crowdsourcing is the process of obtaining needed services, ideas, or content by soliciting contributions from a large group of people (Crowdsourcing, Wikipedia) [21]. Crowdsourcing preferably happens through an online community, rather than from traditional employees or suppliers. Generally speaking crowdsourcing is used to subdivide tedious jobs to lots of workers, where each contributor of their own initiative adds a small portion to the greater result. The idea is to take work and outsource it to a crowd of workers. One example of crowdsourcing is Wikipedia. The principle of crowdsourcing is that more heads are better than one. By canvassing a large crowd of people for ideas, skills, or participation, the quality of content and idea generation is superior.

Collecting data for traffic such as counts, speed, grid blocks, etc. individually is very time consuming and requires lot of manpower. With the emergence of smartphones, people who are not at all involved in the data collection process but are commuters provide significant amount of information through mobile applications, social media, etc. As mentioned before, Google, Apple, Windows and other smartphone vendors collect the navigation related data from individual users who have their GPS on (CNET, 2014) [18]. With the sensors in these devices they collect information such as: velocity, blockages, trip routes, etc. Also, there are some applications such as Google Waze [113], where people do share traffic related information in real-time. These kinds of information are very useful. There are types of crowdsourcing, where people share the information without even being exclusively part of the data collection process. While these techniques cannot replace the conventional methods of data collection methods such as using cameras, sensors, etc. they may be able to supplement them. If the objective is to analyze a specific corridor or streets segment then these crowdsourcing techniques might fail because of insufficient data. Information provided through crowdsourcing is scattered but very useful. It is later shown in current study that on an average 2,200 traffic related tweets are sent each day in U.S. People share information about congestion, blockages, incidents, etc. While not all incidents or congestion is reported this information is useful for real-time analysis.

Microblogging is one similar concept, where people share information through micro blogs in the form of text. A microblog differs from a traditional blog in that its content is typically smaller in both actual and aggregated file size. Microblogs "allow users to exchange small elements of content such as short sentences, individual images, or video links" (Microblogs, Wikipedia) [53]. These small messages are sometimes called microposts. Microposts can be made public on a website and/or distributed to a private group of subscribers. Subscribers can read microblog posts online or request that updates be delivered in real-time to their desktop as an instant message or sent to a mobile device as an SMS text message. One example of microblogging is Twitter and Facebook posts (Microblogs, Wikipedia) [53].

There has been significant amount of information related to traffic published on these microblogs. Even though the percentage of such information is low when compared to the size of the content being published on these microblogs, even small percentages leads to large amounts of information. Most of the content on these websites is public, that is anonymous people can access this information. The idea is to effectively collect the traffic related public microblogs and extract information.

There are various pros and cons of these microblogs. Firstly, they are small in length, hence it is faster to extract the information as only key words are needed to be detected. Secondly, small messages can be shared and edited among the users easily. One major drawback with microblogs is that it is hard to detect sarcasm from the messages if the lengths are shorter (Maynard et al., 2014) [52]. For example: "Traffic is great at Tech Square!". This tweet grammatically and syntactically looks a positive one, however it might be a sarcastic and the user might be scolding the traffic.

## 2.5 Twitter

Twitter is an online social networking and microblogging service that enables users to send and read short 140-character text messages, called "tweets". Registered users can read and post tweets, but unregistered users can only read them. Therefore, Twitter users can create and share ideas, events and information instantly. Close to 16% of U.S. population actively uses Twitter (PEW Research) [69]. With such a large population using Twitter, increasing each year, a lot of information is exchanged over the internet in real-time. Twitter has become one of the fastest sources of information and its propagation through the population. There are many advantages of Twitter over other social media websites. Firstly, tweets (posts on Twitter) are short and therefore information can be easily extracted and stored. Second, large numbers of tweets contain geolocation. Thirdly, it has an extremely large coverage and user base across the entire U.S. and globe.

Twitter has already moved into the mainstream. In the United States, for example, Presidential candidate Barack Obama microblogged from the campaign trail using Twitter (Microblogging, Rouse) [77]. Even news channels now broadcast information in real-time to larger audiences through Twitter. Traditional media organizations, including The New York Times and the BBC, have begun to send headlines and links on Twitter. Other potential applications of Twitter include traffic and sports updates and emergency broadcast systems. Twitter is also being used by many small, large businesses, marketing agencies and organizations to promote their products and services (Business, Twitter) [107]. Almost every celebrity and brand have an official Twitter account. There has been various use cases in the past where people resolved their complaints and issues with service providers through Twitter.

Apart from these, Twitter has also been used to raise awareness through various public campaigns. Almost every community service nowadays furthers their goals and increases participation through these microblogging websites such as Twitter and Facebook. In case of any event such as sports (FIFA World Cup etc.), elections, Oscars, snow day, etc. there is a sudden spike in Twitter activity. Similarly, during the time of accidents or traffic incidents there appear sudden spikes on Twitter in the vicinity of the geolocation where these incidents happen. Although there appear spikes in activity for these traffic incidents on Twitter as well, however they are relatively small. But at micro level, they are observable and this information can be used to infer the incidents. There are many dedicated professional accounts for tweeting about vehicular traffic activity for each city. For example, it is found in current study that the City of Houston has around 10 dedicated professional accounts for traffic, one of them is "Total Traffic", which also maintains vehicular traffic activity for almost every metropolitan city in U.S. Traffic alerts on twitter are so widespread that a Google search for "traffic alerts Twitter" fetches around 87.8 Million results and on closer look almost all of these results in the top 20 pages made sense. It has become a day-to-day activity of the Twitter accounts of news and radio channels to constantly report the updates regarding the traffic.

**2.5.1 Effectiveness of Twitter**

As mentioned before, 16% of the U.S. population is active on Twitter and 500 million tweets are shared each day and out of this 122 million tweets from the U.S. alone (Pew Research) [69]. A significant portion of this information contains geolocation information. Therefore, Twitter contains lots of spatial and temporal information as well. Furthermore, information on Twitter is easily accessible in real-time. Campaigning, sharing, or promoting any information costs nothing. Therefore, it is one of the fastest sources of information. Because of these reasons, Twitter attracts many professional, academic, and general population users.

**2.5.2 Relevance in Academia and Research**

Apart from all these professional applications, Twitter is actively being used in research as well. There are large numbers of researcher that use Twitter data. At the time of writing searching Twitter on Google Scholar reveals 5.13 million articles and Microsoft Academic Research reveals around 2500 research papers. Publications exists in almost all domains of academia such as Computer Science, Engineering, Medicine, Economic & Business, Art, Agriculture, Law etc. Twitter data is so widespread in research that even Twitter itself provides grants for various research projects (Twitter Research Grants) [108].

Twitter data can be analyzed to find the patterns in human behaviors. There are many demonstrated use cases wherein Twitter's data is used for various kinds of predictive analytics. Recently, Choudhury et al. [16] from Microsoft researchers developed an algorithm that predicts those at risk for postpartum depression by analyzing the emotional cues from the tweets of pregnant women. Palen et al. (2010) [67], performed analysis to detect mass emergencies and disasters even before the scientists could validate. Verma et al. (2010) [111] performed Natural Language Processing for extracting situational awareness using tweets during mass emergencies. Chowdhury et al. (2013) [17] uses tweets and classifies crisis related messages. Marcus et al. (2013) [51] developed a methodology to process and visualize the data from the tweets. Periera et al. [68] performs text analysis on tweets for incident duration prediction.

Twitter data when combined with earthquake characteristics reveals ground shaking intensity, with least mean-squared error using Machine Learning approach (Burks et al. 2014) [12]. It can be inferred from this research that when information from Twitter is combined with a particular application it is able to provide much better insights. In another research effort it is found that increasing numbers of patients have turned to social media to share their experiences with drugs, medical devices, and vaccines. Based on this data, drug safety surveillance after monitoring pharmaceutical products on Twitter is established and is found to give very good results (Freifeld et al. 2014) [26]. Epidemic Intelligence is being used to gather information about potential diseases, outbreaks from both formal and increasingly informal sources such as Facebook and Twitter. As mentioned before, social network data can be very helpful to detect mass emergencies such as ebola, swine flu outbreaks, etc. Kostkova et al. 2010 [41] depicted the potential of social networks for early warning and outbreak detection systems for swine flu using Twitter.

Apart from the mathematical analysis various creative and insightful visualizations are also developed using Twitter data. As mentioned before, the activity on Twitter suddenly spikes during sports tournaments. During major sports competitions, such as the FIFA World Cup, billions of people around the world turn their attentions to the results and stories of the tournament. Many researchers have developed different kinds of visualizations to summarize the activities and events during the sports. Wongsuphasawat, 2013 [114] developed visualizations that summarizes a tournament in a way that provides both an overview of the tournament and match details. Rios et al. 2013 [76] visualized the pulse of 50 major cities across the world. Cities are clustered to find the groupings that are similar in the ways their inhabitants use Twitter. After performing the analysis clusters are developed based on culture and spatial-temporal relations. There has been many traffic related publications as well based on Twitter.

## 2.6 Twitter Use Cases for Traffic

There are several demonstrations where Twitter was used for obtaining traffic related information. One of the first implementations, where real-time traffic information was extracted and used for the public occurred in Jakarta (Kosala et al., 2012) [40]. The developed model was straightforward and did not extract any contextual information out of the tweets. Based on the words in geo-tagged tweets, traffic related tweets were detected with some statistical confidence and then they were used for real-time mapping. Minimal information extraction, such as spatial and temporal extraction was performed with these tweets. Ribeiro et al., 2013 [75] developed a Traffic Observatory, which was later used to detect and locate traffic events and conditions using Twitter. Traffic

Observatory is a text mining system that works on Twitter's stream, looking for relevant text patterns that indicate traffic condition in specific locations. Their approach is naive and does not involve any machine learning or text classification, based only on few text patterns defined in the observatory.

Later in 2013, Schulz et al. [83] used microblogs for real-time identification of car crashes using a machine learning approach. Initially, tweets were tagged through manual inspection whether they were related to traffic. Given the efforts required for manual-tagging the sample size was small that was used for the machine learning process. Furthermore, trying to fit the model with respect to features which are highly specific with respect to the small sample size resulted in model to overfit, leading to low accuracy, precision, and recall rates. Their model achieved precision and recall rates of 87% on non-traffic class, which means that 13% of non-traffic tweets will be classified as a traffic related tweet. In microblogs, such as Twitter, 500 million tweets are published each day. Given that traffic related tweets account for significantly less than one percent of tweets precision and recall rates of approaching 87% on the non-incident class leads to a traffic tweets database dominated by false positives.

In 2014 Chen et al. [15] performed road traffic congestion monitoring in social media with Hinge-Loss Markov Random Fields. In this research a novel probabilistic framework for the problem of traffic congestion monitoring is developed. A model is developed to identify congestion locations. Extensive evaluations over a variety of spatial-temporal and other metrics on Twitter and INRIX probe speed datasets is performed. The analyses are demonstrated for two major U.S. cities (Washington D.C. and Philadelphia). Wanichayapong et al. 2011 [112] developed a classification model to classify traffic related tweets vs. non-traffic related tweets. The sample size used is very small at around only 2000 traffic related tweets and the implementation for classification is naive. Not much of a feature extraction is performed for this study. Furthermore, a framework to obtain historic and real-time traffic conditions based on semantic analytics integrated with sensor data is built by Leuce et al. 2014. In this study a system is built to demonstrate how the severity of road traffic congestion can be analyzed, diagnosed, explored and predicted using semantic web technologies. A very straightforward model is built using words, semantics and temporal extraction only.

Extracting real-time events for traffic related information is one of the main objectives of current study. Sakaki et al. 2012 [80] tries to extract this information in real-time by considering people as social sensors. A framework is developed to first classify whether a tweet is related to traffic or not. In the second step spatial information is extracted and then finally this information is portrayed to drivers. Tweets extracted in this study are just based on key words "heavy traffic", and then the analysis is performed. There is potential bias involved in this kind of approach. Since, only "heavy traffic" is used as a feature. This bounds the research by limiting to this phrase. In fact not much efforts are performed for extension of traffic related information for anything other than this phrase. In current study such issues are resolved. Secondly, the sample size for tweets with these words is very small and they have not removed the professional tweets, which definitely have these key words. This kind of study is overfitting and has a limited scope. A major drawback is that it is extracting tweets from professional users and then transferring the information to drivers. Utilizing such professional tweets creates artificially high precision recall rates that might not be achieved using non-professional tweets.

Tostes et al. 2014 [102], used entirely different datasets: Foursquare and Instagram mobile applications. Using these applications people generally check-in (that they have been or they are doing some activity like eating, playing, etc. at some location). In this study, correlation between check-ins (using these applications) and vehicular traffic activity (using congestion information derived from Bing Maps) is obtained.

Most of the existing use cases do not consider metrics other than tweet itself and do not incorporate any contextual information (traffic flow, congestion analysis, traffic perception, extracting incidents, etc.). In current study, various techniques are developed to perform such traffic related analyses, which if implemented in real-time may be highly useful to the transportation industry. Furthermore, the size of the data being considered in current study is one of the largest in this domain.

Collins et al. (2013) [19] developed a transit rider satisfaction metric using rider sentiments measured from Twitter. Tweets were analyzed based on their sentiment score and various conclusions are drawn from these scores. It is a very simple implementation. An advanced analysis could provide information in depth regarding what is being portrayed through these tweets. Hasan et al. (2014) [32] performed urban activity pattern classification using topic models from online geo-tagged data. They have depicted that geo-location data from social media has potential for travel activity analysis. Furthermore, they adopted data driven approach using for pattern classification. They tried extending the models for user specific activity patterns.

## 2.7 Different Types of Analysis Possible With This Data

Tweets can be analyzed in real-time directly from the twitter stream or can be downloaded for significantly large periods of time to obtain higher level insights. It is already mentioned previously that tweets possess lot of traffic information, which can be analyzed in real-time. The following online (real-time) analysis are possible with the proposed framework:

1. Traffic information detection: Tweets appear in real-time in the form of streams. To extract traffic information out of these tweets, it is important to first detect whether a tweet contain any traffic information. This is achieved through machine learning and classification techniques, which are discussed in next section. After detection of traffic related tweets, other contextual information can also be extracted out of the tweets, which are explained in subsequent paragraphs.
2. Incident detection: If a tweet is identified to be related to traffic, the next step is to detect whether it is related to traffic congestion or a traffic incident. This detection is performed in similar way as that of the traffic information detection approach, however the model and data focuses on incidents.

3. Congestion and incident mapping in real-time: After identifying congestions and incidents, they can be mapped in real-time to reflect traffic congestion and related information, so that commuters can benefit from this information.

Apart from real-time, the following offline analysis are possible with the data being collected in the proposed framework. Note that the offline analysis is performed only on the traffic data and not on general tweets obtained from Twitter.

1. Spatial and temporal variation of traffic: The time and geolocation information from the traffic related tweets give a good representation of how traffic varies according to different locations and during different times of the day. The obtained information can also be used for verification of models. If the spatial and temporal variation of vehicular traffic activity on Twitter makes sense and is similar to what is obtained on ground then the model developed on top of the this data should also be logical.

2. Peak hour analysis: 60% of the traffic congestion occurs during the peak hours (TTI) [95] as most of the commutes happen during these times. It is important to control the congestion during these hours. Incidents during peak hours might lead to gridlock and traffic jams. Therefore, peak hour analysis should be performed separately.

3. City level analysis: After aggregating the geolocation information at the cities level, various kinds of analysis is performed. Some of these analyses are used for comparing traffic congestion and incidents between the cities. In fact, traffic flow patterns can be derived when data is aggregated at the city level.

4. Congestion index: Congestion Index is a quantitative measure which provides the degree of traffic congestion for various cities. This analysis or metrics has existed for many years. However, due to emergence of GPS and dependent technologies such as TomTom and INRIX significant quantities of data are now being collected. Since emergence of these technologies, the quantitative method to calculate congestion index has significantly changed. In current study, congestion indices for major cities are calculated and a quantitative method to obtain the congestion index is suggested.

5. Incident index: Similar to the congestion index, an incident index can also be obtained for various cities across the U.S. In concept, an incident index is similar to safety rankings for different U.S. cities, except it is reciprocal of safety rankings. Incident index reflects the propensity of a city to incidents.

6. Traffic perception analysis: Perception or people towards traffic varies based on demographics, age, sex etc. (Dejoy et al., 1992, Dejoy et al., 1989) [23, 24]. Most of the existing perception related analysis are based on surveys. Perception of people varies not only with respect to driving but also incidents, traffic signs, signals, etc. (Lund et al., 2009, Simsekoglu et al., 2012) [46, 88]. In current study, a quantification of traffic perception using Twitter data is performed. A general ranking or index of how people are sensitive to traffic conditions (particularly congestion and incidents is performed).

7. Topic modeling: A topic model is a type of statistical model for discovering the underlying abstract "topics" that occur in a collection of documents. In current study, tweets can be termed as documents. Topic modeling is an essential study for various kinds of analysis. A rich database of relevant topics or words can be created using topic modeling. Furthermore, most relevant topics are also used for extracting features for machine learning models. Apart from above mentioned uses, topic modeling can also provide information of how a city perceives traffic by detecting the top topics for each city.

8. Traffic flow and digital traffic signature: Traffic flow patterns give a good representation of traffic congestion and activity at various streets in a city. Some streets are more prone to congestion and incidents than others. Mapping aggregated data collected over a long period of time provides a sense of incident and congestion prone areas. MIT's Senseable City Lab (Grauwin et al. 2012) [29] has depicted a concept of "Signature of Humanity", by analyzing the data of communication networks. A similar kind of study can be performed on top of the traffic data collected at city level to obtain "Signature of Traffic". The idea is that each city has a unique traffic flow pattern and if it is possible to capture these patterns for each city then what is derived is a representation or signature of traffic for that city.

**2.8 Making Sense of the Data: Machine Learning and Artificial Intelligence**

Tweets are in the form of text. With millions of tweets being sent per day, it is not feasible to manually analyze each tweet and check whether it contains traffic information or not. The process of analysis and data collection should be automated. Automation is a framework where machines or computer programs are able to detect and extract the traffic information. This technique of learning from past data and to be able to predict in the future with reasonable accuracy is known as Machine Learning.

**2.8.1 Machine Learning**

A machine learning algorithm is an algorithm that is able to learn from data. "A computer program is said to learn from experience E with respect to some class of tasks T and performance measure P, if its performance at tasks in T, as measured by P, improves with experience E" (Mitchell, 1997) [57]. In simple words, it is an art of learning from the past, i.e. experience, and predicting the results for the unseen data, i.e. the future. Experience usually consists of allowing algorithm to observe a dataset containing many examples (training data). Each example is a collection of observations called features. The feature extraction is the method through which information is retrieved for each example and from the entire dataset. Once the features are extracted then they are used for a task. A task can be of many types: classification, regression, object detection, anomaly detection, density estimation etc.

The two most widely applied paradigms of machine learning are supervised and unsupervised-learning (Mitchell, 1997) [57]. Supervised machine learning is the search for algorithms that reason from externally supplied instances to produce general hypotheses, which then make predictions about future instances. That is, a class label is present which is then used as extra information in learning the models. For example, if we have labeled data where the class of each sample is already known to be traffic or non-traffic. Then this extra information in the form of label is used to train the models. In other words, the goal of supervised learning is to build a concise model of the distribution of class labels in terms of predictor features. The resulting classifier is then used to assign class labels to the testing instances where the values of the predictor features are known, but the value of the class label is unknown (Mitchell, 1997) [57]. That is, once the model is ready then the same model can be used to predict the label for those samples or examples for which labels are not present, hence automating the process.

In unsupervised learning, the labels are not present and just the examples are used to extract the information out of the data. In current study, only supervised learning is used for analysis. However, there are couple of clustering methods (Latent Dirichlet Allocation etc.) that are used for topic modelling, etc.

Machine learning is used in thousands of applications. Some of the most widely used applications are: spam detection in emails, image detection (Facebook, Instagram, Security etc.), search engines, most modern day predictive systems, etc. Through machine learning, it is possible to make automated mathematical models which can detect whether a tweet is related to traffic or not. This detection task is known as classification: classification into two categories: "traffic" and "non-traffic" classes. The algorithms which perform this detection are known as classification algorithms.

**2.8.2 Artificial Intelligence**

The theory and development of computer systems able to perform tasks that normally require human intelligence, such as such as visual perception, speech recognition, decision-making, and translation between languages is known as Artificial Intelligence (The New Oxford Dictionary of English) [66]. The mathematical modeling for traffic incident and congestion detection is more closely related to machine learning than Artificial Intelligence (AI). AI is more of an art than science. Extracting the information out of the text or any given data is more close to AI. In fact AI uses machine learning techniques to extract the information (Norvig et al. 2003) [78]. Apart from classification of tweets to traffic class or non-traffic class, we are also interested in what information is carried in tweets for example: where has the incident happened or what is the severity of the incident.

**2.8.3 Natural Language Processing**

When the data being analyzed is text, sound, or any form of language (human communication) then the study is classified as Natural Language Processing (NLP). Natural language processing (NLP) is a field of computer science, artificial intelligence, and computational linguistics concerned with the interactions between computers and human (natural) languages. As such, NLP is related to the area of human–computer interaction (Wikipedia, Natural Language Processing) [59]. So, NLP is a kind of AI which deals with natural language be it in any form (Manning et al. 1999) [49]. NLP is used widely in ad targeting on internet, recommendations on various e-commerce industry, web search, speech to text and vice versa (Apple Siri, Google Now etc.), and thousands of other applications. All these applications happen through information retrieval techniques on web. NLP techniques are used in current study for traffic analysis.

Using NLP many kinds of features or attributes which are important to create these mathematical models, or extracting some information, are extracted. In terms of textual data, using NLP many kinds of language related information are obtained. Information includes: syntactic representation of text, parts of speech for each word in text, type of entity for each word (named entity recognition: name, place, time etc.), topic of the tweet etc. The objective of NLP is to somehow convert the text (which is categorical information and cannot be used in mathematical models directly) to some numerical form to be able to be used in mathematical models. Usually, this is achieved by converting text into its numerical vector form. Vectorization is the general process of converting a collection of text documents into numerical feature vectors. Textual data can be converted into numerical form in many ways. This conversion of text to numeric vector form is nothing but feature extraction. There are various techniques of NLP for feature extraction. As previously mentioned some information which can be extracted out of text is parts of speech, named entity recognition, etc. These can be used as features and can be used in some numeric form. In current study, AI and NLP are extensively used to make sense of the textual data, which are discussed in later chapters.

## 2.9 Need of Current Study

Traffic monitoring is an essential part of efficient commuting. Currently, traffic monitoring is conducted through sensors, surveys, manual counts, probes, etc. Firstly, all these infrastructures sensors are limited to spatial-temporal coverage. Thus all the street segments cannot be monitored at all the times. Secondly, the installation and setup of these sensors is very expensive. Furthermore, their maintenance is an overhead. Even the data collected through probe vehicles are oftentimes noisy. The collected data through these approaches is then used to estimate information at those times and locations for which the information or data was not collected. As the majority of the streets are not monitored frequently many predictions are based on estimations.

As mentioned before, people act as sensors and they share a lot of information on web through various applications. Most of this information is freely available or can be easily extracted. In addition, tweets contain spatial and temporal information and people share information from nearly everywhere. In later chapters it is shown that there exists tweets from internal streets, service roads, etc.. Large numbers of tweets are collected from the streets which are neither highways nor freeways. Furthermore, people tweet throughout the day. Even during the night time if people find some information they tend to share it.

The idea is to extract as much information as possible from conventional and these kinds of alternative sources (social sensors). These techniques or sources of data are not new. Previously, mainstream media used to be print media (newspapers and allies). Incidents and congestion related information used to be shared on these media. In fact, incident related information is still shared on these sources. Later on, television media became mainstream source of information and it is faster than print media but most of the times not real-time and limited to spatial-temporal parameters. This medium is also costly and news channels tend to control what they want to show. Currently, social media has become a mainstream source of information and it has been shown previously that many successful campaigns, activities, and movements are being conducted on social media. There are advantages of these social media: i) people themselves are sensors, ii) data is real-time, iii) almost entire spatial coverage and iv) it is freely available. With so many advantages of social media, definitely it is now time to mine these mediums and extract the information.

If a bigger picture is seen than it is clear that information was being delivered in the past, in present, and will be shared in the future, just that medium is changing. In fact NLP, AI, and machine learning can be applied to print media (textual data), television media (sound: natural language, video: pattern recognition) and social media (text and images). Therefore, it can be concluded that even though the medium may change in the future from Facebook and Twitter to something else these NLP, AI, and machine learning based techniques would keep on extracting the information in more efficient and faster manner. In fact, there are many studies in transportation industry itself where researchers are able to predict when an incident is going to happen based on machine learning techniques (Li et al. 2008, Yu et a. 2013) [45, 116]. The data used for these researches is not text, it is generally data collected on the ground using the sensors. Therefore, with the advancement of these techniques, we may be able predict in the future when an incident or congestion is going to happen through the data from tweets or whatever constitutes the mainstream media of the future.

# CHAPTER 3

# METHODOLOGY

Data is the key for performing the desired analyses. Therefore, the first step is to effectively clean and process the data. The second step is to extract the features out of the stored textual data. After extracting the features they are used in mathematical models and analysis is performed. Models are prepared for offline and real-time analysis. Finally, extracted information is mapped.

## 3.1 Data Definition

**3.1.1. Data Source**

Currently, the project utilizes publicly available real-time tweets obtained from Twitter. A tweet is a short 140 characters message, which people share among friends, family, coworkers, and potentially any Twitter user that subscribes to that person's twitter feed. A tweet may also contain photos, videos, and links. Twitter provides various application program interfaces (APIs) for providing content. Twitter search and stream APIs (Twitter APIs) [106] are used to access public tweets at no charge. Out of all the adults using the internet (which is 84% of all the adults in the U.S.,) Pew Research [69] depicts that 19% actively use Twitter. Table 1 provides the demographics and usage statistics of Twitter obtained from Pew Research [69].

Table 1 Twitter Demographics Statistics

| Category | % of all Twitter users |
|---|---|
| Sex: Male | 59 |
| Sex: Female | 41 |
|  |  |
| Age: 18-29 | 49 |
| Table continues.. |  |
| Age: 30-49 | 29 |
| Age: 50-64 | 15 |
| Age: >= 65 | 7 |
|  |  |
| Education: High School Grad | 27 |
| Education: Some College | 36 |
| Education: College + | 37 |
| Income: < $30,000/year | 32 |
| Income: $30,000 - $50,000/year | 21 |
| Income: $50,000 - $75,000 | 18 |
| Income: > $75,000 | 29 |

Typically 500 Million tweets are sent per day across the globe (Twitter Statistics) [105]. Twitter provides approximately 1% of all the tweets from their streaming APIs to the users who want to access the real-time tweets (Twitter APIs, quora.com 2014) [106, 73]. The current effort focuses on tweets within the United States (approximately 24.4%, Pew Research [69]) and those tweets that are geo-tagged (approximately 20%, Wu et al. 2013) [115]. Thus, approximately 240,000 tweets are available per day (500,000,000 x 24% x 20% x 1% = 240,000 tweets/day). Table 2 depicts the described expected twitter data availability. It is cautioned that these estimates are based on the data availability statistics found on the Twitter website and the above cited sources. Some of the data sources mentioned are independent studies performed by various organizations and there is discrepancy in their findings. For instance, while some studies state that approximately 20% of tweets are geotagged (Wu et al. 2013) [115] other studies estimate geotagged tweets at 5%, with some as low as 2% (Burton et al. 2012) [13]. Other studies have also stated that significantly more than 1% of total tweets may be made available on any given day (Twitter APIs, quora.com 2014) [106, 73]. Thus, significant variability exists in the potential size of the available data set. Since, Twitter does not reveal the exact process of providing the tweets, it is possible that the data which is being provided might be biased. However, no study could be found which claims the same. Table 3 depicts the Twitter statistics obtained from current study when data was collected for 50 complete days (the entire 24 hours of each day is included in the data) and 35 partial days (less than 24 hours is included for each day based on various streaming issues) spaced intermittently over the span of 5 months (September, 2014 to February, 2015). Also, as stated, for this effort the collected tweets are focused on urban areas included in the TTI congestion index.

Table 2 Twitter Statistics Based on External Sources Including Twitter

| Tweets sent per day | 500 million |
|---|---|
| Tweets coming from U.S. alone | 24.4% |
| Geo-tagged tweets (Tweets with exact geo-location) | 20% |
| Approximate percentage of tweets obtained by Twitter through its APIs | 1% |
| Approximate number of tweets available for processing | 500,000,000 x 24% x 20% x 1% = 240,000 tweets/day |

Table 3 Twitter Statistics Based on Tweets Obtained in Current Study and Counts of Various Datasets

| Geotagged tweets coming from U.S. per day (For 50 complete days data) | 1.5 million |
|---|---|
| Unique tweets with word "traffic" from U.S. per day | 2200 tweets |
| (averaged over 50 complete days data) | |
| Percentage of geo-tagged tweets with word "traffic" | 0.147% |
| Non-traffic tweets count | 6 million |
| Traffic tweets count | 120,000 |

For this study, for the tweets collected using Twitter's streaming API, those which contain the word (or string) "traffic" and are geotagged are captured as raw-data for the initial analysis. Geo-tagged Tweets having the string "traffic" represent approximately 0.147 % of the sample or approximately 2200 tweets/day across the U.S. Note that while these tweets contain the string "traffic" it does not necessarily imply that they contain traffic related information, for instance the string "trafficking" likely is not related to vehicle traffic. Data Mining and Natural Language Processing approaches are performed (discussed subsequently) to determine tweets which likely contain traffic related information. Subsequent analysis will seek to apply machine learning algorithms trained using the traffic related tweets containing the string "traffic" to the full set of geotagged tweets to identify other traffic related tweets that do not contain the string "traffic".

In addition to the tweets with the string "traffic" approximately 6 million geotagged tweets were collected over the period of November, 2014 to December, 2014 on random days without the string "traffic" were also recorded. These tweets will be used in the analysis to represent non-traffic tweets. As stated above it is recognized that the possibility exists that some set of these tweets may actually be related to traffic however not contain the string "traffic". Ongoing efforts are seeking to help identify these tweets.

**3.1.2. Data Extraction, Traffic Dictionary, and Database Creation**

The process to generate the final database containing traffic related tweets consists of several steps. First all geotagged tweets are logged from the twitter stream. Next the tweets with the string "traffic" are segregated into a separate database. These tweets are then further filtered for tweets related to traffic based on the removal of tweets with the string "traffic" that are unrelated to vehicle traffic (e.g. trafficking) and removing traffic tweets generated by professional traffic monitoring services. Also, traffic related words (e.g. arterial, accident, etc.) present in the traffic tweets are determined.

3.1.2.1 Data Extraction

As stated, a tweet database is created to handle the streams of information being collected in real-time. Every time a tweet is extracted using the Twitter API the database is updated. Data consists of the following information and features (i.e. attributes): (i) tweet text, (ii) geolocation (if available), (iii) date, (iv) time, and (v) user_id. Real-time data may be extracted from the Twitter Streaming APIs (Twitter APIs) [106] using two different methods. In the first method, a rectangular bounding box with south-west latitude-longitude coordinates and north-east latitude-longitude coordinates is given. Tweets which have a geo-location and fall inside this bounding box are retrieved. In second method, tweets with a specific key-word can be obtained while requesting the data from APIs. For example, all the tweets which have the word "traffic" may be retrieved irrespective of the geo-location. Since, the current study seeks information for the entire U.S. the first method is adopted and the entire U.S. is defined by a bounding box

As an aside it is noted that in later analysis for the city level additional bounding boxes are created to identify the tweets that fall in the given urban area. To set the south-west latitude-longitude and north-east latitude-longitude coordinates of the rectangular bounding box for the urban areas coordinates are obtained using Google Maps covering the counties and statistical areas as listed in the Federal Census bulletin for statistical areas for that urban area (OMB Bulleting, 2013) [110]. It is recognized that some inaccuracies in the bounding boxes may exist as all counties in an urban area may not fit neatly with a bounding box, however, utmost care was taken while covering the counties for each urban area. Collected data is verified by the end of each week to see whether the results are consistent. If the counts come out to be varying significantly from expected, then the tweets are checked and visualized to find the anomaly. Many times these anomalies happened due to events such as game day (local effect), or tweets by/or corresponding to a celebrity. In such cases the anomalies (traffic activity due to celebrities etc.) are removed from the raw data.

After the U.S. tweets are extracted using the Twitter API they are filtered to segregate only those tweets that contain the string "traffic" (recall these tweets are all also geo-tagged as this is a condition of the Twitter API data extraction discussed above). For the data set approximately 150,000 tweets containing word "traffic" are identified. As stated previously, an additional 6 million geotagged tweets are also recorded that do not include the string "traffic", resulting in approximately 6.15 million total tweets in the analysis discussed in this report..

Of the 150,000 tweets with the string "traffic" many are unrelated to vehicle traffic, containing the string "traffic" in tweets containing "trafficking", "trafficked" etc. Table 4 represents all such words which contain the string "traffic" but are not related to traffic. Therefore a filter was created to remove these unrelated tweets by making a dictionary of the strings with "traffic" but likely unrelated to vehicle traffic. The potential non-traffic related strings were identified by manually reviewing the tweets with the string "traffic".

Table 4 Non-traffic words appearing in database

| 1 | antihumantrafficking |
|---|---|
| 2 | humantrafficking |
| 3 | trafficking |
| 4 | traficant |
| 5 | endhumantrafficking |
| 6 | airtraffic |
| 7 | sextrafficking |
| 8 | trafficked |
| 9 | trafficker |
| 10 | traffickers |

3.1.2.2 Data filter - Traffic and Incident Dictionary

A traffic dictionary was then created to allow for further filtering of the tweets containing the word "traffic". After the removal of non-traffic tweets words containing the string "traffic" such as "trafficking" from the database, the traffic dictionary is used to increase the likelihood of the tweets which are related to traffic. Traffic dictionary helps in identifying the tweets with the word "traffic" which are related to vehicle traffic. Any tweet containing two or more words from traffic dictionary including the string "traffic" are finally considered as traffic related tweet. This traffic dictionary is created by manually adding traffic related words and their synonyms. Initially, many words related to traffic were collected based on common terminologies used in transportation domain and their synonyms from online dictionary "thesaurus.com" (www.thesaurus.com) [97] were also extracted and stored. Next the most frequent tweets and most frequent words out of all the tweets with word "traffic" are collected. Those words which did not exist previously in the dictionary were added. Finally top topics were derived from the dataset. Topics are groups of words which best represent the data. Top topics are derived using clustering and topic modeling techniques which will be described in the later sections. All tweets which are close (keywords matching) to the traffic dictionary (words such as car, incident, street, congestion, avenue, hit etc.) are considered as traffic related tweets.

Finally, as part of the initial analysis it was seen that there are some professional accounts on Twitter which only publish traffic related tweets. These accounts tend to maintain a standard format of the tweets and purposefully send out the tweets as news to help the commuters in a city. These professional users tweet a significant number of traffic related tweets. For developing classification models, such professional tweets are excluded to remove potential bias to the locations with higher volumes of these professional accounts. For example, the Houston area was found to have a high number of these types of tweets. The data was processed to determine the User_id of these professional accounts. Currently, if a user is observed more than twice in the 150,000 "traffic" raw database in a single day (i.e. approximately 2,200 traffic tweets per day in the data set) then the tweets from that user are ignored for future analysis for that day. It is expected that this is currently overly restrictive, potentially blocking non-professional users and future efforts will seek to improve this filter. However, all such tweets are collected separately to extract the semantics of the traffic related tweets to help inform how the traffic information flows through these tweets so they may be used in later model development and training.

3.2.1.3 Traffic Database

Of the original 150,000 tweets with the word traffic approximately 30,000 tweets were discarded using the above described filtering process. These 30,000 tweets are not used for any part of current study. This database of filtered traffic related tweets (approximately 120,000 tweets) are stored in a database which is termed as **traffic database,** also referred to as Class 1. The six million non-traffic related generic tweets are stored in a database which is termed as **non-traffic database**, also referred to as class 0. Note that the discarded 30,000 tweets are important tweets, they might contain the word

"traffic" and other features corresponding to traffic, therefore these tweets should also be used in the training as a part of class 0. However, in current study this has not been implemented and is currently being performed as an extension to this study.

To determine the effectiveness of the described screening process 500 tweets were randomly selected from the **traffic** database. Based on manual judgement it was found that 497 of these tweets likely were related to vehicle traffic, with the remaining three unrelated to vehicle traffic. However, no further relation could be established to reliably identify the three unrelated tweets through an automated filtering process. Hence, no further screening of data was established and it is assumed that the error rate will be sufficiently small as to not influence findings. Lastly, it is assumed that with time and multiple iterations of the analysis and classifications, data quality will be improved. Table 5 represents the examples of traffic related and non-traffic related tweets.

In summary, a total of two major databases were created: i) *traffic database:* geotagged traffic related tweets containing word "traffic" and one or more word from traffic dictionary and ii) *non-traffic database:* a sample of geo-tagged tweets which do not contain the word "traffic". In current study, the machine learning models are completely trained on the traffic and non-traffic databases described above. Figure 1 and 2 represent the flow chart of all the different databases created and paradigms of analysis considered in this study.

Table 5 Classification traffic and non-traffic words

| Class | Examples |
| --- | --- |
|  | ('i hate traffic'), ('this traffic is ridiculous'), ('traffic sucks'), ('life is a traffic jam'), |
| **Class 1: "traffic"** | ('catch me in traffic'), ('stop and go traffic in downtown on i-10 katy fwy outbound between i-45 and washington delay of 8 mins traffic'), ('this traffic tho'), ('why is there so much traffic') |
| **Class 0: "non-traffic"** | ('#WeLoveLA #Clippers Kings 98, Clippers 92: Season's first loss - ESPN (blog)'), ('He who kneels before God can stand before anyone.'), ('President Obama is headed to @EvergladesNPS on #EarthDay to discuss the need to #ActOnClimate') |

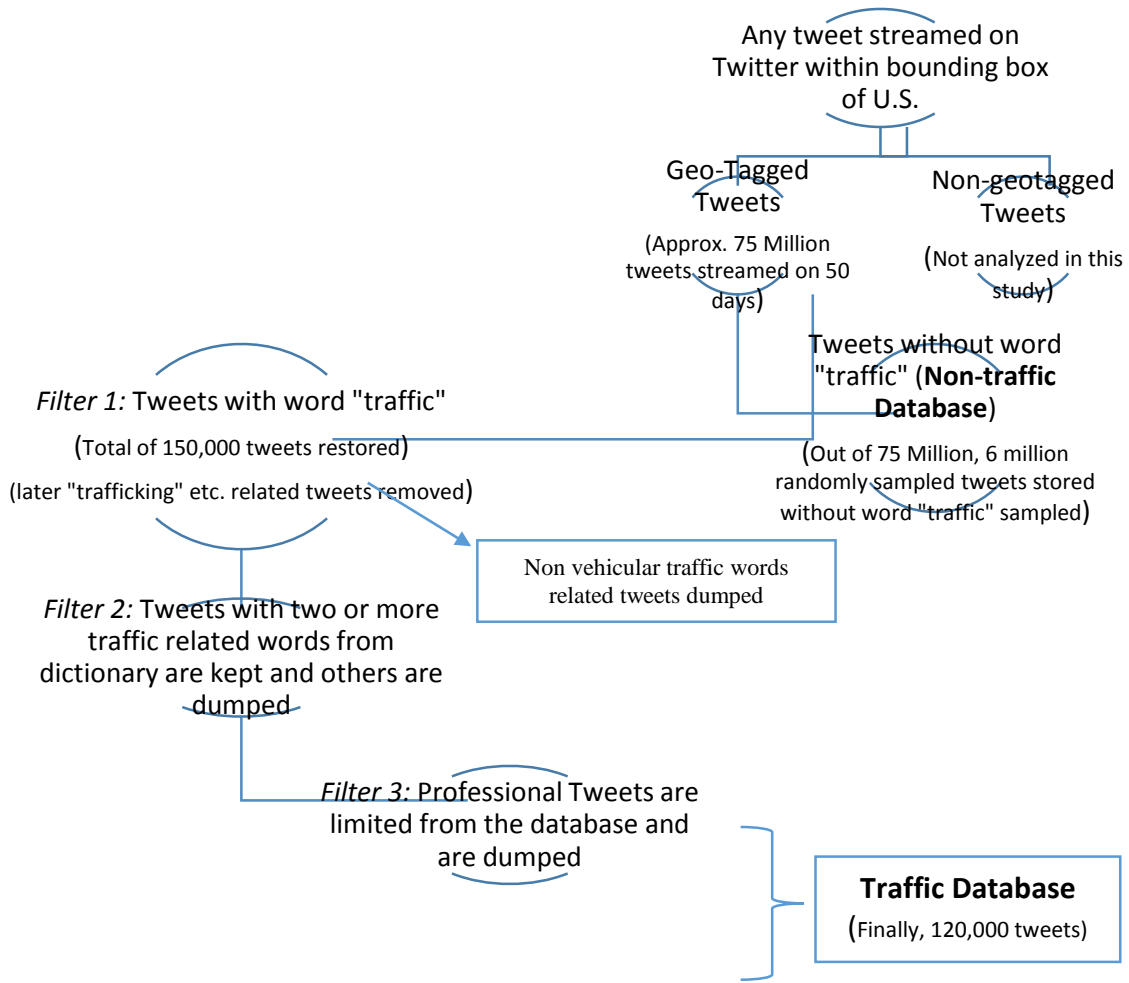

Figure 1 Data processed to develop Traffic Database

## 3.2 Processing the Data

It is to be noted that every tweet in the traffic database contains the word "traffic". The presence of word "traffic" in tweets creates a bias in classification models. That is, the model identifies that whenever the word "traffic" is present in a tweet, then it mostly would be related to traffic. To avoid this bias caused due to the word "traffic", another type of classification model was developed in which "traffic" word was removed from the tweets in traffic database. So, the database remains the same as traffic database for second classification model however the string "traffic" was removed from the tweets.

Note, that this is just an experiment that was performed to create an unbiased model. However, models are trained and tested on both the types of classification models that is with word "traffic" and same tweets with "traffic" word removed from traffic database. Therefore, there are total of two different types of classification models and corresponding databases (Figure 1, Figure 2):

i) *Machine learning models trained on tweets with traffic database:* all tweets in traffic database without removing word "traffic", hence each tweet contains the word "traffic" and there could be bias in the model due to this word.

ii) *Machine learning models trained on traffic database with "traffic" word removed from the tweets:* Same database as above, except "traffic" word removed from the tweets to remove the bias due to "traffic" word.

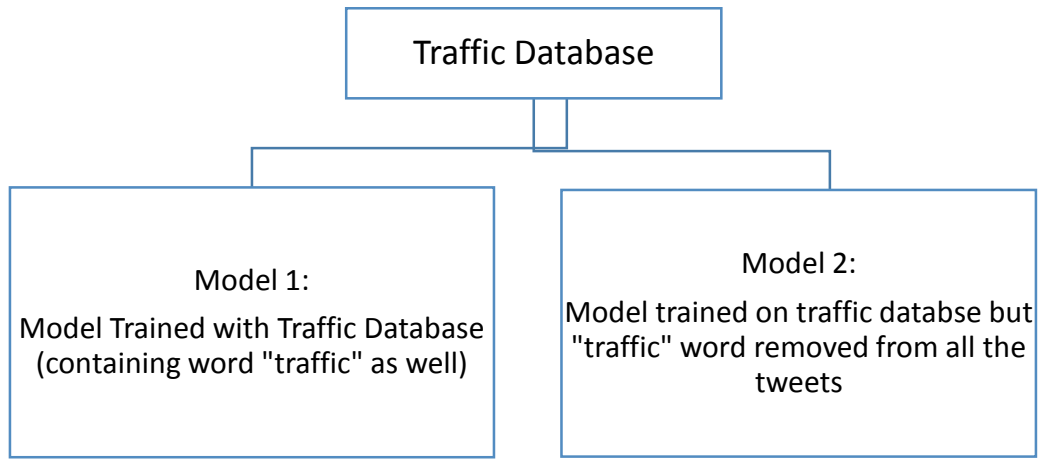

Figure 2 Different classification models trained and tested on various databases

### 3.3 Obtaining Features

Feature engineering is one of the most important aspects of model development. Classification models require those features (discussed below) which may be used to indicate the class (e.g. level of congestion) of the information (i.e. tweets). For example, in a multiple linear regression model, there are two or more explanatory variables which are used for developing a linear relationship with a response variable. In machine learning, these explanatory variables are known as features, while the response variable is termed as label, class, or target variable. The objectives of feature selection (i.e. which features should be included in the model) are: improving the prediction performance of the model, providing faster and more cost-effective predictors, and providing a better understanding of the underlying process that generated the data (Guyon et al., 2003 JMLR) [31]. Many kinds of features can be extracted from the given input. Some of the most straightforward features that may use for congestion classification are temporal data (e.g. peak hour), spatial data (e.g. urban area), and a traffic dictionary obtained from the tweets. The following section discusses all of the features that where derived from the Traffic Database for potential use in later modeling.

**3.3.1 Importance of Feature Selection**

Initially, it is not known which features are important, therefore it is important to obtain as many features as possible and to use them for developing an underlying relationships (i.e. model) between the features and target variable. It is expected that not all the features that are extracted will be used in the developed models, eliminating those that do not improve the model (Tang et al., Wikipedia: Feature Selection) [93, 25]. Furthermore, different features may be used in different kind of analysis. There are many reasons for not choosing all the features for any specific analysis. Some of the main reasons are:

1. Correlation

    If two numerical features are perfectly correlated, then one does not add any additional information. For a fixed number of training examples, increasing the number of features typically increases classification accuracy to a point but as the number of features continues to increase, classification accuracy will no longer benefit. Furthermore, correlated features leads to increase in the training time. Another problem with correlated features is that, if information from one feature is already extracted then the other feature adds little or no extra information. This leads to less weight to the second or features that are used later on (Tolosi et al.

2011) [99]. This give a wrong interpretation that the second feature is not that important compared to other features, even though it might be one of the most important feature. Because the information has already been extracted from another correlated feature makes the later one having lesser weight than other features. (Ando Saabas, Microsoft, 2014) [79].

2. Curse of dimensionality

   As the dimensionality (i.e., number of features) increases, volume of space that is mathematical space required to realize the relationship between target variable and features increases exponentially. This makes the available data sparse for training the classifiers (Curse of Dimensionality, Wikipedia) [22].

3. Reducing computations (training and testing time)

   Apart from the accuracy and space, the model with a greater number of features typically takes much longer for training and testing. Given limited computation power this can significantly slow the analysis process (Feature Selection, Wikipedia) [25].

4. Better understanding and visualization of information

   It is often easier to interpret the models with fewer features. Knowing the most impactful features in a model helps in better tuning of the model (Feature Selection, Wikipedia) [25].

A good model comprises of right features and selecting the right features is the most important part of machine learning models.

**3.3.2 Feature Extraction**

The majority of the data that is used in the current study is textual. This data is converted into some form of numerical metrics. Several different types of features are extracted from the data. Some of the main features that are extracted are:

1. Spatial Feature: The latitude, longitude and their pair that is a tuple of latitude and longitude (e.g.: 33.74,-84.35) called geolocation are used as features. Some locations are more prone to incidents and congestion such as highways, signals etc. The geo-locations are rounded off to two significant digits post decimal for faster computation. By applying this approximation the results can be obtained in the range of +/- 0.02 to +/- 0.067 square miles in the South-East and North-West corners of the U.S. respectively. The idea behind rounding off is to cluster the locations together. For example, if we want to consider a traffic signal for incident mapping, then we need to consider all the incidents in the vicinity of that signal. However, each incident although close to the signal will have its specific geo-location. Rounding off helps in grouping all the locations together in the vicinity of an incident or congestion prone location.

2. Temporal Feature: Traffic congestion often happens during the peak hours, special event, etc. Furthermore, statistics show that most of the incidents occur during the evening peak hours (Miley Legal Group) [55]. Therefore, it is important to consider temporal information for the classification. Note that 40% of the congestion occurs during the non-peak hours. However, only few major streets are congested in the non-peak hours. Aggregating the congestion for remaining 16 hours of the time (non-peak hours) leads to the figure of 40% congestion outside the non-peak hours. Also, city remains almost inactive during night time. So, clearly the pattern of incidents and congestion vary with time and therefore combination of temporal feature and spatial feature may be important. Temporal feature is obtained as a float number at minute level. For example, 5:30 PM can be used as 17 hours + 30/60 hours = 17.5 and similarly 9 Pm would be having value of 21.0. Therefore, temporal feature lies in the range of 0.0 to 23.98.

3. Length of tweet: Twitter has 140 character limit. Thus information has to be represented within this limited number of characters. However, it is expect that some minimal tweet size is commonly needed to portray traffic congestion and incident related information.

4. Word to vector representation (word2vec): It is important to convert words in the text in numerical vector format. Words cannot be used directly as features. Three different ways are adopted to represent words to vectors:

   i. *Count Vectorize:*

   Each word that appears in the corpus (linguistic data) is considered as one feature. The count (total number of occurrences) of each word is noted for each tweet. For example, consider two tweets: "traffic is bad" and "there is bad car crash". The first tweet is corresponding to congestion while the second tweet is corresponding to an incident. A possible corpus would be [["traffic","is", "bad"], ["there", "car", "crash"]]. Therefore, a corpus is list of all the tweets/document/samples at the token level. In the count vectorize each word becomes a feature. The count vectorize for each sample or tweet would be: [1,1,1,0,0,0] and [0,1,1,1,1,1]. Since, there are a total of 6 unique words in the data and each tweet adds some words to the corpus and the data the feature size is 6. The first three features are for the words from the first tweet (congestion related tweet) while the remaining three features are for the words from the second tweet (incident related tweet) which did not previously appear in the corpus. In this example, the count of each word is a feature. The count of the time the word appeared in a document/tweet is the numerical value of this word feature. So, "bad" and "is" appeared in both tweets hence they have value of 1 in both the vectors while all other words which are not there in the respective tweets have value 0.

   ii. *Term Frequency Inverse Document Frequency (Tf-Idf) Vectorize:*

Tf-Idf is a well know method to evaluate the importance of a word in a document or tweet. It is similar to Count Vectorizer except that it diminishes the feature values of those words which are common in all or most of the tweets. The idea is to amplify the effect of unique words with respect to each tweet and diminish the effect of common words because they do not carry any extra information about the document (Manning et al. 2008) [48]. In the above example of two tweets: "traffic is bad" and "there is bad car crash", "is" appears in both the tweets. Hence the power of "is" for reciprocating/reflecting different information (congestion or incident) reduces as it is common to both the tweets. To consider this crucial aspect in model, the count of a word across all documents (i.e. the number of times a word is occurs across all tweets) is divided by number of documents (i.e. tweets) in which they appear.

Tf-idf provides the importance of a word with respect to a document (here tweet) in collection of corpus (i.e. entre vocabulary of words obtained). Tf is an abbreviation of "term frequency". Term frequency of a document is the count of each term or word or token in that document or tweet. Therefore the term frequency (tf) tables for the two documents are:

Table 6 Term frequency table for first tweet (document 1)

| Term | Term Count (tf) |
| --- | --- |
| traffic | 1 |
| is | 1 |
| bad | 1 |

Table 7 Term frequency table for second tweet (document 2)

| Term | Term Count (tf) |
| --- | --- |
| there | 1 |
| is | 1 |
| bad | 1 |
| car | 1 |
| crash | 1 |

Idf corresponds to inverse document frequency. Inverse document frequency is a measure of how much information the word provides, that is, whether the term is common or rare across all the documents. Generally, the count of some words may be in millions for some documents and could be in single digit for other words, therefore to reduce the sensitivity with respect to counts of words, a logarithmic scaled fraction is considered. Idf is a logarithmic scaled fraction of the documents that contain the term, obtained by dividing the total number of documents by the total number of documents containing the term and then taking the logarithmic of that quotient (Wikipedia: TfIdf) [96]:

$$idf(t, D) = log \frac{N}{|\{d \in D : t \in d\}|}$$

where, N is the total number of documents in the corpus. $\{d \in D : t \in d\}$ corresponds to the number of documents where the term/word t appears. Since, it is possible that a term/word does not at all exist in a document/tweet and in that case this metric becomes 0. Therefore, generally the denominator is adjusted with an addition of 1, in this case it is ignored.

Tf-idf is calculated as: $tfidf(t, d, D) = tf(t, d) \times idf(t, D)$. Therefore, tfidf for the word "traffic" in the corpus D: [["traffic","is", "bad"], ["there", "car", "crash"] for document/tweet number 1 is:

$$tfidf(\text{t= "traffic"}, d = 1, D) = tf(\text{"traffic"}, d = 1) \times idf(\text{"traffic"}, D)$$

$$tfidf(\text{t= "traffic"}, d = 1, D) = 1 \times log\frac{2}{|1|} \approx 0.3010 = 0.3010$$

here, N = total number of documents in corpus i.e. 2 and the denominator is count of term "traffic". Tf-idf value for word "is" can be calculated in same way:

$$tfidf(\text{t= "is"}, d = 1, D) = tf(\text{t="is"}, d = 1) \times idf(\text{"traffic"}, D)$$

$$tfidf(\text{t= "is"}, d = 1, D) = 1 \times log\frac{2}{2} = 1 \times 0 = 0$$

It is clear from the above example that the impact of the word "is" is diminished to 0 as it is present in both the documents and hence does not add any extra information to the classification or feature space. However, the word "traffic" has a significantly large value when compared to the common words in both the documents of the example. Therefore, it is clear that tf-idf is a normalization technique for feature extraction. Generally speaking tf-idf normalization helps in reducing the impact of the common words such as: "the", "and", "a" etc., which appear many times in most of the documents. In addition these words are known as stop words. Term frequency (tf - word count) divided by document frequency (df) is used as a feature value instead of the counts.

iii. *Neural Network Vectorize:*

It is possible to capture many linguistic properties (gender, tense, plurality and semantic concepts such as time and capital city of a place) in the form of a vector using a machine learning technique known as Neural Network. (Mikolov et al, 2013, Wikipedia: Artificial Neural Network) [54, 2]. The idea is to find similar words or to make a transformation on the corpus such that the words that appear together get higher similarity score than those words which rarely appear together. This helps in getting all the abstractions of words which happen to reflect traffic (congestion and incident) class and those which do not reflect traffic class (i.e. which are more generic). These abstractions are obtained through non-linear transformation of words to vectors. The idea is to obtain a function which can map input to output using series of non-linear combination of weights. Google's word2vec [30] library is used to obtain these features. A detailed discussion of the implementation of Neural Networks is beyond this document and the reader is referred to the following paper: Linguistic regularities in continuous space word representations, Mikolov et al, 2013 [54].

5. *Named Entity Recognition (NER):* Named Entity Recognition (NER) labels sequences of words in a text which are the names of things, such as person and company names, or street and cities names (Sang et al. 2003) [81]. A well trained Stanford NER library [89] is used to obtain these features. It comes with well-engineered feature extractors for Named Entity Recognition, and many options for defining feature extractors. Consider the statement: "President Obama at five pm on January, 15 through WHO". The named entities for the above mentioned statement would be: [("President", "N"), ("Obama", "N"), ("at", "Prep"), ("five", "T"), ("pm", "T"), ("through", "PREP"), ("WHO", "ORG"))]. Here N corresponds to Name, T corresponds to Time, PREP corresponds to Preposition, ORG corresponds to Organization and LOC corresponds to Location (Natural Language Toolkit) [61]. Through this library the named entities can be extracted. Since, the data in current study deals with spatial, temporal variables and also names of various streets might also appear in tweets, NER provides a good representation of the same. Commonly used named entities are: organization, person, location, date, time, money, percent, facility and geo-political entities (GPE) (Natural Language Toolkit) [61]. A Boolean feature list for these named entities is created and as appended as NER features. Whenever a tweet has a word which corresponds to any named entity then that corresponding named entity feature is made 1 else it remains 0. For example, if a tweet has a word with "GPE" name entity then the feature value corresponding to "GPE" is made 1. In current study, if more than one word contains the same named entity still the count is kept to 1.

6. *Parts of Speech Tagging (POS):* Parts of speech (noun, verb, adjective etc.) or grammatical tagging for each word used in a tweet is obtained and are used as features in the model. The POS features can be used to extract the high-level syntactic and linguistic representation of the data (Manning et al. 1999) [48]. In the current study, POS features are obtained using Stanford's Log-linear Part-Of-Speech Tagger [90]. Consider the statement: "the little yellow dog barked at the cat". The parts of speech for the above mentioned statement would be: [("the", "DT"), ("little", "JJ"), ("yellow", "JJ"), ("dog", "NN"), ("barked", "VBD"), ("at", "IN"), ("the", "DT"), ("cat", "NN")]. Here NN corresponds to Noun, JJ for Adjective, DT for Determiner, IN for Preposition and VBD for Verb (Natural Language Toolkit) [61]. Stanford's POS tagger library provides the POS for each word in a tweet. Natural Language Toolkit [61] provides mentions universal part-of-speech tagset, which mentions 25 different tags. The POS features are used in the same way as NER features mentioned above.

7. *N-gram Model:* In addition to features discussed thus far a group of words in the order of their appearance can be used for extracting the features. In the traffic database many words happen to come together such as: "traffic sucks", "sucks at", "crash at" etc. These groups of words are can be very powerful in depicting information related to traffic. A single word model is known as uni-gram model, a two word model is known as bi-gram model, a three word model is known as tri-gram model, and so on. The more the number of words together, potentially the better the information extracted form one tweet (N-gram, Wikipedia) [62]. However, not every tweet is same and using higher order n-gram models may lead to over-fitting, potentially reducing the accuracy of the model. For example, a tri-gram model may result in tweets with very specific key word groups and hence generality is lost. Thus, while the predictive strength of the tweets reflected in the tri-gram model may be higher fewer data points will be represented than in the bi-gram model. Also, certain powerful two-word tweets will not be considered (e.g. "traffic sucks") in the tri-gram model. Furthermore, in tri-gram model many pairs of words might appear which not necessarily reflect any traffic information however are common in generic English statements. For example: "traffic is bad here"; in this tweet following are the trigrams, which are captured: ["traffic is bad", "is bad here"]. Now the challenge is that "is bad here" is a common phrase and might appear in many generic non-traffic related tweets. Therefore, the applicability of the model is lost and it becomes sensitive to common phrases. In this effort a comparison between a uni-gram and bi-gram model is made. Although, bi-gram model performed better with lesser features, unigram model gave similar performance as it was able to include other features. When using bi-gram model features such as parts of speech tagging, named entity recognition etc. are lost.

8. *Syntactic Features:* Exclamation marks, smileys used in the tweets, and total number of capital characters are also used as features. These provide emotional information. This data is usually sparse, hence these features tend to add less to the accuracy.

9. *Tweet Class Probability Score:* Words from traffic dictionary are used to obtain the probability that a tweet is traffic related. For example, if none of the words in a tweet appear to be in the traffic dictionary then the probability obtained is zero, otherwise the obtained log-probability is added as a feature. For example, if there are total of 1000 words in traffic dictionary and if a tweet has three words which exist in traffic dictionary then the log probability for this tweet would be log (0.003). This is very naive way of implementing the distance or likelihood with the traffic dictionary. Similarity obtained using top topics (next feature) is commonly a better and state-of-the-art technique.

10. *Similarity with Traffic Dictionary in Each Class:* Similarity is the representation of number of common words between a tweet and the traffic dictionary. Whenever a word in tweet appears in the traffic dictionary a counter is increased for that class (traffic/non-traffic) and the final counter (i.e. distance from the top-topics) is used as a feature value for that tweet. Basically counting the number of words that matches with the top-topics. For example, consider the traffic dictionary to have the following four words: ["traffic", "street", "segment", "crash"] and consider a tweet: "traffic is bad at 14$^{th}$ street". Now this tweet has two words which exist in the traffic dictionary. This number of word matches is used as a feature.

11. *Similarity with Top Topics using Neural Network Vectorize:* Neural network based implementation also provides the similarity of one word or feature with other words (Rehurek, Gensim Library) [74]. This similarity is much different from simple counter based similarity mentioned above. For example "San Francisco" has more similarity with "California" than "London". As mentioned previously this technique is shown to retain the syntactic representation of the data. Using the neural network based technique, similarity between the tweets and the top topics (obtained using LDA and most common tweets) is used as a feature. The difference between neural networks based similarity and distance based similarity mentioned previously is that neural network similarity is obtained through some non-linear functions which retain the syntactic and semantic characteristics of the tweets as well.

Apart from the above mentioned features, there is one more category of feature extraction: Auto-Encoders (Autoencoder, Wikipedia) [5]. Auto-encoders can be used for topic modelling and text classification (Mirowski et al.) [56]. Auto-encoder is a type of Artificial Neural Network (discussed above), which transforms the data by reconstructing its own input. They can be used to learn over complete feature representations of data. This technique is shown to outperform many supervised techniques for feature extraction (some of those mentioned above) (Mirowski , 2010) [56]. However, this approach of feature extraction is not currently implemented due to time restriction. For classification and model preparation all of the above mentioned features except auto encoders were extracted. Based on the feature importance, combination of different features is used for one analysis or the other. Table 8 represents various features and their application in the current study.

Table 8 Feature Table

| **Feature** | **Analysis in which feature is used** |
|---|---|
| Spatial Feature (as a lat-long pair) | Spatial Analysis, Incident and Congestion Mapping, City Level Analysis, Incident and |

|  | Congestion Classification, Peak Hour Analysis, Traffic Flow Analysis |
|---|---|
| Temporal Feature (at minutes level) | Temporal Analysis, Incident and Congestion Mapping, City Level Analysis, Incident and Congestion Classification, Peak Hour Analysis, Traffic Flow Analysis |
| Count Vectorize (Boolean list) | Tested in Classification Models. Not used finally. |
| Term Frequency Inverse Document Frequency (Tf-Idf) Vectorize | Incident and Congestion Classification |
| Neural Network Vectorize | Tested in Classification Models. Not used finally. |
| Named Entity Recognition (NER)- List of the size of named entities | Incident and Congestion Classification |
| Parts of Speech Tagging (POS) - List of the size of POS tags. | Incident and Congestion Classification |
| N-gram Model | Tested bi-grams models. Not used finally. |
| Syntactic Features | Incident and Congestion Classification |
| Tweet Class Probability Score | Perception Analysis, Incident and Congestion Classification, Congestion Index, Incident Index, Traffic Sentiment Score |
| Distance with Top Topics in Each Class | Perception Analysis, Incident and Congestion Classification, Congestion Index, Incident Index, Traffic Sentiment Score |
| Similarity with Top Topics using Neural Network Vectorize | Incident and Congestion Classification |

**3.4 Model Preparation and Experimental Setup**

Different kinds of analysis are performed, which are explained in the next section. To perform the classification whether a tweet belongs to a vehicular traffic activity such as a traffic incident or traffic congestion, a mathematical model needed. Some of the state-of-the-art techniques for classification come from the domain of machine learning.

**3.4.1 Machine Learning**

The technique of learning from past data to predict some aspect of future data with reasonable accuracy is known as Machine Learning. Humans are good at finding the similarities and patterns between small number of samples and when the data is small. When the data is in millions it is hard for a human to find the patterns. Secondly, when the complexity is less (linear or quadratic), humans can deal with it, however when the complexity in terms of dimensions and features increases humans are unable to differentiate. Therefore, we need to create mathematical models and corresponding computer programs to deal with these issues. Machine learning allows us to tackle the tasks that are too difficult to solve with fixed programs written and designed by human beings. For example, consider automating an algorithm which can detect traffic congestion or incidents from tweets. Manually, it is not practically possible to look up each tweet and assign it to a class. A classification model is capable of performing this task. In classification, the algorithm is asked to output a function $f : R^n \rightarrow \{1, \ldots, k\}$ (Kohavi et al. 1998) [39]. Here f(x) can be interpreted as an estimate of the category that x belongs to. In current study, the objective is to classify whether and how a tweet is traffic related. There are several machine learning algorithms which are capable of performing classification tasks such as: Naive Bayes, Logistic Regression, Support Vector Machines, Artificial Neural Networks, Decision Trees, Random Forest, etc. Some of these classifiers are used in the current study and are discussed in later sections.

**3.4.2 Natural Language Processing**

Natural Language Process (NLP) is a branch of Artificial Intelligence, where a computer program tries to understand human speech and text (Manning et al. 1998) [48, 49]. NLP is concerned with interactions between computers and humans. The objective of NLP is to infer the information contained in the text or speech. This is a difficult task since human text is often ambiguous and the linguistic structure can depend on many complex variables, including slang, regional dialects, and social context. Whereas computers are constructed to deal with precise and highly structured information. Machine learning and natural language processing techniques are applied to big datasets to improve search, ranking, and many other tasks (spam detection, ads recommendation, email categorization, machine translation, and speech recognition).

In current study, the data of concern is text and therefore NLP is the core of the entire study. The features that are extracted in previous section are different techniques of natural language processing. The text being analyzed is called natural language and the processing to infer relevant information is called information extraction (Natural Language Processing, Wikipedia) [59]. For example consider the tweet: "traffic is great at Tech Square". It contains the positive word "great" and grammatically and syntactically what a person may infer is that there is no traffic congestion. However, this is likely not the case as this tweet is likely sarcastic. What user is implying is that traffic is horrible. A strong NLP model is the one which can deal with ambiguity, sarcasm, and context within the text. There are several studies to depict that people tweet ambiguously and there is significant potential for sarcasm (Ibanez et al. 2011) [34]. Dealing with sarcasms is one of the toughest challenge in NLP.

### 3.4.3 Analysis

Once the data was collected and features were extracted three primary categories of analysis were undertaken as discussed in the follow subsections. In general, spatial and temporal features were used for spatial and temporal analyses. Features which maintain the linguistic structure, parts of speech, and dictionary based features were used for traffic perception analysis. For performing the analyses at cities level, the data is aggregated at the city level and analyzed to extract various information. For many of these analyses existing metrics and rankings, which are adopted widely, are used for validation of the approach. For topic modeling (determining the topmost topics from the data) clustering and database creation words were directly used as features. The three different categories of experiments that are performed in current study include:

1. <u>Online and Real-time Analyses:</u>

    Online analysis means analyzing the data as soon as it appears in real-time. At the time of writing this information, approximately 9,000 tweets are sent per second and 500 Million tweets are tweeted per day (Twitter Statistics) [105]. Performing real-time analysis on such Big Data is a challenging task. Firstly, it is important to make sure that the data stream does not break, that is, the stream of the data which is being downloaded and analyzed should not be interrupted. All errors such as connection issues, quota limits, data fallacy, etc. must be handled accordingly. Secondly, the needed data must be extracted in real-time to perform analysis. For each tweet, features are extracted and are used in classification models to classify whether a tweet contains any traffic related information. These classification models are trained using historical data, which is collected over several months. Dynamic models are built which can classify and store the data in the database accordingly. Hence, making the models richer and better over time.

2. <u>Offline Analyses and Visualization</u>

    The difference between online and offline analysis is that the objective is not to classify in real-time but to extract important contextual information from the traffic data. Contextual information is extracted from traffic database. Once the data is stored, it is analyzed offline. This kind of analysis is also important in obtaining a holistic picture of the traffic within a region. For example, collecting

one year data for a particular region can provide many insights on traffic behavior with respect to that region. Visualizing this information can be even more insightful for performing comparative analysis between regions. Visualization also helps in obtaining the variation of traffic congestion and incidents with respect to spatial and temporal parameters such as time-of-day, day-of-week, seasons, etc. In current study, various kinds of offline analysis are performed. Their implementation is described in the next section.

3. <u>Post-incident and Post-congestion Analyses</u>

In real-time, if it is detected that traffic congestion or an incident has occurred at a particular location then this information may be used to improve the situation at that location. The developed classification models can be used by various transportation organizations as an extra source of information apart from traditional sources such as cameras, loops, etc. Furthermore, there are many places which usually are not covered by cameras and other sensors for collecting traffic related information. Tweets can be observed at these locations as well. During an occurrence of congestion due to some natural incident such as snow etc., this technique can be highly useful. Based on prediction models congestion or incidents can potentially be reduced when status of traffic at various locations is known through the pattern of the information being received. In the current study, suggestions are made to use these models for post-incident and post-congestion analyses. In current study, post-incident analysis is not performed fully.

**3.4.4 Experimentation**

There are several different metrics to analyze machine learning models. It is important to check the performance of the model being built. Generally, model performance can be obtained using validation techniques. Out of all the analyses that are performed in current study, some are based on classification models and while others are not. One example of non-classification based analyses is obtaining traffic congestion ranking for various cities. To check the performance of the models, various statistical tests are performed.

For classification models, cross-validation technique is used. Cross validation is a model validation technique for assessing how the results of a statistical analysis will generalize to an independent data set (Cross-validation, Wikipedia) [20]. A model learns using some data set (training data) and then the same model is then used on a different, independent data set (test data). The model is validated based on predictions on this new data set. This is the basic concept for a class of model evaluation methods called cross validation.

To perform cross validation, some portion of the available data is used for training the model and the other remaining portion is used for testing model's performance. There are many ways of splitting the data. Generally, a portion of the data is randomly selected for testing with the remaining data used for training. The errors made by the model on the testing set are known as testing error. A model which gives the least testing error can be considered the best model. This type of cross validation approach is known as a hold-out method. The disadvantage of this approach is that the model may face high variance issues. If the data in the testing set or training set contains a high number of anomalies etc. then the testing error is going to be large.

To deal with this issue a more adopted technique known as k-fold cross validation can be used. The data set is divided into k subsets, and the holdout method is repeated k times. Each time, one of the k subsets is used as the test set and the other k-1 subsets form a training set. Then the average error across all k trials is computed. The advantage of this method is that it matters less how the data gets divided. Every data point gets to be in a test set exactly once, and gets to be in a training set k-1 times. The variance of the resulting estimate is reduced as k is increased. The disadvantage of this method is that the training algorithm has to be rerun from scratch k times, which means it takes k times as much computation to make an evaluation.

These methods tend to work well when the number of samples in each class is similar. Consider a case where the number of samples of one class (majority class) is significantly greater than the other class (minority class). In such a case, splitting the data for k-fold cross validation is further going to reduce the samples for the minority class for training. The model may not be trained well as it may not have sufficient samples for training the other class. Another problem with random selection of the data included in training and testing data sets is that most of the minority samples might get placed in the testing set leaving insufficient data in the training set and vice versa.

To avoid such issues, Stratified sampling can be useful. When subpopulations for each class within an overall population vary, it is advantageous to sample each subpopulation (stratum) for each class independently. Stratification is the process of dividing members of the population into homogeneous subgroups before sampling. The strata should be mutually exclusive: every element in the population must be assigned to only one stratum. The strata should also be collectively exhaustive: no population element can be excluded. This kind of sampling improves the representativeness of the sample by reducing the sampling error.

In current study, the population or samples have majority and minority class. To train the models without having sampling issues, stratified sampling is used. However, once the data was sampled with stratification 10-fold cross validation was performed to test the performance of models.

**3.4.5 Performance Measure**

Once the tasks are defined and models are built by using the extracted features, the models are then tested on some performance measure. In order to evaluate the performance of a machine learning algorithm, it is important to design a quantitative measure of its performance. Usually the performance measure P is specific to the task T being carried out by the system (Bengio et al.) [7]. For tasks such as classification, object detection (or model accuracy), one of the most used statistics, is used to determine the accuracy of the model. This is simply the proportion of examples for which the model produces the correct output.

If the model accuracy is high, that does not necessarily mean the model is good. For example, suppose there is a majority class and a minority class. That is, in current dataset 99.853% of the tweets are non-traffic related and while 0.147 % of the tweets are related to traffic. In such case if a model predicts all data as non-traffic the accuracy is 99.853%, which may appear as a good model. However, the model is not acceptable. To tackle this issue, other metrics such as precision and recall are used in the current study. Precision is the ratio of the number of relevant records retrieved to the total number of irrelevant and relevant records retrieved. Recall is the ratio of the number of relevant records retrieved to the total number of relevant records in the database. High precision means that an algorithm returned substantially more relevant results than irrelevant, while high recall means that an algorithm returned most of the relevant results. Relevant records above can be counts for traffic class or for non-traffic class.

### 3.5 Online and Real-time Analysis

As mentioned previously, online and real-time analysis are performed on the tweets appearing in real-time. The first objective is to classify whether a tweet in real-time is related to traffic. The second objective is to classify whether it is related to an incident or congestion. To obtain these objectives, classification models are built to detect the classes (congestion, incident, non-traffic). Many offline techniques which are discussed in the next section (section number 3.6) are also used. The following are the different analysis that are performed in real-time:

**3.5.1 Traffic Information Detection**

The approximately 120,000 instances of traffic which were obtained through data extraction and 6 million instances of non-traffic were used for developing the classification models. The objective is to build an algorithmic framework which is able to classify whether a given tweet or data is related to traffic or not. This will allow for determining the existence of traffic congestion and incidents in real-time from tweets. This goal is achieved by creating a mathematical model, which is trained from past data. To train the model, tweets were cleaned, mined and various filters were applied. All those tweets which passed the previously explained traffic related filters only are annotated as traffic tweets. While the data which does not have "traffic" word is used as the data for the other class ("non-traffic") class. This annotated data was collected previously and was used so that the algorithm learns what kind of tweets or data are related to traffic and what is a non-traffic information.

3.5.1.1 Model Preparation and Analysis

The technique used in the current study is supervised learning, in which we try to infer the function from the labelled training data. For example, say Y represents the class (1: "traffic", 0: "non-traffic") and say X represents the list of features ($x_1$, $x_2$, .., $x_n$), then in supervised learning the objective is to represent Y = f(X) such that for all the input values of $x_i$, a corresponding class (Y = 0 or 1) is obtained. So, in current study the previously labelled/annotated data were used to train the models to obtain the function. The following are the three steps involved in the model development:

*3.5.1.1.1 Feature Selection*

Out of all the features that are extracted (explained previously in feature extraction section) several are used for developing the classification models. In current study, Tfi-df vectorizer features were used instead of Neural Network vectorizer because the later did meaningfully improve the accuracy. Furthermore, extracting NN features is highly time consuming. Either of the Neural Network vectorizer or Tf-idf vectorizer could be used. Dimensions of these vectorizer models (Tf-idf and Neural Network vectorizer) is very high and therefore they both cannot be used simultaneously. The models are trained and their performance is calculated based on both kinds of models: Tf-idf and NN vectorizer. These word vectorizers carry significant amounts of information at word and syntactic level. There are several thousand unique words in the corpus and the dimensions of the features has already exceeded so much that it was not possible to perform the computations in current study due to hardware limitation. It is desirable to limit the number of additional features to limit the computation time. There exist many techniques to select the features for machine learning models.

A naive approach to select the features is correlation: obtaining correlation between target variable (1 for "traffic" and 0: for "non-traffic") and various features. Features which depict higher correlation (positive or negative) with the class feature (1: "traffic", 0: "non-traffic") can be selected as the primary features in the model. Another approach is to use all the features in the model and select those features which have high statistical significance. There are several strong classification methods. One such classification algorithm is Support Vector Machines trained using Stochastic Gradient Descent (described later) (Bottou et al. 2012) [11]. Such models eventually provide parameter or coefficient values for each feature. Based on the parameter values, features can be selected or dropped. This is also a very naive approach as there are other parameters also involved in the model such as regularization (used for generalization, handling the correlations among features, and avoiding underfitting and overfitting of the model). Therefore, this approach also has limitations.

A much better approach for feature selection is using Information Gain. Information Gain (IG) measures the amount of information in bits about the class prediction which is provided by any feature. To obtain IG, the only information available is the presence of a feature and the corresponding class distribution. Concretely, it measures the expected reduction in entropy (uncertainty associated with a random feature) (Mitchell et al., 1997) [57]. The expected value of the information gain is the mutual information *I(Y; A)* of *Y* and *A*, that is reduction in the entropy of *Y* achieved by learning the state of the random variable A. Usually a feature with high mutual information should be preferred over other features. Information gain is used to obtain the most relevant features. Note that word vectorizer features are not considered for feature selection because they provide the features at word level and add the most to the accuracy of the models. Although, dimensionality of the word vectorizer features can be used using dimensionality reduction techniques such as Principal Component Analysis (PCA), however, it is not performed in current study.

*3.5.1.1.2 Model Selection*

Because data size is significant and all the analysis is being performed on a personal laptop, classifiers that are fast and robust are used. Classifiers used in the current study are Stochastic Gradient Descent (SGD) (Bottou 2012) [11] and Naive Bayes Classifier (NB) (Norvig et al. 2003) [78].

*Stochastic Gradient Descent*

SGD is a discriminative learning approach of linear classifier under convex loss function such as Logistic Regression (Bottou 2012) [11]. Following is the pseudo-code for SGD (Stochastic Gradient Descent, Wikipedia) [92]:

- *Choose an initial parameter vectors (W) and a learning rate α*
- *Repeat until convergence or approximate minimum threshold is reached:*
  *Sample n examples in batches from training set*

$$\text{For } i = 1,2,3,\ldots,n:$$
$$\omega := \omega - \alpha\left(\frac{dQ_i(\omega)}{d\omega}\right)$$

where, $Q(\omega)$ is the is the true gradient approximated at each iteration. Each iteration can be performed either on each sample from training or in batches of samples. Batchwise SGD performs gradient descent on more than one example and provides smoother convergence (Mini Batch).

After convergence is obtained the same set of parameters which are obtained from the SGD are then used to predict the value of Y for given input X. For a given tweet, first features are extracted. These feature values correspond to the vector representation of X. If the value of dependent variable (Y) comes out to be 1 then it is classified as "traffic" tweet and Y=0 corresponds to "non-traffic" tweet.

*Naive Bayes Classifier*

The Naive Bayes Classifier technique is based on the Bayesian theorem and is particularly suited when the dimensionality of the inputs is high (which is the current case) (Norvig et al. 2003) [78]. Despite its simplicity, Naive Bayes can often outperform more sophisticated classification methods. Naive Bayes is a probabilistic classifier and is based on applying Bayes theorem with strong independence assumption between the features. It is one of the most prevalent classifier for text classification. Following is the algorithm of Naïve Bayes Classifier (StatSoft, Dell Inc.) [58]:

Given a set of variables, $X = \{x_1, x_2, \ldots, x_n\}$ the objective is to construct the posterior probability for the event $C_j$ among a set of possible outcomes $C = \{c_1, c_2, \ldots, c_n\}$ like class "traffic" or "non-traffic". In a more familiar language, X is the predictors and C is the set of categorical levels present in the dependent variable. Using Bayes' rule:

$$p(C_j \mid x_1, x_2, \ldots, x_n) \propto p(x_1, x_2, \ldots, x_n \mid C_j)\, p(C_j)$$

where $p(C_j \mid x_1, x_2, \ldots, x_n)$ is the posterior probability of class membership, i.e., the probability that X belongs to Cj. Since Naive Bayes assumes that the conditional probabilities of the independent variables are statistically independent, we can decompose the likelihood to a product of terms:

$$p(X \mid C_j) \propto \prod_{k=1}^{d} p(x_k \mid C_j)$$

and posterior:

$$p(C_j \mid X) \propto p(C_j) \prod_{k=1}^{d} p(x_k \mid C_j)$$

Using Bayes rule above, a new case X from training data is labeled with a class level Cj that achieves the highest posterior probability. An advantage of naive Bayes is that it only requires a small amount of training data to estimate the parameters necessary for classification. Some classifiers have been shown to outperform Naive Bayes such as Boosted Trees and Random Forests, but they require many training examples and also their computational complexity is much higher than that of the Naive Bayes. Since, current study is performed on a single machine, also the sample size and number of features are large therefore these classification techniques are not implemented. Furthermore, when trying to implement these algorithms the data did not fit in the memory of the laptop eventually they are not implemented.

*3.5.1.1.3* Experimentation

This is the last step of a single iteration of model construction. Many iterations are required for fine tuning of the model and every time the performance of the existing model is analyzed using experimentation. As mention previously, in current study Stratified Sampling is used for obtaining training set and testing set with stratification of datasets based on number of samples of each class. Furthermore, a 10-fold cross validation is performed for model selection.

Two different types of classification models are adopted for deciding the input for the models. In first classification model, each unique word in the dataset (traffic dataset, size: 120,000) is considered including word "traffic". Although, every tweet has the word "traffic", therefore the model will have very high correlation with class Y=1 ("class: traffic") and the model tends to be highly biased towards the word "traffic". To avoid this issue, another classification model where "traffic" word is removed from the tweets of the traffic database and then the same experiments are performed.

**3.5.2 Incident Detection**

Once it is detected whether a tweet is traffic related or not, the next step is to find whether it is related to any incident or not. One of the major problems with this classification is that the samples for such classification are small and with respect to number of features, the sample size tends to underfit. The classification technique used for this analysis is Naive Bayes and it is implemented in the same way as traffic detection. To make the incident database significantly large enough to be used for classification incident related words such as "accident", "crash", "hit", "emergency" etc. are used for data extraction. This filtering added significantly large number of samples (close to 1/6th of all the traffic related tweets), which are later used for incident classification. As mentioned before, the approach for this remains similar to that of the traffic detection method.

### 3.5.3 Congestion and Incident Mapping in Real-time

After classifying whether a tweet is traffic or incident related or not, it is visualized on top of maps in real-time. While visualizing any vehicular traffic activity, the intensity of an event, either congestion or incident, is obtained using posterior probability and number of tweets in that geolocation. The mapping/visualization is dynamic, that is, it updates with time. At a given time stamp the visualization would be different from another time stamp. The coloring of the tweet implies its intensity. The higher the number of tweets and output probability for class Y=1, more highlighted they are mapped on top of maps.

The mapping is based on geolocation information which was collected and also local time of the tweet or tweets. Furthermore, with geolocation information available in 20% (Wu et al. 2013) [115] of the tweets, it might be possible to use it to know how traffic is flowing within the cities during morning and evening peak hours. At micro-level the direction of traffic is also visualized with vehicular traffic activity on Twitter which is depicted in results and discussions. Visualization frameworks are designed so nicely that it is possible to see how the traffic is moving within a city in real-time. Furthermore, the same framework is also being used for incident mapping. However, the number of incident related tweets are very small in number and therefore it is not always possible to detect and map them in real-time.

### 3.6 Offline Analysis

Various kinds of offline analysis are performed in the current study. Each of these analyses provide different insight from other remaining analyses. Many of these analyses specially related to spatiotemporal information, safety and congestion rankings are also used to validate the study. Results of all the analyses are mentioned in the results and discussion chapter (Chapter 5). Following are the description of each of these analyses:

**3.6.1 Spatial and temporal variation of traffic (location and time)**

3.6.1.1 Spatial Analysis

Spatial analysis includes the techniques which study entities using topological, geometric, or geographic properties. There are many challenges and issues that arise while performing spatial analysis. One of the most fundamental challenges is defining the spatial location of entities being used. While performing the analysis it was difficult to define the boundaries for various cities. There are no standard boundaries defined for metropolitan and other cities across the U.S. There are many congestion and other related rankings provided by various organization such as Texas A&M Transportation Institute (TTI) [95] etc., but there is no common reference or metrics for considering the boundaries of cities while performing the analysis. The same challenge was faced while performing current study. Therefore, boundaries based on counties are considered and the list of counties for each city is obtained from Federal Census bulletin for statistical areas (OMB Bulletin, 2013) [110]. An exhaustive list of statistical areas for each metro, micro, and combined statistical area is provided in this bulletin. The coordinates of each city are obtained using this list and further the same coordinates are used as bounding box to obtain the tweets corresponding to each city.

After obtaining the tweets, vehicular traffic activity on Twitter for each city was calculated. A simple model based on counts was created, where all the vehicular traffic related tweets were counted and the count was used in obtaining the traffic activity. A novel tweet normalization technique was developed and was used for obtaining the cities statistics, which is explained in section 3.6.4. In current study, spatial analysis is restricted to 101 most congested cities according to TTI. Since, it is very time consuming to perform spatial analysis for each and every city, the aggregated vehicular traffic activity is mapped and visualized to see how it varies according to geo-location across the U.S. The same information can be used to find the activity at any location within the U.S.

3.6.1.2 Temporal Analysis

Traffic temporal analysis is the variation in traffic composition along with time in a study area. Temporal and spatial traffic information are correlated, however, time data is ordered, whereas there is no clear ordering in spatial data. The ordering property of temporal data is used to obtain the traffic variation with time periods. The United States uses nine standard time zones. In current study, only four time zones which constitutes majority of the population are considered: Eastern Standard Time (EST), Central Standard Time (CST), Mountain Standard Time (MST) and Pacific Standard Time (PST). Twitter data was collected at the local time in EST and therefore it was important to convert the time of collection for each tweet to corresponding time zone.

After time conversion, the data is aggregated and is used to analyze the variation of traffic across the day. Similar to spatial analysis, the aggregated vehicular traffic activity is mapped according to time to find the traffic activity at any location and at any time within the U.S. Furthermore, the information is also used to plot the histogram of traffic activity according to time. The obtained histogram is also used for validation and verification of the models built in the current study

**3.6.2 Peak hour and Traffic Flow Analysis**

Most of the traffic congestion and incidents happen during the morning and evening peak hours. It is found that 60% of the entire traffic activity lies in the morning and evening peak hours (TTI) [95]. Therefore, a dedicated analysis is performed for these periods. Initially, a simple count based model is used to obtain the traffic activity for cities during the peak hours. While analyzing the data over several months, it is very clear that the traffic related activity on twitter is maximum during the morning and evening peak hours. Since, more samples or observations are collected during these times it is evident to perform the analyses during peak hours. The impact of anomalies and outliers is reduced on the analysis due to larger sample sizes during the peak hours. Once the analysis is performed the findings are used to validate the models built in the current study. Aggregated vehicular traffic activity on Twitter is mapped during the morning and evening peak hours.

**3.6.3 Traffic perception analysis**

Perception and attitude towards the traffic changes from person to person. Perception is how people perceive traffic in their surroundings. For some people even a very small incident becomes a big matter of concern, while others people might be okay with harsh incidents and congestions. Perception is a very subjective concept and it changes from person to person. There are numerous studies which depict that there is a user bias involved in traffic perception (David M. DeJoy, 1989) [23]. For instance, they depicted that people are excessively and unrealistically optimistic when judging their driving competency and accident risk. Optimism increases with driving experience and marginally with age. Males tended to be more optimistic than females, particularly when judging their driving skill (Dejoy et al., 1992) [24]. In a comparative study between Turkish and Norwegian responds revealed that Turkish respondents perceived traffic risk to be higher than Norwegian respondents. Turkish respondents reported safer attitudes towards drinking and driving than Norwegian respondents, while Norwegians reported safer attitudes towards speeding. Turkish respondents reported a lower frequency of speeding behaviors than Norwegian respondents, whereas Norwegian respondents reported a lower frequency of drinking and driving. Traffic risk perception was related to road safety attitudes and behaviors among Norwegian respondents but not among Turkish respondents (Simsekoglu et al. 2012) [88].

Another cross-cultural study (Lund et al. 2011) [46] between India, Norway, Sub-Saharan African countries, and Russia revealed significant differences in traffic risk perception, attitudes towards safety, and driver behaviour. For example, Tanzania reported the highest willingness to take risks both in traffic and in general. Participants from Sub-Saharan Africa and India reported safer attitudes in regard to speaking out to an unsafe driver, rule violations and sanctions, attitudes towards pedestrians, and traffic rules and knowledge. One study even suggested that professional drivers may constitute a risk group in road traffic because they are involved in more accidents than the non-professional drivers, who are more cautious (Nordfjaern et al., 2012) [63].

From all these studies it is very clear that traffic perception is subjected to the type of user and there is significant variance involved in traffic perception information. In general traffic perception is obtained by using conventional survey based methods. There are limitations with these survey based methods:
1. It is hard to pursue survey at city level and usually the sample size is very small
2. Humans tend to limit the survey by imposing assumptions, which hides the true nature of the problem
3. Performing surveys across many cities is very costly and time consuming

All the above mentioned studies are based on surveys but currently there is no quantitative representation of user-perception in collecting the traffic information or finding the congestion index. Therefore quantification of the problem provides information about many hidden parameters, which are hard to infer due to limitations in data or samples in survey based methods. Furthermore, it is very hard to incorporate traffic perception feature in the conventional methods of calculating the congestion index.

Currently, there exists no standard ranking system regarding traffic perception. In the current study, a unique ranking system is developed and then this ranking is used to reflect the traffic perception of people in various cities. TTI and other congestion rankings do not consider traffic perception as a factor. All the existing rankings are based on ground truth data on congestion, delays, and other demographic details. Traffic perception is obtained at the city level and is then used for comparative analysis between the cities. Since, tweets are sent by individuals, it can be a good source of information for traffic perception. When the tweets are collected at the city level, the combined level of traffic perception can be obtained for each city. Twitter is well suited for obtaining the perception of people which conventional methods for calculating congestion index etc. fail to do. Because Twitter provides information at user level (that is considering the perception at individual) unlike survey based or traffic count methods, tweets are a better representation of perception.

3.6.31. Traffic Perception Score

Sentiment analysis or opinion mining for Twitter is a very well-studied problem. Sentiment analysis aims to determine the attitude of a speaker or a writer with respect to some topic or the overall contextual polarity of a document (Sentiment Analysis, Wikipedia) [82]. The attitude may be his or her judgment or evaluation. In current study, the sentiment of each tweet is obtained using a dictionary developed by Stanford's Natural Language Processing group [91]. Each word in the English dictionary has a score. Using these scores and the syntactic representation of the sentence or document log probabilities are calculated. These probabilities represents how positive or negative the document is.

In the current study, documents are tweets. Moreover, traffic congestion and incidents generally carry negative sentiments. This is also evident as most of the top words and topics obtained from the data possess negative sentiments or are curse words. In current study, sentiment analysis is obtained by deriving the log probabilities of tweets using syntactic and semantic representations of the tweets. Obtained log probability represent the probability of the tweet to be of positive sentiment. These probabilities are then multiplied with "traffic sent score" for each tweet (explained in section 3.6.9). This concept is new and is defined in the context of this study. When we talk about traffic perception, then it should incorporate two things: (i) how grammatically and syntactically a tweet is positive or negative (that is does the sentence depict a positive or negative message/feeling) and (ii) how much the tweet has to do with traffic. If a tweet is related to a traffic incident then it would have higher negative scores while if it is related to congestion then it will have lower negative scores. When multiplied with log probability of the sentence being positive or negative provides the grammatical representation of the sentence as being positive or negative.

Traffic sent score is calculated using the traffic dictionary. The traffic dictionary also has manually annotated scores. Incident related words have much higher negative score than simply congestion related words and so on. A cumulative score for each tweet is obtained by summing up the value for each word within a tweet (if it exists in the database). This score is known as "traffic_sent_score" and the "traffic_sent_score" is then multiplied with the log probabilities calculated for each tweet to obtain the overall perception score of the tweet. Section 3.6.9 describes the traffic sent score.

If p is the log probability of tweet being positive obtained using syntactic and semantic representation of the tweet. Then the perception score for each tweet is calculated as:

*if sent_score is negative:*
    *perception_score = (1-p)\*traffic_sent_score*
*else:*

$$perception\_score = p*traffic\_sent\_score$$

What this represent is that if a traffic_sent_score itself is positive then p*traffic_sent_score represents probability multiplied by actual score. Similar to a posterior representation of the perception of the tweet and vice versa. Once the perception score for all tweets are obtained then they are aggregated at the city level and the mean of all the scores of each city represent the actual aggregate perception of the traffic for that city. These perception scores are later visualized for top cities and also used for verification of the models in current study.

An example related to traffic perception can be taken by considering Washington D.C., which is the most congested city in U.S. according to TTI [95]. The average travel time and delay of commuters in the most congested cities would be much higher than the average travel time and delay of other cities. A commuter getting delayed by 15 minutes might be okay for a commuter in Washington D.C. however this delay would be too much for a commuter in Pittsburgh or any other lesser congested and medium sized city. That is, people may perceive the same traffic condition differently in different cities. Therefore, the actual traffic index and congestion index provided by TTI, TomTom etc. [95, 101] could be supplemented by adding perception information to rankings.

**3.6.4 Congestion index**

To understand the level of congestion at various places, it is important to determine an appropriate congestion measure. Texas A&M Transportation Institute (TTI) and INRIX [95, 36] published the Urban Mobility Report, wherein congestion trends of various cities across the United States is provided. The 101 most congested cities in U.S. are ranked based on various metrics such as yearly delay per auto commuter, travel time index, congestion cost per auto commuter etc. TomTom [101] also provide Traffic Index for the Americas, China, and Europe.

Aftabuzzaman et al., 2007 [3] has developed a framework for developing a measure of public transport congestion. A range of features have been suggested for a measure of congestion. Turner et al., 1992 [103] examined indicators of congestion and suggested that measures to quantify the level of congestion should:

1. "deliver comparable results for various systems with similar congestion level"
2. "accurately reflect the quality of service for any type of system"
3. "be simple, well-defined and easily understood and interpreted among various users and audiences."

Levinson et al., 1996 [44] discussed desired attributes of a congestion index and suggested that a congestion index should:

1. "communicate easily"
2. "measure congestion at a range of analysis level (a route, subarea or entire urban region)"
3. "measure congestion in relation to a standard"
4. "provide a continuous range of values."
5. Be based on travel time data because travel time based measures can be used for multimodal analysis and for analyses that include different facility types
6. Adequately describe various magnitudes of congested traffic conditions.

In the current study, a novel approach for obtaining congestion index is developed. Traffic related tweets are collected at cities level and then are normalized based on Twitter activity for each city. Some cities are more active on Twitter than others. Furthermore, the twitter activity of each city is also used for normalization. The obtained normalized score is then used to obtain congestion index for various cities in U.S. Normalization of the congestion scores is important to avoid any bias towards number of active twitter users per city and effect of professional tweets.

3.6.4.1 Tweet Normalization

Twitter data depends on the online (internet) activity of the people as well as the absolute size or population of a city. Cities such as Atlanta, Houston, Philadelphia, and Dallas are not the most populated or largest cities in terms of area or population but are among the most active cities on Twitter. Therefore, normalization of the data or tweets for the analysis based on the population or statistical area is not correct. Furthermore, there is no standard study or research about rankings of the cities in the order they tweet across the U.S. Therefore, a simple technique was developed to obtain the individual penetration of each city on Twitter as part of this study. Around 6 million geo-annotated tweets were collected and the rankings for 101 most congested cities is obtained. This ranking was based on the number of tweets collected over a long period of time divided by the population of that city. This ranking is called "Cities_Activity_Rank" and is used for normalization of the traffic related tweets for each city. Instead of obtaining number of traffic tweets per person or number of traffic tweets per square feet a novel normalization factor called traffic tweets per city_activity is calculated. This normalization factor is then used for all the comparisons. Cities are ranked in the descending order of the tweet normalization factor. A high score implies that the city has high vehicular traffic activity on Twitter. This technique works for small cities as well because they might not rank high in number of traffic tweets but would be high when normalized with Cities_Activity_Rank, which would be low for such cities.

Tweet normalization score for each tweet is calculated and then aggregated at city level. The mean of all the aggregated normalized scores is calculated to obtain the congestion index for each city. Later the congestion indexes are used in validation and verification of the models and the data being used in current study. Congestion indices for each city are also visualized on top of maps for comparative analysis.

**3.6.5 Incident Index and Safety Rankings**

After clustering the topics and aggregating the traffic sentiment scores for each city, a perception of safety based on the tweets may be developed. The top topics obtained are used to develop the incident database. The technique used here is the same as tweet normalization technique, which is used in congestion index. The obtained scores are then compared with the results of traffic incidents, fatalities etc. provided by Allstate's America's Best Drivers Report [4]. The obtained incident indices for each city are also visualized on top of maps for comparative analysis.

**3.6.6 Cities Statistics, Traffic Flow and Signature of Traffic**

Using the bounding box obtained from United States Census Bureau (OMB Bulletin, 2013) [110], tweets are extracted for 101 most congested cities. A tweet normalization factor is obtained for these cities using the Tweet Normalization mentioned previously. Once the data is aggregated at cities level for the span of over five months, traffic flow patterns and related information at various times of the day are calculated.

3.6.6.1 Digital Signature of Traffic

Using the framework built for visualization, it is possible to detect the vehicular traffic activity varying across different times of the day. Heat-maps (intensity) of traffic activity are used to visualize the activity. When this information is aggregated at the city level, traffic flow patterns are obtained for each city. As Twitter data is unique from other traffic data collected this provides novel traffic visualization. Twitter has not been previously utilized for similar traffic analysis. The city level information is may be thought of as the signature or representation of the traffic for each city and therefore a new concept of "Digital Traffic Signature" is coined in current study. It is the named as digital signature because the data which is being used to obtain the signature is through internet in digital form and it varies and adapts according to how people reflect traffic in the digital media.

**3.6.7 Topic modeling**

Topic models are a suite of algorithms that uncover the hidden thematic structure in document collections. These algorithms provide state-of-the-art ways to search, browse, and summarize large archives of texts. Generally topic modeling algorithms are statistical methods that analyze the words of the original texts to discover the themes that run through them, how these themes are connected to each other, and how they change over time (Blei, 2012) [9]. The objective is to find the main topics or themes that pervade a large structured/unstructured collection of documents. Using topic modeling, the collection of documents can be organized according to the discovered themes.

Documents in current study can be referred as tweets. Top topics out of the traffic related tweets are extracted and are used for various purposes. While obtaining the top topics, stop words ("the", "is", "are" etc.) are dropped and are not used since they do not provide extra information and they appear in almost all the topics. Top twenty topics are derived from all the aggregated raw data using a state of the art unsupervised topic modeling technique called Latent Dirichlet Allocation. Obtained topics represent the perception of people towards the traffic. Apart from the topics obtained through LDA, the top 30 common tweets are also obtained.

3.6.7.1 Latent Dirichlet Allocation (LDA)

The method assumes that each document is a mixture of small topics and that each word's creation is attributable to one of the document's topics. The idea is to find the topmost topics out of all the documents. Topics which can represent or capture the intention of the maximum part of the documents will be classified as top topics. Similarly, a topic is combination of words and those set of words which can provide near to the true representation of a topic will be classified as the body of the topic. It is a generative approach and Dirichlet distribution is used to assign the words to topics and then topics to words. This process is repeated many times by updating using Gibbs sampling (Latent Dirichlet Distribution, Wikipedia) [42]. After many iterations, the algorithm converges to near to a true solution and the topics stabilize. The topics obtained at this point represent approximate representation of words and the topics out of the set of documents. In current study tweets are the documents and combination of words are topics. The detailed explanation of LDA is out of the scope of this study, however it can be referred from Blei 20123 et al. "latent Dirichlet Algorithm", 2013 [10].

In current study, topic modeling is used for developing a traffic dictionary, as mentioned earlier. Top topics derived are also used in classification for traffic congestion and incidents. Distance from the top topics of the training data set is used as one of the features for classification models (closely related to clustering, where these topics are considered as clusters). Topic modeling can also be performed for each city and the findings can be used for obtaining traffic perception at a city level. Furthermore, if a city has more incident related topics then information related to safety in various designs within the city can be obtained. After clustering the words for different cities using topic modeling many inferences can be obtained regarding how the city perceives traffic. Grauwin et al [29] from MIT's Senseable City Lab has derived the "Signature of Humanity" by analysing the data from communication networks. Similarly, "*Signature of Traffic*" is also obtained for each city using current analysis.

**3.6.8 Extracting information from non-geotagged tweets**

There are two different ways of obtaining the public tweets: geo-location boundaries and keyword. In current study, only geo-tagged tweets are considered and are used for analysis. But this geo-tagged data is only 20% of all, and the remaining information regarding the traffic is lost as it is not possible to capture almost 80% of the information. To deal with this issues, classification models which are built can be useful.

Sometimes, even non-geotagged tweets contain some geolocation information such as city, past location of the user, etc. These information are provided by Twitter as an external parameter apart from the tweet. Recently, IBM [47] published a paper where they were able to find the location of homes of Twitter users at different granularities, including city, state, time zone or geographic region, using the content of users' tweets and their tweeting behavior (Mahmud et al. 2014) [47]. However, the primary objective of this study is not to find the geo-location but to extract traffic information with respect to some location. An extension to this study and an implementation of location finding can make this study more robust.

The online classification models can predict whether a tweet is related to traffic or not. Once those tweets which are classified as traffic related tweets, they may be used is subsequent analysis. Firstly, it may be determined whether the traffic related tweet is for congestion or incident related. Later it can also be used for performing many offline analysis mentioned previously such as city level statistics, congestion and safety levels for each city, etc. Furthermore, with time and appearance of more data, the classification models will become richer. Furthermore, the model will also be able to handle the changing pattern of the traffic and its perception with time and season within a given demographics. This is kind of an iterative process can be used in the long run to extract information with the geo-tagged and non-geotagged tweets with time.

**3.6.9 Traffic Sentiment Score**

Traffic Sentiment Score is a score which reflects the level of frustration of commuter tweeting about the traffic. Lower scores imply more frustration and negative scores imply significant dissatisfaction about the traffic of the user. The lowest scores are obtained for the accidents, crashes, and other traffic incidents. It is also used in calculating the traffic perception as mentioned previously. The traffic sentiment score is obtained by breaking the tweet at the token level (word level) then obtaining the cumulative score for each word within the tweet. The cumulative score of a tweet is the final traffic sentiment score for that tweet. The score for each word is obtained from a sent score database (Figure 3), which consists of the words and their sent-scores.

| 49 | street | -2 |
| 50 | congestion | -3 |
| 51 | insane | -2 |
| 52 | shut | -2 |
| 53 | collision | -4 |
| 54 | car | -2 |
| 55 | bus | -2 |
| 56 | alleviate | 1 |
| 57 | hate | -1 |
| 58 | traffic | -2 |
| 59 | vehicle | -2 |
| 60 | stopped | -3 |
| 61 | drive | -2 |
| 62 | driving | -2 |
| 63 | injured | -4 |
| 64 | slow | -3 |

Figure 3 Sentiment Score Database

All of the words in the Traffic Dictionary discussed previously are manually assigned a Sentiment Score. The score for a word is intended to reflect its relative importance. The current score assignment is based on judgement. Incident related words such as "crash", "collision" and curse words were scored the least while positive words such as "good", "lucky", "awesome" etc. are scored positively. Stop words such as "the", "and" etc. are inserted with zero score, since they do not impact the sentiment of a statement. An example for the traffic sentiment score for the following tweet is depicted in Figure 4. It is readily recognized the current scoring system is based on the bias of the score assignments and different word score assignments could change the analysis. Future efforts will seek a formalized method to assign word score values and an understanding of the sensitivity of the analysis to variations in score assignments.

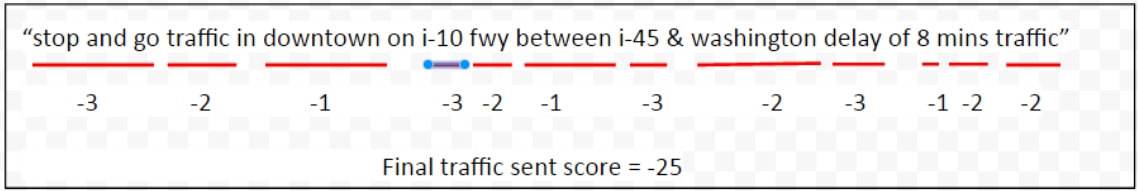

Figure 4 Traffic Sentiment Score calculation

### 3.7 Visualizing Offline Information

Various kinds of offline analyses are performed in current study that are mentioned in previous section. Visualization of these analyses greatly aids in the ready interpretation of the results. The following are the analyses which are visualized with the data collected over time:

1. Spatial Analysis & Cities Level

    The information extracted at cities level and spatially across the entire United States is plotted on top of a map and a comparative analysis between the cities is performed. This kind of visualization also provides insights about the most active demographics for traffic activity. Heat maps can also be drawn which allow for better insights and provides information of how traffic is trending spatially.

2. Temporal Analysis

    Visualizing temporal information is very important to know how traffic activity trends with time across the U.S. Moreover, the same information can be used to see the traffic inflow and outflow during the peak hours across the cities leading to traffic flow analysis. Since, offline analysis includes high volumes of data collected over time, anomalies are diminished and general trends and patterns can be observed.

3. Incidents over Time

    Visualizing incidents can provide the information regarding the areas or geolocations which are most prone to incidents. As mentioned previously, there

can be many locations which are not directly on the highways or main streets hence might not be monitored actively, but are prone to incidents. A benefit of the data collected from Twitter is that it is crowdsourced allowing for those locations which are not actively monitored by local communities and DOTs to be brought included as long as user of the social media are active in that location. Once identified with incident-prone locations, corresponding steps can be taken to make the situation better and cities safer to improve quality of life.

4. Visualizing Top Topics: Spatially and with Time

    Traffic perception related information can be obtained using topic modeling and same can be visualized on the map for various cities. During an event such as snow or celebrity accidents (e.g. death of a TV anchor in traffic accident in Atlanta (2013)) [86], there may be a sudden spike in number of tweets and vehicular traffic activity on Twitter. This is one of the drawbacks and advantages of data from social media that events are escalated quickly. Usually these events are anomalies and their variation can be normalized over time. Moreover whenever there is a spike in vehicular traffic activity at some location visualization of topics can provide much better insight to handle such situations.

# CHAPTER 4
# IMPLEMENTATION

The foundation of this study is data. Therefore, the first task is to extract and store data in an efficient manner. Once the data is handled in real-time using streams and stored in the database the next step is to convert this data into a usable format. Therefore, cleaning of the data is required and later feature and information extraction is required. Several kinds of information are extracted in different ways or using different tools. This information retrieval tasks happens using feature extraction techniques that include state-of-the-art libraries that aid in the text extraction.

After feature extraction comes model building where features are used in classification model development or for other spatial, temporal, and other kinds of analysis. Verifications and validations are performed by analyzing the output of the models. Most of the verifications are performed by visualizing the output and matching the observed counts for various tasks with expected counts. Validation of classification and other machine learning based models are performed using a technique known as cross-validation as previously explained. Validations of other ranking based analyses is performed using statistical techniques such as hypothesis testing. Rank-correlation tests are performed and root mean square errors are calculated to compare observed rankings and existing rankings, which are widely adopted in academia and the industry. Validations are performed using standard Python libraries described later in this chapter. After validating the models, analyses results are visualized. Apart from the statistically verified congestion and incident rankings, various kinds of novel inferences are also obtained. A typical lifecycle of any analysis performed using machine learning is depicted in Figure 5

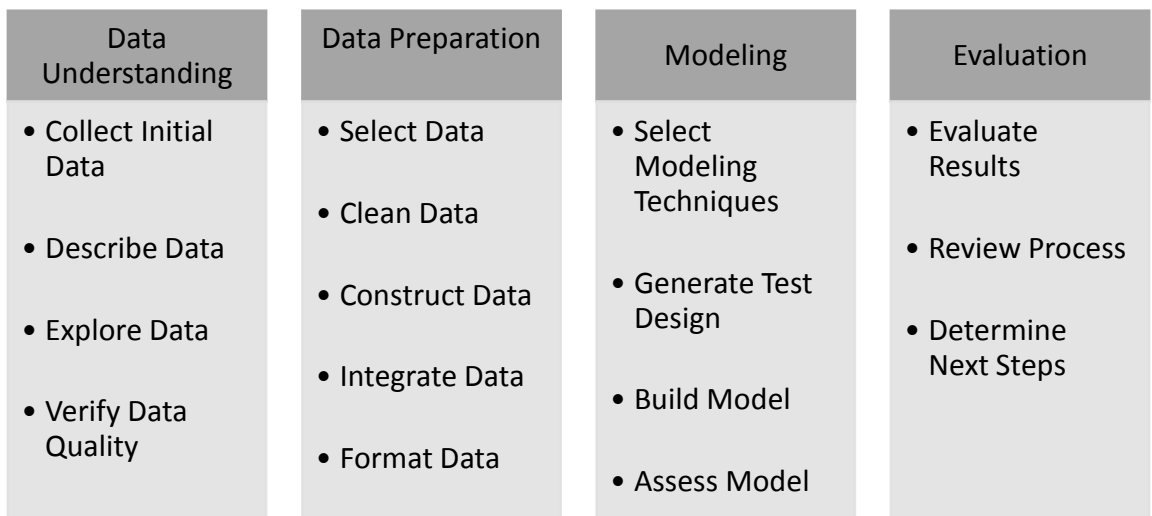

Figure 5 Machine learning life cycle

## 4.1 Language of Implementation

Python is used for almost all the tasks such as data streaming, data extraction, data cleaning, feature extraction, model generation, topic modeling, experimental and statistical analysis, etc. Even the final output, rankings, etc. are also obtained using Python. Database and dictionary creation for traffic incidents, traffic congestion, words related to traffic, etc. are performed using SQLite3 API of Python. It is easier to store and access the values using SQLite3 database and it can access the queries up to million lines with fairly fast speed.

## 4.2 Getting Data

Twitter's Streaming APIs [106] are used to download the data in real-time. Data can be extracted in two ways: i) based on geolocation using bounding boxes with south-west and north-east coordinates and ii) based on key-words. In first scheme, geotagged tweets lying within that boundaries are returned, while in second scheme tweets containing specific keywords are returned. Since, this study involves lots of spatial analysis the the first approach was used for streaming the data in real-time from Twitter. Applications on Twitter were created for streaming the public data. The application requires authorization keys (OAuth Keys) and tokens which once allocated by Twitter can be used for creating the application. There are some robust Python libraries for streaming Twitter data. One such library called Tweepy [104], which is used in current study for streaming the data. Generally the data from web is provided in XML or JSON formats. In the current study the format of the data being retrieved is JSON.

While Tweepy [104] is used for smooth streaming of data there were many issues encountered while downloading the data. Since, tweets were being downloaded 24 hours a day non-stop for several months, there were many errors which were faced such as data fallacy (empty data, faulty json pattern etc.), connection errors, quota limits (there is an upper bound on number of Tweets that can be accessed from Twitter), encoding-decoding errors, etc. These issues were resolved using error handlers in Python and the process was restarted again whenever the automated process stopped. As part of this a proper date and time format is followed to collect all the tweets.

Collected data is verified by the end of each week to see whether the results are consistent. If the counts come out to be varying significantly from expected, then the tweets are checked and visualized to find the anomaly. Many times these anomalies happened due to events such as game day (local effect), or tweets by/or corresponding to a celebrity. In such cases the anomalies (traffic activity due to celebrities etc.) are removed from the raw data.

### 4.3 Creating Database

After making sure that the data retrieval is smooth, the next step is to create relevant databases. The information which is being collected for each sample includes: tweet, date, time, latitude, longitude, location (city), and user id. A database is created which contains all these information without any refinement. This database is known as raw database. The format of the data being stored is TSV (tab separated values). Now filters are created on top of this raw database to create a traffic database. These filters are discussed in detail in sections 3.1 and 3.2.

The traffic related dictionaries are stored as python objects using a python library known as Pickle [70]. Using Pickle, python objects can be stored. The traffic dictionary is used in many places including feature extraction, topic modeling, etc. In the same way the incident dictionary is also created and stored. These dictionaries contains all the relevant words appearing multiple times or are synonyms of traffic related words or those obtained as top topics after performing topic modeling on the data. Using the words from traffic dictionary and incident dictionary, a final traffic related database and incident database are generated and are stored for classification and Natural Language Processing. Apart from the traffic and incident database, a non-traffic database is also created. This is the database without any filter except that the string "traffic" should not be present in the tweet. If this filter would not have been imposed then the traffic related data would have been the subset of this non-traffic data, making the classification models hard to train.

Using these traffic, incident, and non-traffic database classification models are created. Firstly, features are extracted from these datasets and are stored in the form of Python objects using Pickle. Storing data in the form of python objects has many advantages such as data does not need to be read every time if processing is required. Since the samples being analyzed in current study are in millions extracting feature from the data is very time consuming job. If the data is already processed and features are stored as Python objects then they can be loaded as another variable in Python in minimal time. These features are then used for training and testing of the models.

Lastly, for spatial, temporal, incident, perception, and congestion analysis, corresponding databases were created. These databases have much smaller size and therefore there is no need of using Pickle for these. Furthermore, these datasets are used for visualization purposes which used altogether a different framework, therefore it is important to store this information in more generic form such as TSV or CSV.

**4.4 Feature Extraction, Tools and Softwares Used**

This study involves several kinds of statistical and numerical analysis. In-built libraries were developed to perform these tasks, but they were computationally expensive. Reasons for this could be lack of optimization, memory usage, absence of code at lower level, and not leveraging multi-core processing. All these advantages are provided by some Python libraries. For numerical linear algebra, one of the most prominent libraries, called Numpy [85], is used. Furthermore, Scipy [85] is used for statistical analysis.

For extracting the features using natural language processing technique a couple of state-of-the-art Python libraries are used. To extract parts of speech (POS) tagging, named entity extraction, tokenization, getting stop words ("is", "the", "a" etc.) Stanford CoreNLP [91], Natural Language Toolkit (NLTK) [60] is used. Furthermore, to get the term frequency inverse document frequency (Tf-Idf) vectorizer feature NLTK is used. To obtain neural network vectorizer, Google Word2Vec [30] library is used. Topics are obtained using the Gensim library [74].

**4.5 Model Preparation and Analysis**

Features which are extracted are then used in machine learning models. Firstly, features which are most important are obtained using various methods such as information gain, correlation, etc. The Python libraries used for feature importance are obtained from Scikit-Learn [84]. This library is one of the most robust machine learning libraries used currently for almost all the machine learning applications. After obtaining the most relevant features, they are used in preparing the models. Classification models implemented in current study are: Stochastic Gradient Descent, Naive Bayes Algorithm and Support Vector Machines. To implement these algorithms Scikit-learn [84] libraries are used. As mentioned previously that Scikit-learn is a very robust library, many parameters were tweaked to obtain the best results.

The analysis of the performance of these models is conducted using 10-fold Cross Validation technique as previously mentioned. One major challenge in preparing the models for this kind of data is to deal with the majority class. The number of samples of non-traffic and non-incident instances is very small. To deal with this issue, stratified sampling technique is used. For performing stratified sampling and 10-fold cross validation, Scikit-Learn library is used again. Various statistics primarily precision, recall, and accuracy are calculated for testing model performance. Model preparation is an iterative process. The parameters and features for which the model gave best performance on testing data (unseen/new data not used for training the models) are finally selected and their performance is noted.

Apart from classification, various other models are also prepared. These models generally use the data from features and extract various kinds of information. Other analysis such as spatial, temporal, incident, city level, traffic perception, etc. comes under this category. Since, these analyses are specific to the current research and are ad-hoc analyses their implementation is performed without using any libraries or tools. However, some of these analyses are used for validation and verification purposes. Their visualization provides good insight for verification while hypothesis testing is performed using Spearman rank correlation tests and nonparametric Kolmogorov Smirnov tests. These are the means for validation of ranks obtained in current research with existing well adopted rankings. For these statistical analysis, Scipy library [85] is used.

### 4.6 Visualization Tools

This research has numerous implementations using visualizations, in particular, spatial-temporal analysis requiring visualizations. In the current study each data point contains geolocation and time, which is used to infer lots of spatial and temporal information. Various kinds of indexes, ranks, and information are mapped based on geo-locations. Information can be an aggregated version at the city level or each and every individual sample can be a point on top of the map. Most of the analyses which are not related to classification, are related to some sort of rankings. To visualize all this information an Openstreetmap [65] based platform is chosen. A free open source platform called CartoDB [14] is used in current study. This platform integrates georeferenced data with shapefiles, which is again integrated with Openstreetmap for mapping functionality. PostgreSQL [71] is used as a query language to visualize only a few segments of the entire data. Because the platform already exist and is open source much of the coding which was performed for visualization is on CSS and PostgreSQL. CSS is a creative visualization language for managing the look of the data on the web pages. This is the final layer used for visualization.

The following are the steps for visualizing the information on map. First the data is loaded in the TSV format and then it is georeferenced on the go. Later PostgreSQL queries are run to extract only important pieces of information. In last step, the data is mapped based on geolocations and finally CSS is used to make it much more creative and intuitive.

**CHAPTER 5**

**RESULTS, DISCUSSIONS AND VERIFICATION**

Various kinds of analysis are performed in current study. This chapter provides the results which are findings as well as verification of models in some cases.

**5.1 Topic Modeling, Clustering tweets and database creation on entire dataset**

The top thirty topics are derived from all the aggregated data having the word "traffic" using an unsupervised Natural Language Processing technique called Latent Dirichlet Allocation. The geo-tagged database with the word "traffic" (from which traffic database is created) as shown in figure 1 is used in this analysis. Section 3.6.7 explains the process of topic generation or topic modeling. Obtained topics represent the perception of people towards the traffic. Apart from the top topics some of the most frequent tweets are also captured. Following are the 25 most frequent tweets from the aggregated data of U.S.:

**5.1.1 Top 25 most common tweets**

Note: Curse words have been redacted modified with "*".

('i hate traffic'), ('traffic'), ('f*** traffic'), ('in traffic'), ('f*** this traffic'), ('stuck in traffic'), ('so much traffic'), ('holy traffic'), ('this traffic is ridiculous'), ('traffic sucks'), ('this traffic'), ('f*** you traffic'), ('life is a traffic jam'), ('catch me in traffic'), ('f***ing traffic'), ('f*** all this traffic'), ('so much traffic), ('this traffic tho'), ('why is there so much traffic'), ('in traffic '), ('this traffic though'), ('i love traffic'), ('why is there traffic'), ('traffic blows').

As seen it is possible that some of the tweets may not be related to vehicle traffic, such as 'life is a traffic jam' demonstrating the challenge of distinguishing between vehicle traffic related tweets and non-traffic tweets.

**5.1.2 Top topics derived using LDA for two different classification models of analysis**

<u>5.1.2.1. Topics without word "traffic" in it</u>

Each topic consists of 10 words and with the probability with which these words appear in that topic. The word "traffic" is removed from the data to perform topic modelling to remove the bias caused by the word "traffic" as it exists in every tweet. The objective is to obtain all sets of words which when appear together to imply a traffic related tweet. The number associated with each word is the probability with which a word exist in a topic. Topics consists of words and may be thought of in a similar vein as keywords that are used to describe a book or an article . A topic model is a type of statistical model for discovering the underlying abstract "topics" that occur in a collection of documents. LDA provides the probability of each word representing that topic. A topic is built by combination of words and probability of each word represents its contribution to the topic. Following are the top 10 topics that are retrieved and the probability score of each word constituting a topic:

0.208*hate + 0.134*bad + 0.092*getting + 0.037*lane + 0.034*best + 0.034*ud83dude2d + 0.030*w + 0.029*parkway + 0.027*guys + 0.024*start

0.105*sucks + 0.077*worst + 0.074*life + 0.069*late + 0.045*first + 0.030*ud83dude24 + 0.022*may + 0.021*ppl + 0.021*awesome + 0.019*year

0.119*light + 0.082*play + 0.040*making + 0.040*miles + 0.032*past + 0.029*place + 0.027*working + 0.022*ur + 0.022*yes + 0.017*-)

0.291*amp; + 0.058*oh + 0.044*street + 0.035*nice + 0.030*crash + 0.029*driver + 0.028*fire + 0.014*"im + 0.013*90 + 0.010*incident

0.190*right + 0.050*15 + 0.039*though + 0.030*also + 0.024*deal + 0.019*haha + 0.017*web + 0.016*situation + 0.015*tweet + 0.014*killing

0.070*downtown + 0.041*ave + 0.032*must + 0.029*train + 0.022*cool + 0.019*kind + 0.017*el + 0.015*standing + 0.015*wonder + 0.015*telling

0.081*took + 0.072*always + 0.049*lot + 0.032*please + 0.029*hill + 0.019*stadium + 0.018*fix + 0.016*ud83dude05 + 0.014*bruh + 0.013*angels

0.164*today + 0.142*2 + 0.102*late + 0.060*street + 0.045*tonight + 0.044*sure + 0.034*sunday + 0.029*free + 0.028*swear + 0.025*instead

0.053*stopped + 0.051*hell + 0.049*whats + 0.049*last + 0.044*tho + 0.030*worth + 0.028*finally + 0.028*google + 0.019*listen + 0.016*passenger

0.094*minutes + 0.063*gonna + 0.056*front + 0.048*come + 0.045*freeway + 0.044*lane + 0.040*30 + 0.032*find + 0.031*min + 0.026*without

5.1.2.2. Topics with word "traffic" in it

0.072*traffic + 0.050*on + 0.027*at + 0.027*accident + 0.018*blocked + 0.016*rd + 0.015*is + 0.011*i + 0.011*hate + 0.010*lane

0.084*traffic + 0.023*is + 0.019*this + 0.018*on + 0.017*i + 0.011*at + 0.011*that + 0.010*all + 0.009*so + 0.007*me

0.085*traffic + 0.046*on + 0.038*go + 0.037*delay + 0.036*stop + 0.036*mins + 0.025*from + 0.025*fwy + 0.022*accident + 0.017*at

0.072*traffic + 0.026*on + 0.013*i + 0.012*is + 0.011*be + 0.010*my + 0.009*wreck + 0.008*at + 0.007*as + 0.007*just

0.068*traffic + 0.056*at + 0.032*on + 0.031*st + 0.028*stop + 0.028*ave + 0.027*accident + 0.026*or + 0.026*pdx911 + 0.024*ne

0.075*traffic + 0.052*i + 0.019*my + 0.017*me + 0.015*this + 0.011*so + 0.011*it + 0.009*just + 0.008*get + 0.008*hate

0.063*traffic + 0.016*wait + 0.016*i + 0.014*on + 0.013*no + 0.009*de + 0.008*is + 0.007*at + 0.007*today + 0.007*stop

0.057*traffic + 0.030*is + 0.019*you + 0.016*at + 0.014*stuck + 0.009*or + 0.008*your + 0.008*amp; + 0.008*have + 0.007*when

0.082*traffic + 0.023*se + 0.020*alert + 0.018*incident + 0.018*sent + 0.017*tac + 0.016*st + 0.015*at + 0.012*b + 0.012*police

0.072*traffic + 0.035*on + 0.032*at + 0.022*accident + 0.020*is + 0.009*ave + 0.008*i + 0.007*get + 0.007*stuck + 0.007*rd

The words like "ud823…." are actually smileys within the tweet converted in the UTF-8 format. Secondly, "amp" represents & again in UTF-8 format and so on. Generally speaking the character "i" within the topics is not the pronoun I but the names of inter-state highways (I-75 etc.). Furthermore, from top topics those tweets which have curse words are not shown and the curse words from the most frequent tweets are replaced with "*".

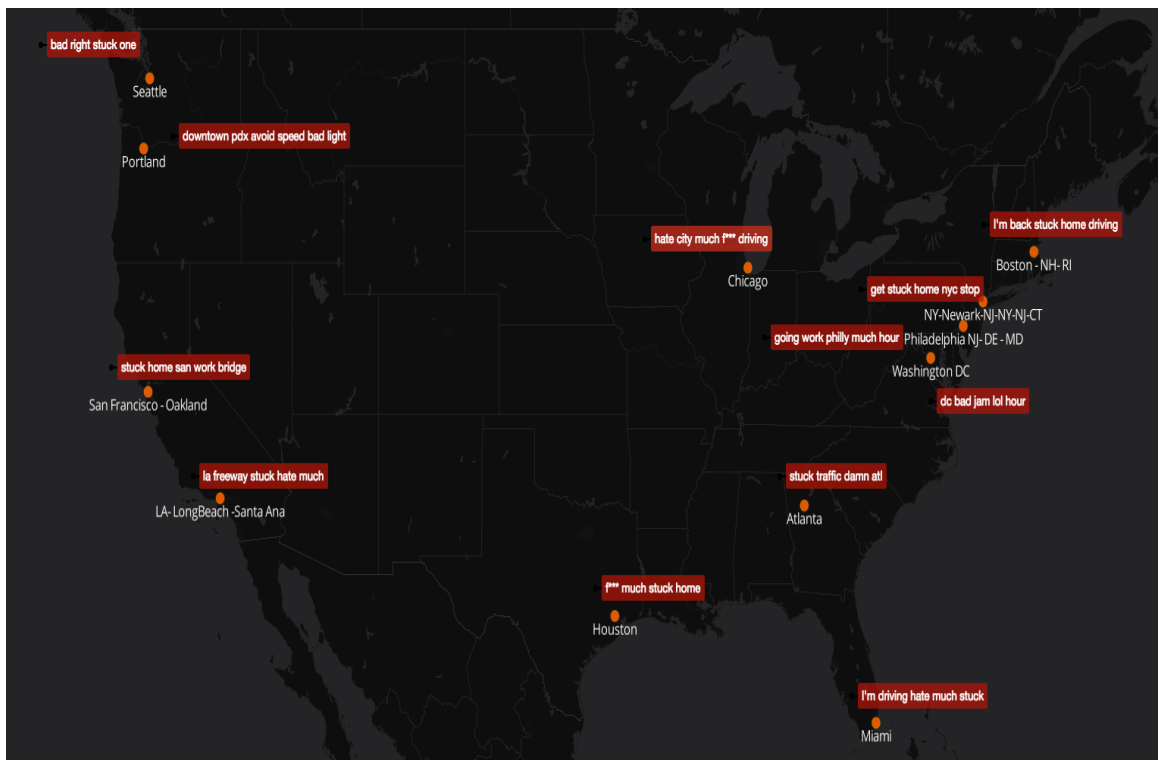

Figure 6 Visualization of top topics at city level

Table 9 Top Topics for Various Cities

| City | Top Topic |
|---|---|
| Washington DC | dc bad jam lol hour |
| LA- LongBeach | la freeway stuck hate much |
| San Francisco | stuck home san work bridge |
| NY-Newark-NJ-NY-NJ | get stuck home nyc stop |
| Boston - NH- RI | I'm back stuck home driving |
| Table continues… | |
| Houston | f*** much stuck home |
| Atlanta | stuck traffic damn atl |
| Chicago | hate city much f*** driving |
| Philadelphia | going work philly much hour |
| Seattle | bad right stuck one |
| Miami | I'm driving hate much stuck |
| Portland | downtown pdx avoid speed bad light |

**5.2 Traffic Activity and Incident Classification**

The primary aim of current study is to create mathematical models which are smart enough to identify any vehicular traffic activity or incident activity through tweets.

To extract this information, it is required to first classify whether any given tweet is related to vehicular traffic such as congestion, incident etc. or not. Once the mathematical models are able to provide good results then they can be used for any tweet in real-time. Following are the findings of the two different classification models created in this study:

**5.2.1 Traffic Activity Classification**

This first classification explores the ability to classify models in real-time. The traffic database and non-traffic database as seen in Figure 1 is utilized for this analysis. Using the classification models developed the performance of the algorithms are measured. There are two models of classification: (i) using the word "traffic" for learning models and (ii) models without the word "traffic" in them. The algorithms used for classification are: Stochastic Gradient Descent (SGD) and Naive Bayes Classifier (NB). Naive Bayes outperformed SGD and the results depicted in Table 10 and Table 11 are with Naive Bayes classifier.

**Sampling Method**: Stratified Sampling
**Sample Size:** 600,000 "non-traffic" tweets (10 % random sample of all 6 million non-traffic tweets from non-traffic database) and 120,000 "traffic" tweets from traffic database
*Model*: Naive Bayes
*Experiment*: 10-fold Cross Validation

Table 10 Results for model 1: Congestion and incident classification for models with word "traffic"

| Performance Metrics | Value |
|---|---|
| Accuracy | 99.85 % |

| | | |
|---|---|---|
| Precision | "traffic" : 99.20 % | "non-traffic" : 99.8 % |
| Recall | "traffic": 89 % | "non-traffic" : 99.98% |

Table 11 Results for model 2: Congestion and incident classification for models without word "traffic"

| Performance Metrics | Value | |
|---|---|---|
| Accuracy | 98.92 % | |
| Precision | "traffic" : 95.89 % | "non-traffic" : 98.91 % |
| Recall | "traffic": 58 % | "non-traffic" : 99.97% |

Table 10, 11 represents the performance of the classification models prepared in current study. The overall accuracy of the models is fairly good for both the classification models for traffic. The precision and recall rates are very good for model 1, however this is being driven by the word "traffic" in the tweet. It should be noted that majority (~ 99.99 %) of the tweets obtained in real-time are for class 0, that is non-traffic dominates the database whereas the above analysis has 83% "non-traffic" tweets. The false positive rate (predicting "traffic" even though it is "non-traffic") for class "non-traffic" for model 2 is nearly 1 %, which is very large because most of the tweets from Twitter are "non-traffic" and 1% of that (majority class: "non-traffic") being classified as "traffic" indicating most of the tweets referred to as "traffic" will be "non-traffic". This demonstrates the difficulty that will exist in automating a real-time classification. One filter which may help restrict the false positives is finding the similarity of a tweet with top topics of traffic and setting an acceptance a threshold. That is, if a tweet has similarity above the decided threshold then consider it as a traffic related tweet. Future efforts will explore this next step.

**5.2.2 Incident Classification**

**Sampling Method**: Stratified Sampling

**Sample Size:** 100,000 "traffic congestion" tweets and 20,000 "incident" tweets

***Model***: Naive Bayes

***Experiment***: 10-fold Cross Validation

Once a tweet is classified as traffic related, then the second classifier which is implemented on the top of it is to test whether the tweet is related to incident or congestion. The traffic database is then segregated to incident database and non-incident traffic database. These databases are used for incident classification. Same models which were used in traffic activity classification are used to perform this classification just that the databases are changed. Results for incident classification are shown in Table 12. It is assumed that all those traffic related tweets which are not incidents are related to congestion. Also, this test is based on tweets that have been correctly identified as "traffic".

Table 12 Results for incident and congestion classification

| Performance Metrics | Value |
|---|---|
| Accuracy | 77.64 % |
| Precision | "congestion" : 81.07 %     "incident" : 39.57 % |
| Recall | "congestion" : 93.17 %   "incident" : 17 % |

It is clear from Table 12 that the accuracy of the model is low. Furthermore, precision and recall rates for incident class are indicate a weak model. One potential reason for this poor result is lesser number of samples for model training. There is not sufficient samples with respect to the features and the model tends to underfit with respect to the number of samples. Reducing number of features or trying new models may also improve the performance but it is not currently tested.

### 5.3 Spatial and Temporal Variation of Twitter for Vehicular Traffic

Traffic activity on Twitter and also on the streets vary with time and space. Some locations are more prone to congestion and incidents than others. Urban areas depict more traffic activity than the sub-urban and rural areas. Within these areas some signals, street segments etc. have higher congestion and incident rates than others. Similar to spatial variation, traffic activity varies with time as well. Peak hours constitute majority of traffic activity (TTI) [95] and it is important to track these activities so that resources can be utilized in better way. In this section, visualization of vehicular traffic activity with respect to time and space is prepared to obtain better insights about the traffic information from Twitter. The results obtained in this section also provide verification for various studies performed in this study.

**5.3.1 Spatial Analysis**

Figure 7 represents the heat map of spatial distribution of vehicular traffic activity on tweets obtained through Twitter (i.e. all geotagged traffic related tweets recorded over five month period). Traffic database is used to perform this analysis. Therefore, 120,000 traffic related tweets are mapped according to their geo-location. Areas with highest intensity (visualized in red) are generally the most congested areas in U.S. Generally, they are medium or large size urban areas. It is clearly visible from the figure that there is a large area in red close to the east coast in the north. It is the most congested zone in the U.S. and it is evident from TTI rankings that four (NY, Boston, Washington D.C., Philadelphia) of the top 10 most congested cities lies in that zone. The other six cities are: LA, San Francisco, Houston, Atlanta, Chicago and Seattle. It is also clearly visible from the figure that these cities are also highly congested. The other clusters are close to the remaining six cities. The congestion index are calculated for all the cities in later section, that provides much better insight by implementing a quantitative method for calculating the level of congestion for each city.

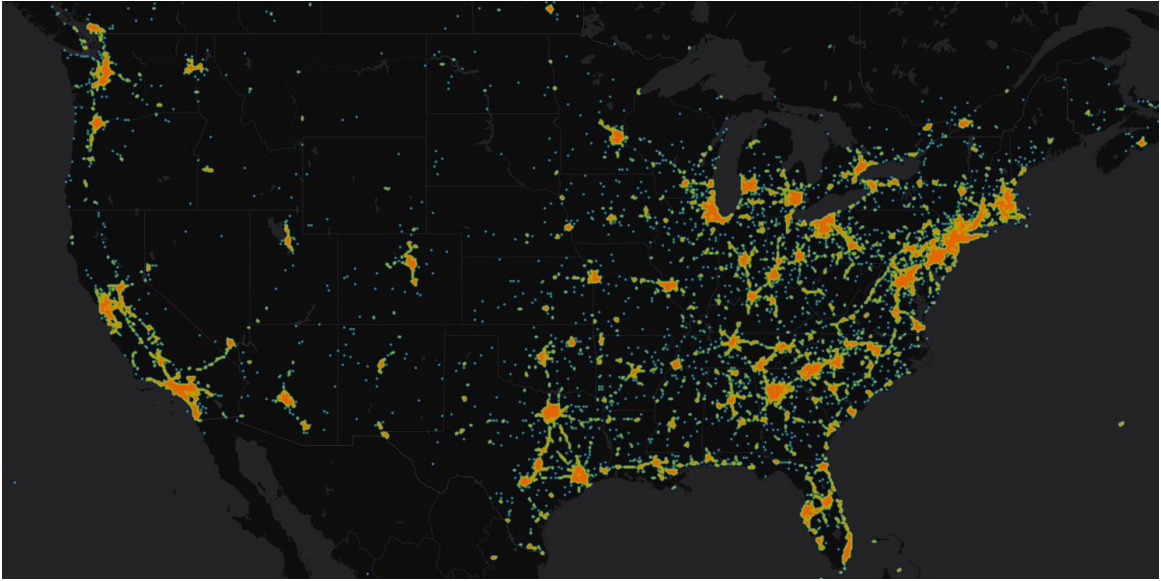
Figure 7 Vehicular traffic activity on Twitter

**5.3.2 Temporal Analysis**

The traffic activity on twitter varies according to the time as it happens on the ground. Almost 60% of the entire traffic related twitter activity lies in the morning (6-10 AM) and evening peak hours (3-7 PM). The Traffic database is used to perform temporal analysis. The temporal variation of all the traffic related tweets with respect to their local time are collected and their corresponding histogram is created. Figure 8 represent the aggregated histogram of traffic activity on Twitter. Note that Twitter is most during the late afternoon. It is interesting to note how strongly the appearance of this graph matches the typical vehicle traffic graphs for urban areas.

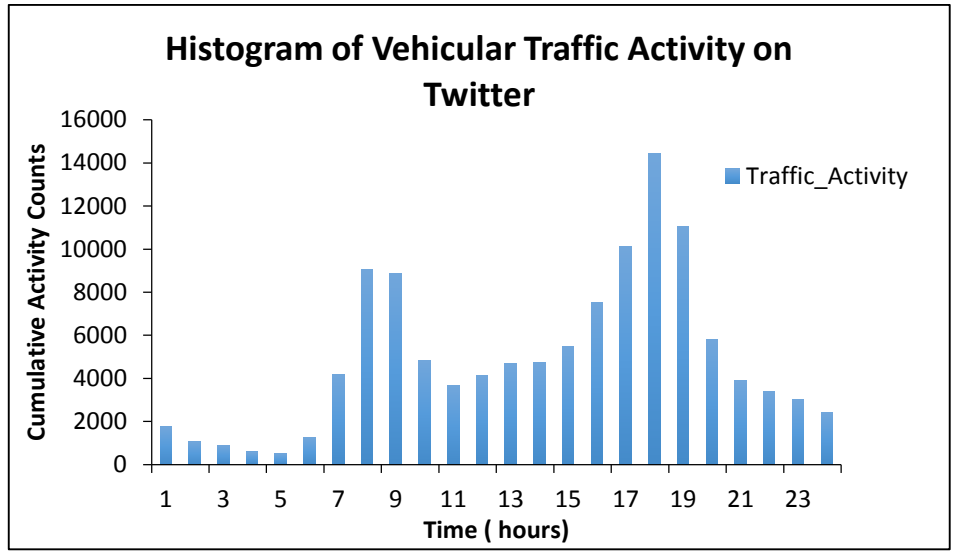

Figure 8 Histogram of vehicular traffic activity on Twitter

## 5.4 Peak hour and Traffic Flow Analysis

Peak hours are the primary focus in most of the traffic analysis, as 60% of the entire traffic congestion happens during the peak hours (TTI) [95]. This section provides better insight and variation of vehicular traffic activity on Twitter. The database used here is traffic database. Tweets are extracted which were sent during the peak hours and are then plotted according to their geolocation.

**5.4.1 Peak Hour Analysis**

On closer look at Figure 9 for morning peak hours (6-10 AM) and Figure 10 for evening peak hours (3-7 PM) that the two distributions are not much different in nature. Evening peak hours has longer trails and the heat maps at various locations and have more intensity than those of in the morning peak hours. The nature of both the distributions is almost similar because in the morning, traffic flows inwards (towards the center of the cities) and in the evening traffic flows outwards (away from the centers of the cities in outwards direction). Generally, people commute through the same routes for their work and businesses during morning and evening, which is the reason why the nature of the distributions are similar but with opposite directions. The evening peak hours have longer trails because people prefer doing other regular activities such as grocery shopping etc. (trip chain) in the evening.

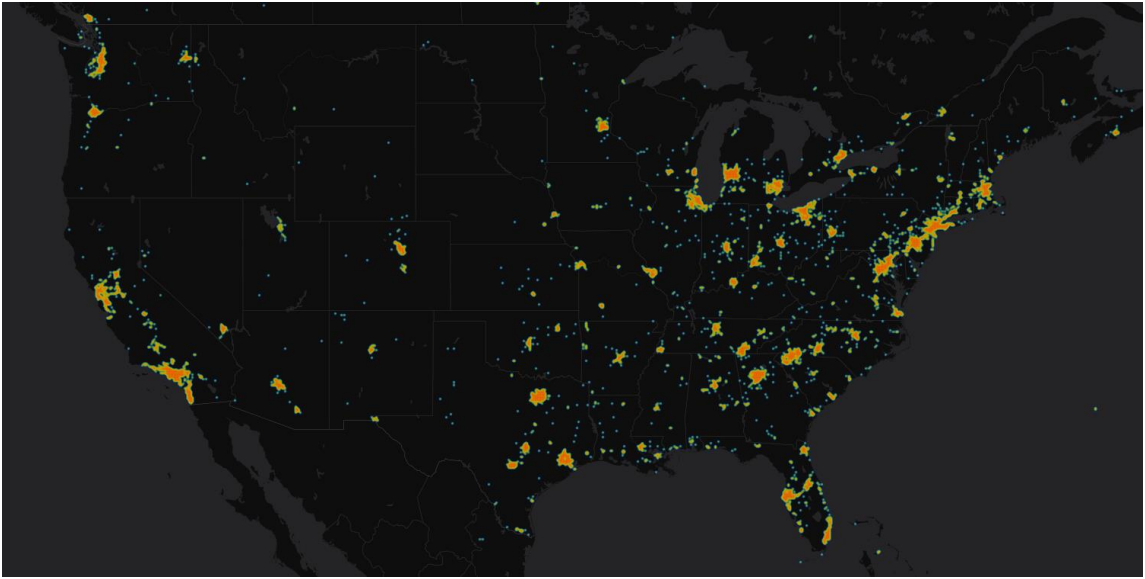
Figure 9 Vehicular traffic activity during morning peak hours (6-10 AM)

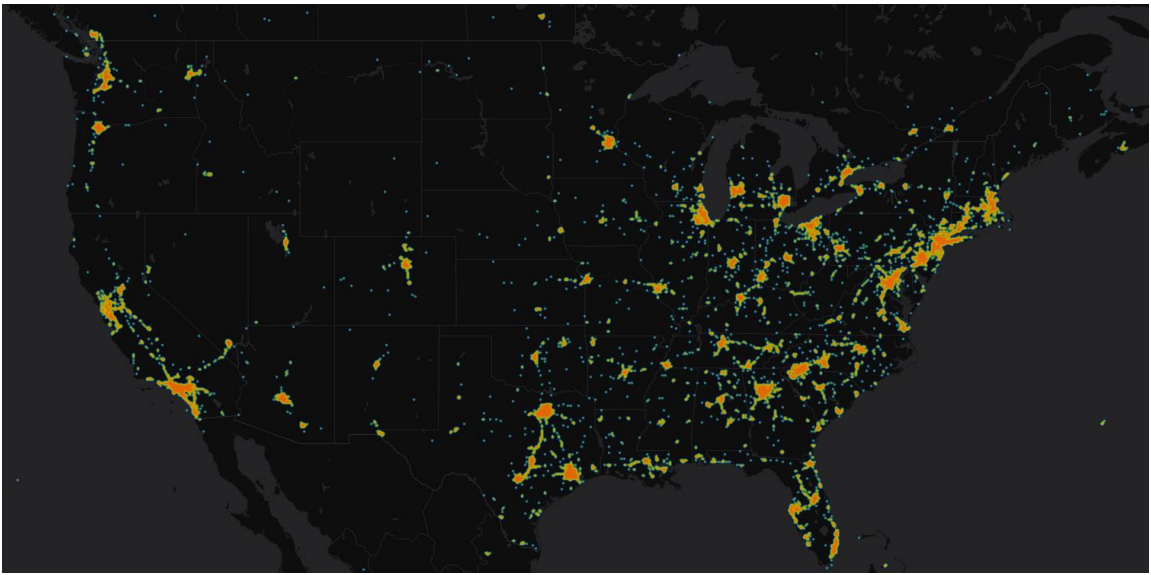
Figure 10 Vehicular traffic activity during evening peak hours (3-7 PM)

When comparing the peak hour distributions with aggregated traffic activity (Figure 7), it is clearly visible that the aggregated data map has much longer trails and intensities as the combined or the entire traffic database is the aggregation of the vehicular traffic activity on Twitter and is not restricted to only peak hours. Figure 7 is representation of almost each and every trip captured on Twitter unlike those in the peak hours, which are mostly due to daily commutes.

**5.4.2 Traffic Flow Analysis**

Using the same information traffic flow at various locations is also observed across the day. It is found that traffic activity on Twitter behaves very similar to what happens actually on the ground. Traffic gradually starts from the exterior/outwards of the cities (where population resides: residential zones) towards the business and economic zones in the cities (generally in the interior and center of the cities) in the morning at around 5-10 AM and then the activity remains within the city until around 4 PM. After 4 PM traffic slowly flows to the exterior of the cities following the trip chain in the direction of residential zones. Still most of the activity remains in the vicinity of the internal part of the cities and then at around 7 PM starts fading away. Traffic activity slowly fades away continuously till around midnight. Finally, the traffic is almost relatively inactive during 1-4 AM and the cycle continues. A much better insight can be obtain at city level and on per hour basis. A use case with visualizations is provided in later section which explains the traffic flow patterns.

**5.5 Traffic Perception Results**

Table 13 depicts the perception index calculated in the current study for various cities. As previously mentioned perception index is the attitude or perception of people towards traffic. The lower the score the more that city is sensitive to traffic incidents and congestions. This analysis uses the traffic database, as shown in Figure 1. Moreover, traffic dictionary is also used to obtain the perceptions scores for various cities. Tweets for various urban areas (New York, Atlanta, SFO etc.) are extracted from traffic database based on their geo-location and then perceptions score for each city is obtained. Perception index score is the mean of perception score for all the traffic related tweets for a city.

Houston and Portland are seen as the cities which are most sensitive to the traffic activities. Houston is actually an anomaly and it is because Houston has some dedicated professional Twitter users who constantly update the city with tweets regarding the incidents, congestion etc. While these professional tweets were removed in the initial filtering based on the sender twitter ID it was found that there are several retweets of these professional tweets. Even after limiting the retweets, there were significant traffic related tweets which were variation of these professional tweets from numerous users. Another reason for this anomaly can be social media: on social media activities gets escalated quickly. If a city has more number of professional users pumping the tweets they would be escalated more quickly than those in other cities with lesser professional users.

However, Portland is altogether a different case. The city government is renowned for its superior land-use planning and investment in public transportation (Urban Mobility Corporation, 2003) [109]. Portland is frequently recognized as one of the most environmentally conscious or "green" cities in the world because of its high walkability, large community of bicyclists, expansive network of public transportation options and 10,000+ acres of public parks (Sheppard et al., 2007) [87]. One of the main reasons for these achievements of Portland is community participation. In today's world a lot of participation happens through social media. Twitter is known for promoting such activities and agendas on social media. All these reasons may make Portland much sensitive to traffic activities than its peers.

From Table 13 it is clear that most of the large metropolitan areas are less sensitive to traffic related activities than the medium size cities. A potential explanation for this is that the delays and long drives are common to large cities. When delays and long drives becomes inherent parts of people's life then they might get used to tend to become more accustomed to it. However, cities still are active on Twitter but are relatively less active than medium size cities such as Columbus, Virginia Beach, Pittsburg etc. Figure 11 is the visualization for the cities with the top 29 perception scores. Cities in red are more sensitive to those in yellow.

Table 13 Perception Index scores

| Rank | City | Perception Index Score | TTI Travel Time Index, Rank | Yearly Delay Hours Auto per Commuter (TTI) |
|---|---|---|---|---|
| 1 | Houston | -8.18 | Houston - 6 | 52 |
| 2 | Portland, OR | -4.30 | Portland, OR - 19 | 44 |
| 3 | LA- Long Beach - Santa Ana | -3.65 | LA- Long Beach - 2 | 61 |
| 4 | Columbus OH | -3.38 | Columbus OH - 26 | 40 |
| 5 | Denver-Aurora | -3.06 | Denver- Aurora - 14 | 45 |
| 6 | Baton Rouge, LA | -3.01 | Baton Rouge - 21 | 42 |
| 7 | Pittsburgh PA | -2.99 | Pittsburgh PA - 28 | 39 |
| 8 | Bridgeport-Stamford CT-NY | -2.98 | Bridgeport-Stamford - 22 | 42 |
| 9 | Virginia Beach, VA | -2.97 | Virginia Beach, -20 | 43 |

Table continues…

| | | | | |
|---|---|---|---|---|
| 10 | Detroit MI | -2.95 | Detroit MI - 27 | 40 |
| 11 | Nashville-Davidson, TN | -2.94 | Nashville-Davidson, TN - 12 | 47 |
| 12 | Baltimore MD | -2.92 | Baltimore MD - 23 | 41 |
| 13 | San Jose | -2.91 | San Jose – 29 | 39 |
| 14 | Boston - MA - NH- RI | -2.89 | Boston MA - 5 | 53 |
| 15 | Indianapolis IN | -2.88 | Indianapolis IN - 24 | 41 |
| 16 | Seattle WA | -2.87 | Seattle WA - 10 | 48 |
| 17 | Dallas-Fort Worth-Arlington TX | -2.83 | Dallas-Arlington TX - 13 | 45 |
| 18 | NY-Newark-NJ-NY-NJ-CT | -2.82 | NY-Newark-NJ - 4 | 59 |
| 19 | Las Vegas | -2.80 | Las Vegas - 18 | 44 |
| 20 | Chicago | -2.78 | Chicago - 8 | 51 |
| 21 | Austin TX | -2.78 | Austin TX- 17 | 44 |
| 22 | SFO- Oakland, CA | -2.71 | SFO- Oakland, CA -3 | 61 |
| 23 | Charlotte NC-SC | -2.70 | Charlotte NC-SC -25 | 40 |
| 24 | Atlanta | -2.62 | Atlanta - 7 | 51 |
| 25 | Philadelphia PA NJ- DE - MD | -2.55 | Philadelphia PA - 9 | 48 |
| 26 | Memphis TN-MS-AR | -2.52 | Memphis TN-MS-AR - 30 | 38 |
| 27 | Washington DC-VC-MD | -2.06 | Washington DC - 1 | 67 |
| 28 | Orlando, FL | -1.93 | Orlando, FL - 13 | 45 |
| 29 | Miami FL | -1.58 | Miami FL - 11 | 47 |

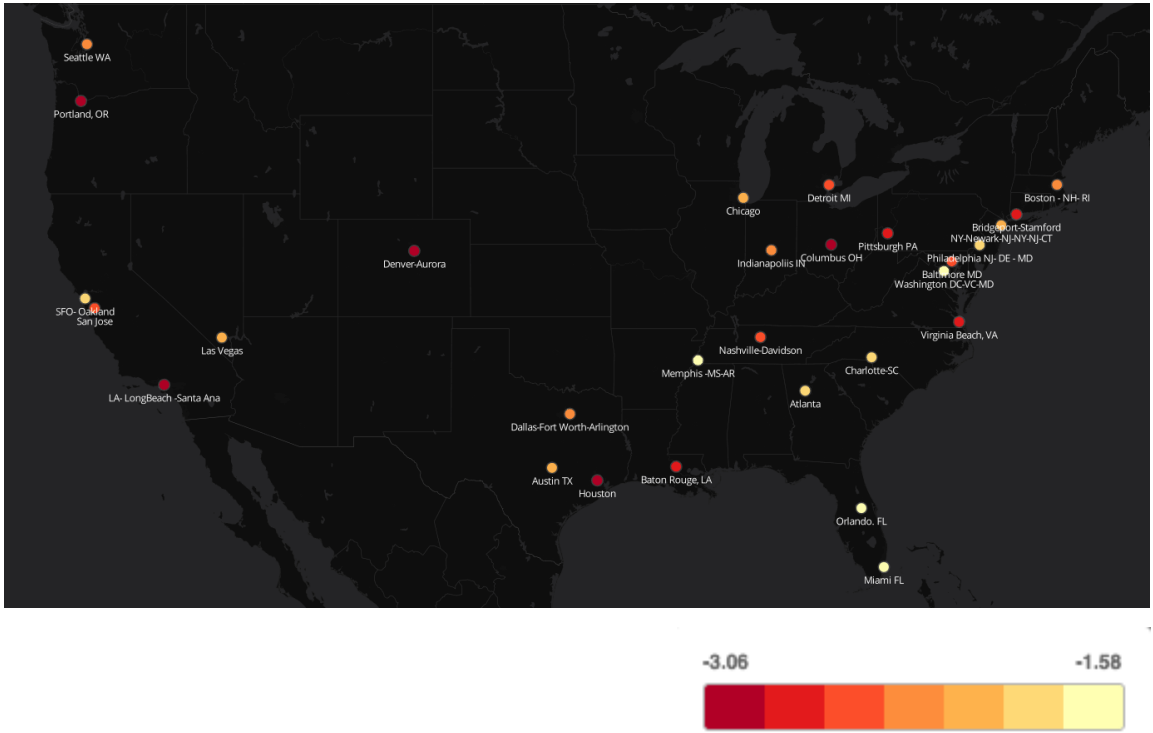

Figure 11 Traffic Perception Index for top 30 cities

Traffic perception is a very important parameter for considering or comparing traffic between various cities. Generally, most of the existing travel time index based rankings such as TTI, INRIX, TomTom [95, 36, 101] etc. do not consider perception while calculating the rankings. There is a potential drawback in the methodology of these rankings (TTI: yearly delay hours). For example, 30 minutes of travel time and delay of 15 minutes is too much for city like San Jose when compared to that of Washington D.C. where the exact same figures for travel time and delays might be normal. It is important to account traffic perception as well in the rankings. As seen in Table 13 then TTI congestion index does not align well with the perception scores, with many of the cities having much high perception scores having significantly lower than congestion rankings.

## 5.6 Congestion Index and Comparison with TTI's Travel Time Index

When comparing just the ranks of most congested cities obtained from TTI's traffic index [95] and tweet based congestion index for various cities calculated in the current study (see section 2.2 and 3.6.4 for description), it is clearly visible that they follow the same trend. Traffic database is used for this analysis. Tweets for various urban areas (New York, Atlanta, SFO etc.) are extracted from traffic database based on their geo-location and tweet normalization scores for each city are obtained. Table 14 depicts the top 30 most congested cities ranked in descending level of congestion.

Table 14 TTI rank vs. Tweet normalization ranks

| TTI Travel Time Index (Ranking, City, Yearly Delay Hours Auto per Commuter) | | | Tweet Normalization (Rank, City, Tweet Normalization Factor) | | |
|---|---|---|---|---|---|
| 1 | Washington DC | 67 | 5 | Washington DC | 9.66 |
| 2 | LA- Long Beach | 61 | 1 | LA- Long Beach | 20.58 |
| 2 | SFO- Oakland, CA | 61 | 6 | SFO- Oakland, CA | 8.35 |
| 4 | NY-Newark-NJ | 59 | 4 | NY-Newark-NJ | 12.48 |
| 5 | Boston - MA | 53 | 7 | Boston - MA | 6.8 |
| 6 | Houston | 52 | 2 | Houston | 18.23 |
| 7 | Atlanta | 51 | 9 | Atlanta | 9.37 |
| 7 | Chicago | 51 | 8 | Chicago | 5.86 |
| 9 | Philadelphia PA | 48 | 13 | Philadelphia PA | 3.07 |
| 9 | Seattle WA | 48 | 12 | Seattle WA | 4.07 |
| 11 | Miami FL | 47 | 10 | Miami FL | 5.12 |
| 11 | Nashville-Davidson, TN | 47 | 22 | Nashville-Davidson, TN | 1.18 |
| 13 | Dallas-Arlington TX | 45 | 11 | Dallas-Arlington TX | 4.43 |
| 13 | Denver-Aurora | 45 | 14 | Denver-Aurora | 1.99 |
| 13 | Orlando, FL | 45 | 19 | Orlando, FL | 1.51 |
| 13 | Honolulu HI | 45 | NA | NA | NA |
| 17 | Austin TX | 44 | 18 | Austin TX | 1.56 |
| 17 | Las Vegas | 44 | 24 | Las Vegas | 0.81 |
| 17 | Portland, OR | 44 | 3 | Portland | 15.12 |
| 20 | Virginia Beach, VA | 43 | 29 | Virginia Beach, VA | 0.43 |
| 21 | Baton Rouge, LA | 42 | 28 | Baton Rouge, LA | 0.54 |
| 21 | Bridgeport-Stamford | 42 | 27 | Bridgeport-Stamford | 0.60 |
| 23 | Baltimore MD | 41 | 20 | Baltimore MD | 1.46 |
| 23 | Indianapolis IN | 41 | 26 | Indianapolis IN | 0.74 |
| 25 | Charlotte NC-SC | 40 | 15 | Charlotte NC-SC | 1.90 |
| 25 | Columbus OH | 40 | 23 | Columbus OH | 1.03 |
| 25 | Detroit MI | 40 | 17 | Detroit MI | 1.63 |
| 28 | Pittsburgh PA | 39 | 16 | Pittsburgh PA | 1.79 |
| 28 | San Jose | 39 | 21 | San Jose | 1.40 |

| | Table continues… | | | | |
|---|---|---|---|---|---|
| 30 | Memphis TN-MS-AR | 38 | 25 | Memphis TN-MS-AR | 0.79 |

When performing the Spearman Rank Correlation Test, following are the findings:

$$spearman\ correlation\ coefficient, \quad \rho = 0.747$$
$$p\ value\ for\ null\ hypothesis\ that\ two\ sets\ are\ uncorrelated\ = 3.27e - 06$$

Based on the very high spearman correlation coefficient and almost zero p value of the rank correlation test, it is very clear that there is very strong correlation between the two rankings. Most of the cities have similar positions in the rankings, except for few. Reason for such anomalies (large displacement from the actual rank) may be due to the perception differences. That is, not only are the tweets generally more negative in cities more sensitive to traffic congestion there are also likely more traffic related tweets. Perception index scores are very different from the congestion index scores. It can be inferred after comparing the Tables 13 and 14 that the cities which are out of order in the congestion rankings actually rank very high in the perception index scores. This explains the fact that perception should also be considered while calculating the congestion index by various organizations such as TTI, INRIX etc.

### 5.7 Incident Index and Safety Rankings

Incident index represents how much a city is prone to incidents. The larger the number the more it is prone to incidents. Incident index is calculated in the same way as that of the Congestion Index, the only difference is that incident database is used for this analysis. Table 15 depicts the rankings obtained for top 30 most congested large and medium size cities. In Table 15 incident index is the column which is derived from current study, while the last column represents the relative accident likelihood compared to national average. This data is obtained from Allstate's America's Best Driver's Report [4]. This report contains 200 large and medium sized cities ranked according to their safety, however for current study only top 30 most congested cities of U.S. are considered and their rankings are used to perform Spearman's Rank Correlation test. When performing the Spearman Rank Correlation Test, following are the findings:

$$spearman\ correlation\ coefficient, \quad \rho = 0.578$$
$$p\ value\ for\ null\ hypothesis\ that\ two\ sets\ are\ uncorrelated = 0.00102$$

Based on the high spearman correlation coefficient and small p value of the rank correlation test, it is clear that there is strong correlation between the two rankings. Most of the cities are close to the actual rankings, except for few which is due to traffic perception (explained previously).

Table 15 Incident Index vs. Relative Accident Likelihood Compared to National Average

| City | Incident Index | City | Relative Accident Likelihood Compared to National Average |
|---|---|---|---|
| Houston | 19.70 | Baltimore MD | 1.29 |
| LA- Long Beach -Santa Ana | 14.66 | Boston - MA - NH- RI | 1.29 |
| Portland, OR | 11.74 | Washington DC-VC-MD | 0.97 |
| NY-Newark-NJ-NY-NJ-CT | 2.76 | Philadelphia PA NJ- DE - MD | 0.641 |
| Table continues… | | | |
| Washington DC-VC-MD | 2.59 | Miami FL | 0.584 |
| SFO- Oakland, CA | 1.70 | SFO- Oakland, CA | 0.546 |
| Table continues… | | | |
| Boston - MA - NH- RI | 1.59 | Pittsburgh PA | 0.51 |
| Miami FL | 1.38 | LA- Long Beach -Santa Ana | 0.485 |
| Chicago | 1.19 | Bridgeport-Stamford CT-NY | 0.412 |
| Seattle WA | 1.16 | NY-Newark-NJ-NY-NJ-CT | 0.411 |
| Atlanta | 0.83 | Atlanta | 0.39 |
| Dallas-Fort Worth-Arlington TX | 0.69 | Portland, OR | 0.372 |
| Philadelphia PA NJ- DE - MD | 0.68 | Dallas-Fort Worth-Arlington TX | 0.36 |
| Pittsburgh PA | 0.64 | Seattle WA | 0.36 |
| Denver-Aurora | 0.52 | Austin TX | 0.3 |
| Charlotte NC-SC | 0.45 | Houston | 0.29 |
| Detroit MI | 0.45 | San Jose | 0.274 |
| Austin TX | 0.39 | Chicago | 0.22 |
| Baltimore MD | 0.38 | Baton Rouge, LA | 0.19 |
| Columbus OH | 0.37 | Columbus OH | 0.19 |
| Nashville-Davidson, TN | 0.29 | Orlando, FL | 0.167 |
| San Jose | 0.23 | Charlotte NC-SC | 0.164 |
| Orlando, FL | 0.22 | Las Vegas | 0.154 |
| Las Vegas | 0.20 | Virginia Beach, VA | 0.13 |

| Virginia Beach, VA | 0.17 | Detroit MI | 0.106 |
|---|---|---|---|
| Table continues… | | | |
| Baton Rouge, LA | 0.17 | Memphis TN-MS-AR | 0.087 |
| Indianapolis IN | 0.15 | Denver-Aurora | 0.072 |
| Bridgeport-Stamford CT-NY | 0.13 | Indianapolis IN | 0.052 |
| Memphis TN-MS-AR | 0.12 | Nashville-Davidson, TN | 0.03 |

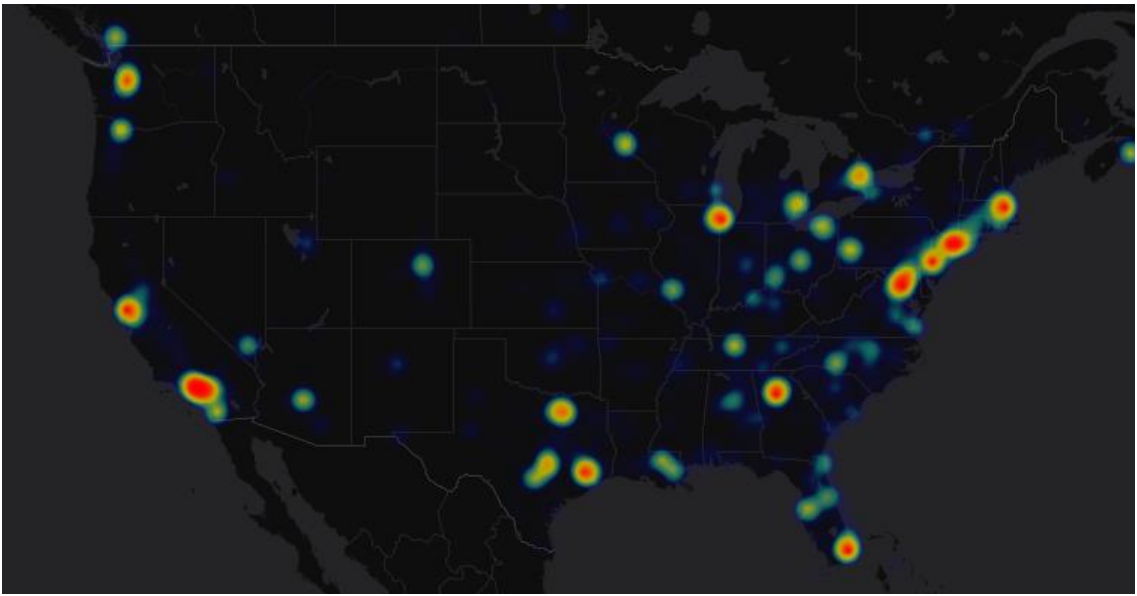

Figure 12 Traffic incident Index heatmap based on Twitter data

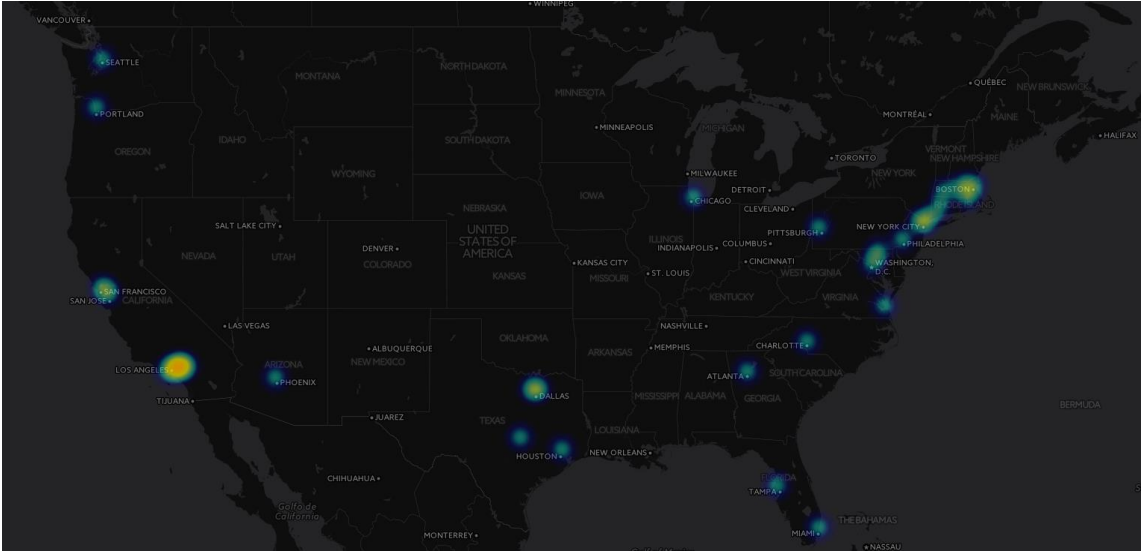

Figure 13 Relative accident likelihood compared to national average obtained from Allstate Rankings [4]

Safety rankings is just opposite to the incidents rankings. The lesser the incident index the safer the city. Figure 12 is the heatmap for incidents collected through Twitter data for all the entire US, while Figure 13 reflects the heatmap for relative accident likelihood compared to national average only for top 30 cities as provided by Allstate Rankings [4]. It is clearly visible from comparing the two figures: traffic incidents derived from Twitter and those obtained from Allstate Rankings [4] that Twitter provides a good reflection the actual incidents and congestion. Apart from the rank correlation test, verification of the traffic data from Twitter is also visible from these visualizations.

**5.8 Cities Statistics, Traffic Flow and Digital Signature of Traffic**

In current study, geotagged tweets are considered which is used to extract the data at cities level. Several kinds of analysis are performed at cities level, some of which have already been mentioned, for example congestion index, incident index, perception index, etc. One of the most important aspect of Traffic Engineering is to study traffic flows. After visualizing the aggregated tweets from traffic database on top of a map provides the patterns of traffic flow within the cities and across various geolocations. As mentioned before, it is assumed that the traffic related tweets (leading to creation of traffic database) are being sent by the commuters, drivers, passengers and pedestrians. Therefore, the vehicular traffic activity obtained on Twitter is the same as Twitter traffic activity in current study. Whenever traffic activity or Twitter activity is mentioned in current study it should not be confused on Twitter activity or Twitter traffic. Twitter activity or the traffic on Twitter is a different thing and is basically the rate of tweets being sent out on Twitter. They may or may not be geotagged and may or may not be related to traffic.

**5.8.1 Digital Signature of Traffic**

Each city has a unique pattern of vehicular traffic flow obtained through Twitter and the actual traffic on the streets. Vehicular traffic activity from Twitter can be used to develop a concept called Digital Signature of Traffic. The digital traffic signature for city of Atlanta is analyzed in current study. Traffic Signature for every 3 hours is mapped for the aggregated data for Atlanta. The following figures (Figure 14-21) represent the digital signature of traffic for Atlanta.

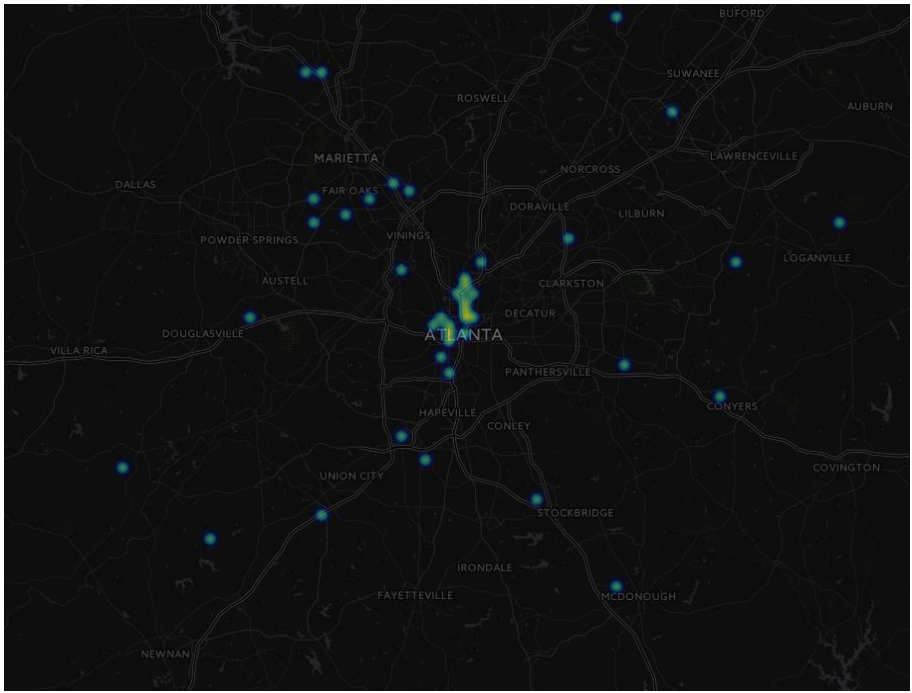

Figure 14 Traffic Signature of Atlanta from 0-3 AM

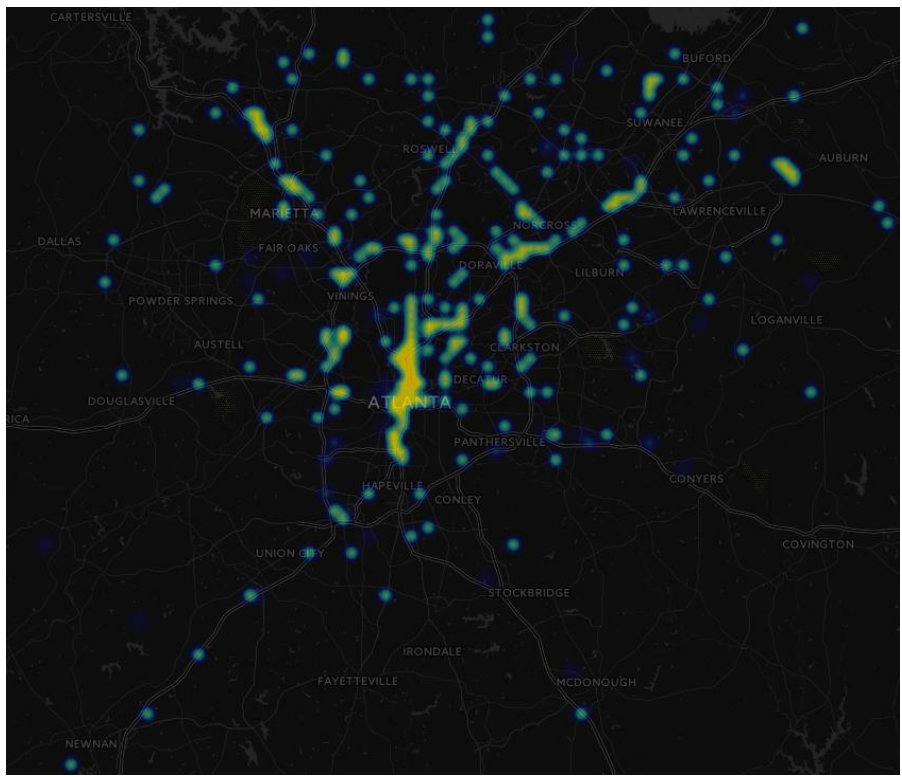

Figure 15 Traffic Signature of Atlanta from 3-6 AM

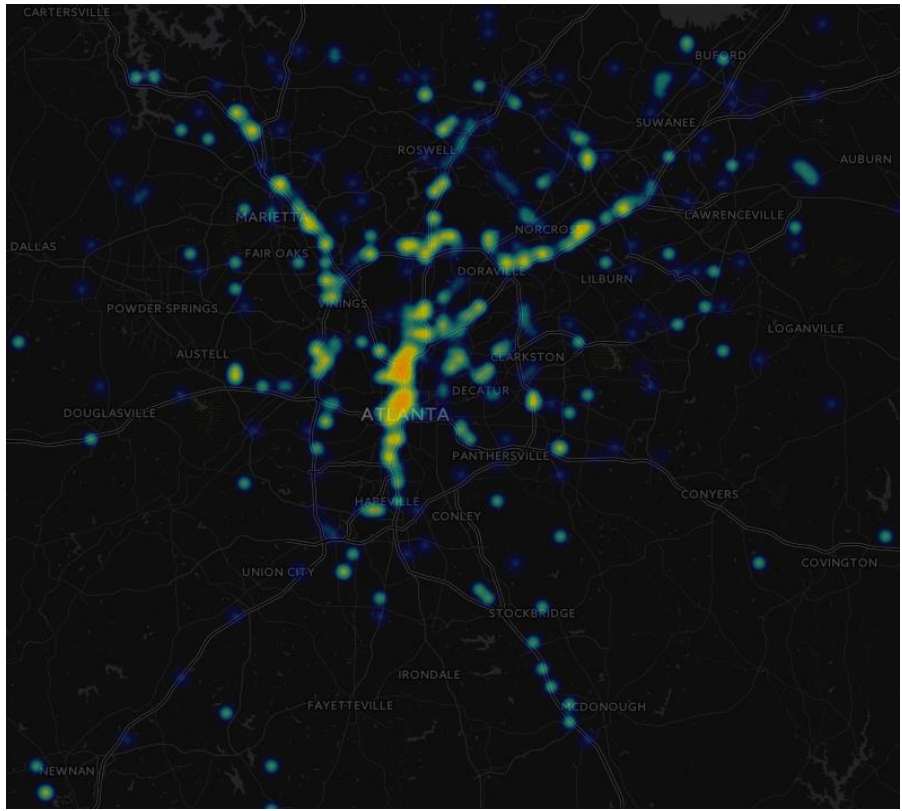
Figure 16 Traffic Signature of Atlanta from 6-9 AM

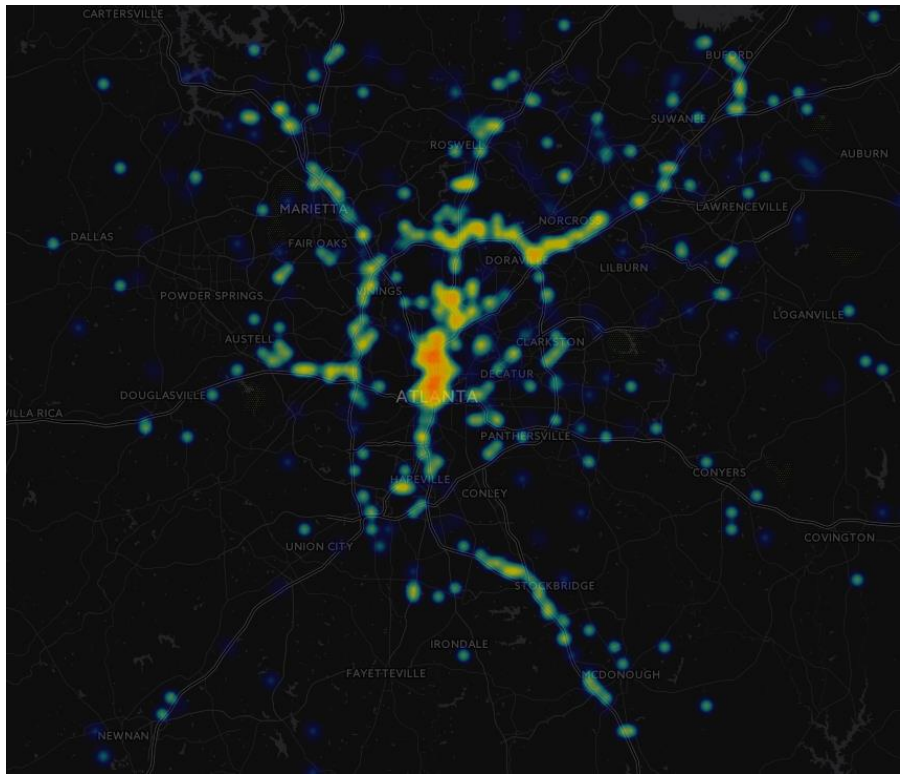
Figure 17 Traffic Signature of Atlanta from 9-12 PM

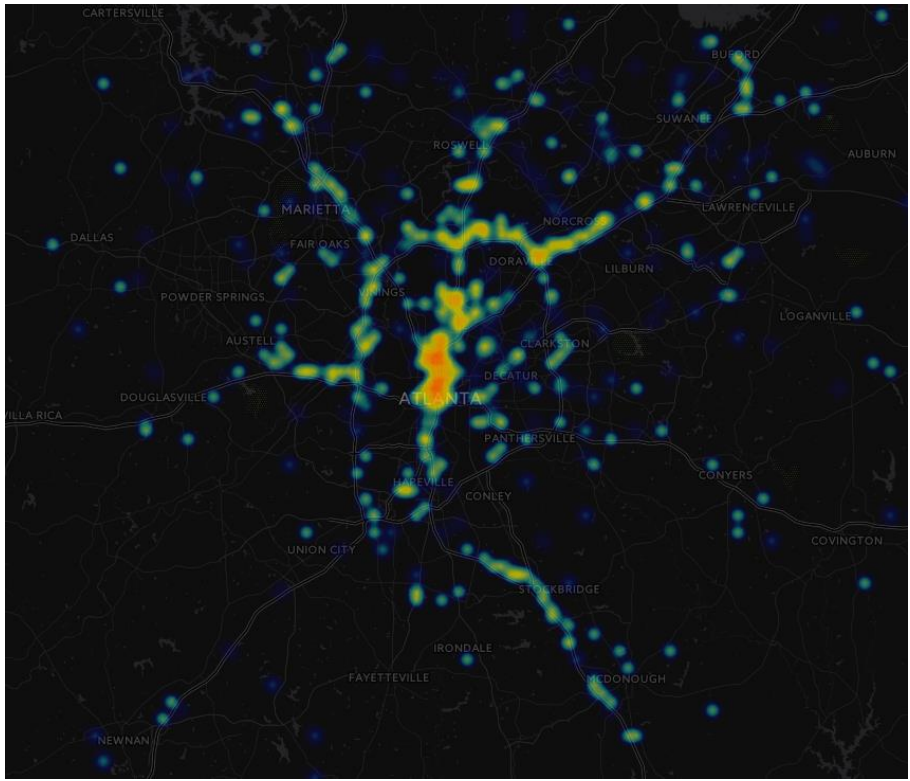

Figure 18 Traffic Signature of Atlanta from 12-3 PM

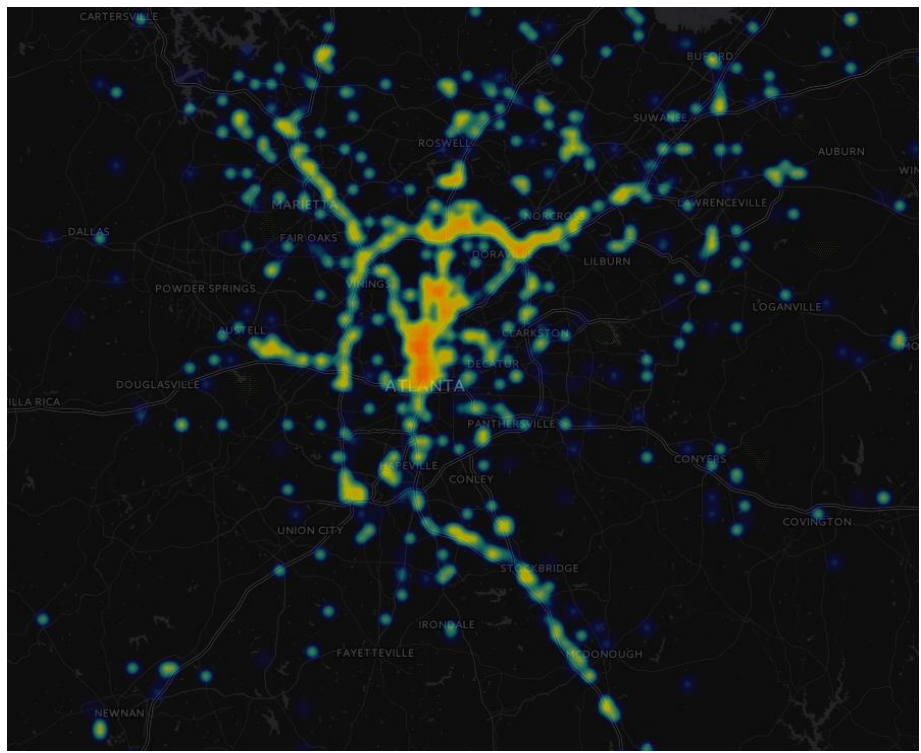

Figure 19 Traffic Signature of Atlanta from 3-6 PM

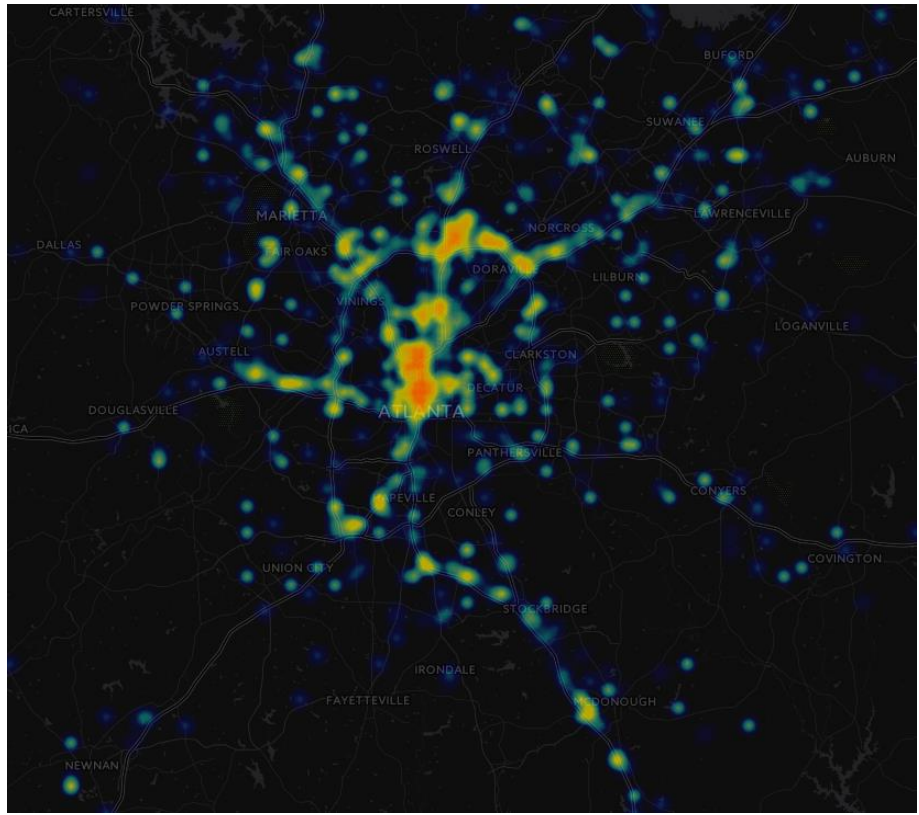

Figure 20 Traffic Signature of Atlanta from 6-9 PM

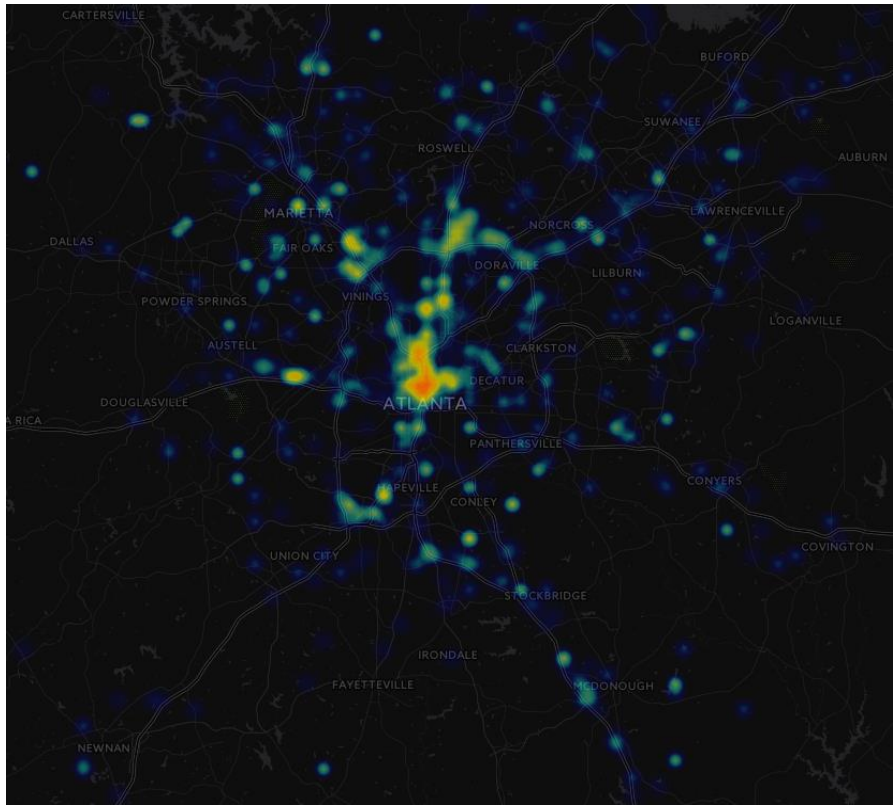
Figure 21 Traffic Signature of Atlanta from 9-11:59 PM

From the above mentioned figures (Figures 14-21) it can be inferred that the vehicular traffic activity on Twitter in Atlanta starts at around 4 AM at the areas around Atlanta such as Vinings, Panthersville, Marietta, etc. Vehicular traffic activity from Twitter increases in the morning peak hour from 6-10 AM it moves inwards the city of Atlanta. From 10 AM onwards traffic remains almost static and does not vary much until 3 PM. After 3 PM the vehicular traffic activity on Twitter again changes and moves outwards towards the counties around Atlanta. But this does not happen rapidly. Because of trip chains vehicular traffic activity on Twitter remains maximum during the evening peak hours of 3-7 PM and then slowly fades away. Vehicular traffic activity on Twitter also drifts from freeway corridors to interior streets within the counties. From heat maps it is visible that during morning and evening peak hours, most of the highways and freeways are highlighted. From midnight to 4 AM, the vehicular traffic activity on Twitter is hardly visible on the highways and freeways and they are within the counties.

The figures mentioned above represent the signature of traffic for Atlanta. These are static images. A visualization provides much better insights on the signature of the traffic for any city. The idea behind signature of traffic is that information which cannot be obtained using sensors and traffic counts using various equipment can be obtained using this digital signature. For example, there are many interior streets which do not have cameras or sensors and it is very tedious to get these kinds of information vehicle traffic information for these roadways. However, using social media people act as sensors and tweet about the issues they face. People can tweet about the incidents or other issues on the internal street. These types of information can be collected and used to improve the situation and to improve the quality of life. During an outbreak of traffic due to some natural incident such as snow etc., this technique can be highly useful. It is hoped that future work will be able to develop prediction models based on this data stream allowing for a reduction in congestion.

**CHAPTER 6**

# CONCLUSIONS AND RECOMMENDATIONS

Current study provides a novel methodology and data source for obtaining traffic related information such as congestion and traffic incidents. Real-time data from web is used as traffic sensors or the source of traffic information. Currently, Twitter is used as the primary data source. Twitter data is not bound by space and time unlike other sources of data (sensors for traffic counts etc.). Furthermore, this data is free, its processing is faster and it is observed from results that Twitter can be very effective in identifying traffic congestion and incidents in real-time.

Geotagged tweets within the boundaries of U.S. are streamed in real-time and relevant tweets are stored to retrieve traffic related information. After collecting the data two kinds of analysis are performed: (i) online analysis and (ii) offline analysis. Through online analysis, automated models are developed which can analyze the tweets in real-time and predict whether a tweet belongs to traffic congestion, traffic incident or not. While offline analysis is performed on the traffic related data stored over long period of time. This data stored over a long period of time is used for build or train classification models. Geotagged tweets are streamed at the rate of 1.5 million tweets per day. It is found that about 0.147% of all the geotagged tweets contain the word "traffic". All such tweets are stored, cleaned and analyzed. Data mining techniques are performed on top of these retrieved tweets to extract traffic related information. Over 120,000 traffic related tweets are retrieved finally over 50 days randomly sampled within the period of five months (September 2014 to February 2015).

Once the data is retrieved machine learning, artificial intelligence and natural language processing techniques are used to extract the traffic related information and make sense out of the data. Mathematical models are proposed which can infer the information prevailing within these tweets in the similar manner how humans infer the text. The idea is to automate the process of information retrieval for vehicular traffic from these tweets or any given text. Once the algorithms became smart enough to retrieve the information out of the text then they were used to create the classification models. The objective of these classification models is to identify the pattern and classify which tweet contains traffic related information such as congestion, incidents etc. and which does not. Those which are classified as traffic related tweets are further stored and are used for information extraction while others are scrapped. The performance of the classification models developed in current study is extremely good. Two different classification models for traffic congestion detection are proposed. The best model is able to classify a vehicular traffic activity with accuracy of 99.85% and mean precision and recall rates of 99.60% and 93.5% respectively. Since, traffic incidents are less likely to occur therefore its data was sparse because of which traffic incidents were predicted with lesser accuracy of 98.92% and mean precision and recall rates of 97% and 79% respectively. These classification models can be used in real-time for prediction purposes.

In the offline analysis, past data is aggregated over a five month period and analyzed to obtain various kinds of information such as congestion and safety rankings for various urban areas. The obtained rankings for safety and congestion are also validated and compared with with existing state of the art rankings provided by TTI, INRIX and Allstate Rankings [95, 4]. A completely novel technique and methodology for calculating traffic congestion and safety for various cities is proposed. The Spearman rank correlation tests are performed to check the validity of obtained rankings with existing rankings widely adopted by various organizations. Based on the very high spearman correlation coefficients and almost zero p values of the rank correlation tests, it is very clear that there is very strong correlation between the rankings proposed in current study and those widely adopted by others. Therefore, we can reject the idea that the correlation is due to any random sampling.

Apart from these rankings, a novel technique to obtain traffic perception for commuters at various cities is proposed. Currently, there exist no standard ranking system for calculating traffic perception. A concept of perception index is introduced. Perception index provides the sense of how people perceive traffic and their attitude towards the same. Perception index at cities level is calculated and various inferences are also retrieved from the obtained perception indices. Commuters at New York might perceive the traffic and corresponding delays and incidents differently than those in smaller urban areas such as San Jose. In current study, it is shown that perception is also an important factor which should be considered for various kinds of traffic analysis and comparison between various cities should be done accordingly.

The traffic related data is also mapped on top of US maps for various kinds of spatial analysis such as traffic flows. Visualizations are developed to map the vehicular traffic related activity on Twitter based on their geo-coordinates. The spatial distribution of the vehicular traffic activity on Twitter is very similar to the actual traffic activity on streets. Apart from spatial analysis, the data was plotted and visualized according to the time and the flow of traffic inwards and outwards the city is observed during the morning and evening peak hours. It is found that the traffic patterns observed during the peak hours is similar to what is expected for most of the urban areas. This variation of vehicular traffic activity on Twitter according to space and time gives realistic feel similar to the actual traffic flows. Therefore, it is concluded that the Twitter can also be a good source of traffic related information. It might not become the primary source of information however, when collected over a long period of time can reveal much more than what is collected through normal traffic sensors and manual counts on the streets.

Data from web in the form of tweets proved to be very informative with respect to vehicular traffic information. Therefore a new concept of "Digital Signature of Traffic" is introduced in current study. It is proposed that Twitter data or any form of textual data on web represents the signature or fingerprints of traffic. Every city has a unique traffic flow pattern obtained through Twitter. Since, the data considered in current study is converted in digital form to be analyzed through mathematical models and each city has a specific subset of tweets therefore these data converted in digital form represents the digital signature for any city.

In current study, an exhaustive vehicular traffic related study is performed using Twitter data. Various kinds of analysis and feature extraction techniques depicted in this study from Twitter data are unique with respect to traffic related studies. Current study is first of its kind. The objective is not just to create classification models able to predict traffic congestion and incidents but also to explore as much traffic related information as possible. This study is so rich in its content that even traffic perception and traffic flow patterns are also extracted from the Twitter data.

If a dedicated framework is developed for identifying the traffic related information (perception, congestion and incidents) from Twitter then it can help various DOTs and regional offices of traffic in ameliorating the worsening traffic situations. The core concept of entire study is that people reflect their perception and share any information that they possess in the form of news or social media. May be in the future some other platform other than Twitter or social media will exist and techniques might be even more advanced but the idea will not change. People act as sensors as events occur around them they tend to share it with others. The trick is to extract as much information as possible using people as sensors, instead of setting up a dedicated infrastructure for the same.

# APPENDIX A

## TRAFFIC DICTIONARY

This table contains all the words extracted from top topics and most common tweets.

The dictionary of these words is termed as traffic dictionary.

| accident | driving | downtown | commute | rush hour |
|---|---|---|---|---|
| block | injure | drive | crazy | bad |
| emergency | slow | light | time | stupid |
| street | f*** | drunk | highway | hit |
| congestion | shit | bus | city | hitting |
| insane | god | slow | lol | death |
| shut | stop | signal | late | fu**ing |
| collision | don't | home | wait | outbound |
| car | not | hell | rush | between |
| bus | way | morning | hour | inbound |
| alleviate | worse | evening | love | toll |
| hate | worst | night | day | much |
| traffic | haha | day | amp | holy |
| vehicle | #traffic | hit | gridlock | police |
| stopped | rush hour | transportation | freight | fence |
| drive | lane | shipment | transit | dicker |
| traffic accident | incident | service | transport | touch |
| traffic incident | against | travel | movement | market |
| blocked | now | influx | transfer | interact |
| Horrendous | freeway | barter | passage | handle |
| minutes | sw | exchange | flux | dont |
| delay | nw | connection | jam | fwy |
| massive | rd | communion | passengers | interface |
| least | mins | truck | vehicles | damn |
| rush | i- | close | cartage | drive |
| serious | sucks | peddling | truckage | contact |
| closed | blvd | network | trade | push |

# APPENDIX B

## STOP WORDS

This table contains all the words that are used as stop words mainly in feature extraction and topic modeling.

| i | them | does | after | both |
|---|---|---|---|---|
| me | their | did | above | each |
| my | theirs | doing | below | few |
| myself | themselves | a | to | more |
| we | what | an | from | most |
| our | which | the | up | other |
| ours | who | and | down | some |
| ourselves | whom | but | in | such |
| you | this | if | out | no |
| your | that | or | on | nor |
| yours | these | because | off | not |
| yourself | those | as | over | only |
| yourselves | am | until | under | own |
| he | is | while | again | same |
| him | are | of | further | so |
| his | was | at | then | than |
| himself | were | by | once | too |
| she | be | for | here | very |
| her | been | with | there | s |
| hers | being | about | when | t |
| herself | have | between | where | can |
| it | has | into | why | will |
| its | had | through | how | just |
| itself | having | during | all | don |
| they | do | before | any | should |

# APPENDIX C

## PROFESSIONAL USER'S TWITTER ID

This table contains the Twitter Ids of professional users on Twitter, who publish tweets regarding the Traffic.

| | | | |
|---:|---:|---:|---:|
| 249818911 | 29407288 | 2749163574 | 979531062 |
| 1606906374 | 27962969 | 199324673 | 1606472113 |
| 249405759 | 1589692776 | 64465376 | 979570111 |
| 65171105 | 39049373 | 20781010 | 930739604 |
| 54304799 | 56266341 | 2856823310 | 23074218 |
| 71052378 | 930753054 | 55540658 | 930781484 |
| 930758376 | 53969028 | 22651125 | 2715157034 |
| 249397229 | 979644014 | 55823412 | 1568864394 |
| 18984207 | 54988729 | 59678572 | 57976284 |
| 39822711 | 22932788 | 74998364 | 16374678 |
| 467751855 | 60452453 | 86938456 | 16246600 |
| 81913437 | 250322598 | 13719342 | 56117318 |
| 1301668658 | 55218049 | 478647114 | 70381121 |
| 58101529 | 24905550 | 149710722 | 70291924 |
| 92771483 | 831251732 | 593137955 | 1399647162 |
| 325862777 | 1709482140 | 282079946 | 16294929 |
| 2432324029 | 53932318 | 242984239 | 67972476 |
| 113711103 | 62050708 | 66179557 | 62413708 |